\documentclass[final,3p,times]{elsarticle}

\usepackage{color}
\usepackage{soul}
\usepackage{amsmath} 
\usepackage{amssymb}  
\usepackage{graphicx}
\usepackage{algorithm}
\usepackage{algpseudocode}
\usepackage{mathtools}
\usepackage{multirow}
\usepackage{nicematrix}
\usepackage{svg}
\usepackage{subcaption}

\usepackage{xspace}

\usepackage[hyphens,spaces,obeyspaces]{url}

\usepackage[title]{appendix}

\newcommand{\cE}{\mathcal{E}}
\newcommand{\cR}{\mathcal{R}}
\newcommand{\cB}{\mathcal{B}}

\newcommand{\cL}{\mathcal{L}}
\newcommand{\cM}{\mathcal{M}}
\newcommand{\cS}{\mathcal{S}}

\newcommand{\cC}{\mathcal{C}}

\newcommand{\be}{\mathbf{e}}
\newcommand{\bI}{\mathbf{I}}
\newcommand{\bt}{\mathbf{t}}

\makeatletter
\DeclareRobustCommand\onedot{\futurelet\@let@token\@onedot}
\def\@onedot{\ifx\@let@token.\else.\null\fi\xspace}

\def\eg{\emph{e.g}\onedot} 
\def\ie{\emph{i.e}\onedot} 
\def\cf{\emph{cf}\onedot}

\def\etal{\emph{et al}\onedot}
\makeatother

\journal{Aerospace Science and Technology}

\begin{document}

\begin{frontmatter}



\title{Event-RGB Fusion for Spacecraft Pose Estimation Under Harsh Lighting}


\author[adelaide]{Mohsi Jawaid\corref{cor1}}
\ead{mohsi.jawaid@adelaide.edu.au}

\author[adelaide]{Marcus M\"{a}rtens}
\author[adelaide]{Tat-Jun Chin}
\affiliation[adelaide]{organization={AI for Space Group},
            addressline={The University of Adelaide}, 
            country={Australia}}

\cortext[cor1]{Corresponding author}
\begin{abstract}

Spacecraft pose estimation is crucial for autonomous in-space operations, such as rendezvous, docking and on-orbit servicing. Vision-based pose estimation methods, which typically employ RGB imaging sensors, is a compelling solution for spacecraft pose estimation, but are challenged by harsh lighting conditions, which produce imaging artifacts such as glare, over-exposure, blooming and lens flare. Due to their much higher dynamic range, neuromorphic or event sensors are more resilient to extreme lighting conditions. However, event sensors generally have lower spatial resolution and suffer from reduced signal-to-noise ratio during periods of low relative motion. This work addresses these individual sensor limitations by introducing a sensor fusion approach combining RGB and event sensors. A beam-splitter prism was employed to achieve precise optical and temporal alignment. Then, a RANSAC-based technique was developed to fuse the information from the RGB and event channels to achieve pose estimation that leveraged the strengths of the two modalities. The pipeline was complemented by dropout uncertainty estimation to detect extreme conditions that affect either channel. To benchmark the performance of the proposed event-RGB fusion method, we collected a comprehensive real dataset of RGB and event data for satellite pose estimation in a laboratory setting under a variety of challenging illumination conditions. Encouraging results on the dataset demonstrate the efficacy of our event-RGB fusion approach and further supports the usage of event sensors for spacecraft pose estimation. To support community research on this topic, our dataset has been released publicly.
\end{abstract}




\end{frontmatter}




\section{Introduction}\label{sec:intro}
Spacecraft pose estimation is the problem of determining the 6-degrees-of-freedom (6DoF) pose consisting of the position and orientation of a space-borne object, typically a satellite. It is a critical step in a wide range of space applications, including rendezvous, close proximity operations, debris removal, refueling and on-orbit servicing~\cite{renaut2025deep, bechini2025robust, pasqualetto2019review, cavaciuti2022}. Robust pose estimation is paramount to safely and effectively executing these tasks~\cite{reed2016restorel, activedebrisremoval}.

\begin{figure}[t]\centering
\includegraphics[width=0.6\columnwidth]{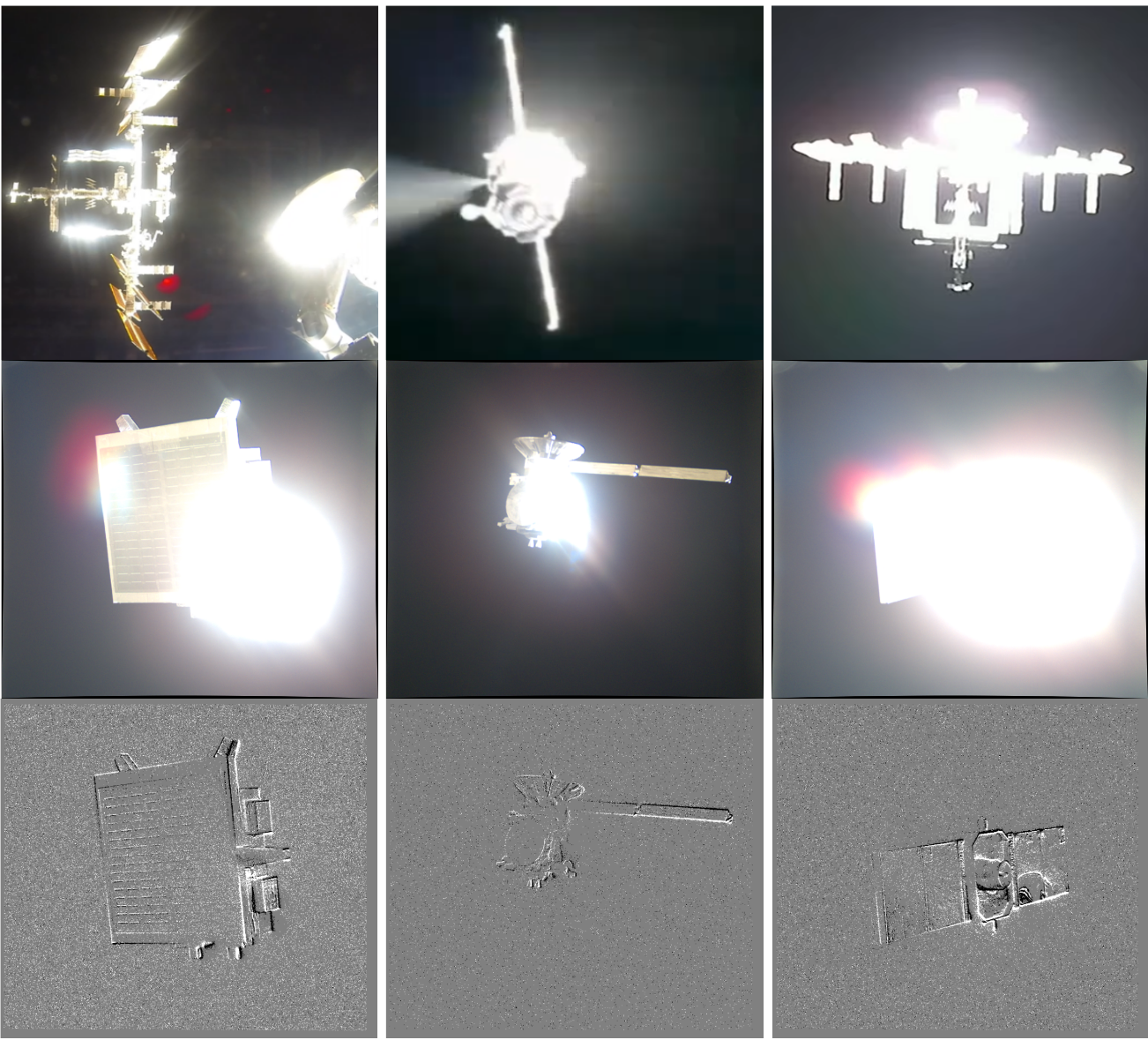}
\caption{A comparison of harsh lighting conditions in real docking and rendezvous scenarios~\cite{issdockingteaser1,  soyuzdockingteaser, issdockingteaser2} (top row) along with samples from our real dataset (RGB frames in the second row and optically aligned event frames in the third row). Even in the harshest example where most of the satellite is occluded by glare/over-exposure in the third row third column, an event camera is able to see the structure of the object.}
\label{fig:teaser}
\end{figure}

Several types of sensor technologies can be employed for spacecraft pose estimation, but they are all subject to size-weight-power and cost (SWaP-C) constraints. Optical sensors such as RGB imaging sensors are favored due to their low SWaP-C requirements, high resolution and the availability of established vision-based algorithms. However, operating in the space environment can present nontrivial challenges to vision-based systems. Chiefly, harsh lighting conditions caused by strong sunlight at certain angles to the spacecraft can severely affect the operation of RGB cameras. The first row of Fig.~\ref{fig:teaser} illustrates the effects of harsh lighting encountered in the docking operations of Soyuz~\cite{soyuzdockingteaser} and the International Space Station~\cite{issdockingteaser1, issdockingteaser2}. More specifically, harsh lighting can manifest as:
\begin{itemize}
    \item Glare: Strong incident light on the sensor causing occlusion of part or whole of the target object. The stronger the glare is the higher the possibility of over-exposure, where the pixel values are clipped to the maximum of the intensity range. Glare is exhibited in all the examples in the first row of Fig.~\ref{fig:teaser}.
    \item Blooming: Soft radial glow on the edges of the object. See Fig.~\ref{fig:teaser}, row 1, column 2.
    \item Lens flare: Partial occlusion due to the optics creating a dispersion of colors in a circular shape or resembling stars or streaks. See top-left quadrant of Fig.~\ref{fig:teaser}, row 2, column 1.
\end{itemize}
The effects of harsh lighting make it challenging or impossible to recover the full structure and texture of the target object through basic image processing techniques such as rescaling the image brightness or equalizing the contrast. Although the periods of harsh lighting can be temporary during on-orbit operations, a temporary loss of accurate relative positioning data during a critical phase can be catastrophic.

Recently, deep learning has been actively adopted to conduct vision-based pose estimation, whereby deep neural networks (DNNs) are used to predict the pose of the object in an input image~\cite{sharma2018pose, kisantal2020satellite, park2024robust}. In general, to perform effectively, DNNs need to be trained on large amounts of labeled data that are representative of the operating environment. Given the limited availability of real data from actual in-orbit operations, researchers have resorted to training DNNs on synthetically generated data from simulation environments such as OpenGL~\cite{Park_2022} and Unreal Engine~\cite{ursoesaunreal}. However, DNNs trained on rendered data suffer from the \emph{domain gap} problem when the synthetic data does not match the real-world conditions. A major cause of domain gap  is harsh lighting~\cite{kelvinsspec21, park2023adaptiveshirt, dixon2025uncertaintyevidental}, whose effects are difficult to simulate in software due to the need for precise knowledge of lighting, optics and material properties~\cite{bao2021rendering,delavennat2021physically,chen2021self,sun2021end}. In short, the adoption of deep learning compounds the problem of harsh lighting in spacecraft pose estimation.

Mitigating the domain gap via domain adaptation (DA) techniques~\cite{csurka2017comprehensive, Peng_2018_CVPR_Workshops} has demonstrated promising results. Broadly, DA aims to refine a pretrained DNN on unlabeled or partially-labeled test-time data. Two popular domain adaptation paradigms are unsupervised domain adaptation (UDA) and test-time adaptation (TTA). UDA reorients the feature map of the DNN such that the common characteristics of the training and testing domains that are useful for the task (\eg, geometric structure of the target object) are accentuated, while distracting differences (\eg, color and lighting variations) are attenuated~\cite{kelvinsspec21results, wang2023bridging,kelvinsspec21}. However, UDA requires numerous (\eg, thousands) testing samples and several training iterations, hence it is not suitable for online operation. TTA strives for the same goal as UDA but performs the update online as the data becomes available at test-time~\cite{jawaid2024iros, park2024robust}. However, TTA can be computationally costly, and its efficacy is generally not close to that of UDA methods.

Another approach to deal with harsh lighting is employing sensors that are intrinsically more robust to the deleterious effects of harsh lighting. Event or neuromorphic sensors have much higher dynamic ranges than RGB sensors, \eg, 120 dB versus 50 dB~\cite{eventsensor128, eventsensor640}, which enable the former to distinguish minute intensity differences in a scene with huge brightness range, thereby perceiving details that would otherwise be occluded by harsh lighting effects in the latter; \cf~rows 2 and 3 of Fig.~\ref{fig:teaser}. Previous works on event-based spacecraft pose estimation have demonstrated higher robustness under harsh lighting without the need for DA~\cite{jawaid2023towards, jawaid2024iros}. However, event sensors have a fundamental weakness: since their pixels operate asynchronously to detect intensity changes, the signal-to-noise ratio (SNR) of event sensors is inherently low during periods of low relative motion, which negatively affects pose estimation accuracy; see Fig.~\ref{fig:datasamples} for examples of the impact of low relative motion on the SNR of event sensor outputs.

A fusion technique that combines the relative strengths of RGB and event data for spacecraft pose estimation is highly desirable. Previous event-based pose estimation works~\cite{jawaid2023towards, jawaid2024iros, rathinam2024spades} have not explored such a fusion. The recent work by Yishi~\etal~\cite{yishi2025crossast} proposed a fusion method based on cross-modal self-attention, however, their method has only been demonstrated on synthetic RGB and event data. Moreover, sensor fusion via the attention mechanism, which involves a significant number of learnable parameters, makes the fusion itself susceptible to overfitting on the synthetic data~\cite{crossattentionuda,Yuan_2024_WACV}, which somewhat detracts from the aim of bridging the domain gap induced by harsh lighting.

Our key contributions are:
\begin{itemize}
    \item \textbf{Event-RGB fusion for spacecraft pose estimation:} Our novel method consists of two main components:
    \begin{itemize}
        \item A dual-channel sensor rig for spacecraft pose estimation that utilizes a beam splitter to achieve accurate optical and temporal alignment between separate RGB and event sensors; see Fig.~\ref{fig:beamsplitter_setup}.
    
        Compared to event sensors that can simultaneously output both event data and active pixel sensor (APS) frames (\eg, DAVIS346), our capture system does not require compromising on the data quality (\eg, spatial resolution, dynamic range, SNR)  of either channel necessitated by co-locating the sensors at the pixel level. Moreover, our rig provides flexibility of changing the RGB and event sensors post hoc.

        \item An event-RGB fusion algorithm that exploits the alignment by the dual-channel sensor rig to achieve learning-free fusion via the classical RANSAC method~\cite{fischler1981ransac}. Our pipeline also examines geometric consistency of pose predictions across the two channels to detect potential break down of the pipeline, as well as conducts uncertainty estimation in the DNNs to account for the varying reliability of each channel.

    \end{itemize}
    
    Unlike previous techniques~\cite{kelvinsspec21results, wang2023bridging,kelvinsspec21,jawaid2024iros, park2024robust}, our method is trained solely on synthetic data and does not require domain adaptation during test time to operate under harsh lighting conditions. Moreover, it is less susceptible to low SNR situations encountered by event-only methods during low relative motions.

    
    \item \textbf{An optically and temporally aligned event-RGB dataset for spacecraft pose estimation:} To enable hardware-in-the-loop evaluation of the proposed algorithm, we collected real RGB and event data using the dual-channel sensor rig of target satellite mock-ups under real-world harsh lighting conditions. This dataset consists of labeled sequences of event data, event frames and RGB frames capturing satellite-like objects with realistic materials and geometrical structures. See rows 2 and 3 of Fig.~\ref{fig:teaser} for samples of our dataset.

    Our dataset has been publicly released at~\cite{jawaidnoveldataset25} to help stimulate research in this area.
\end{itemize}

\section{Related work}

\subsection{Spacecraft pose estimation}

Accurate spacecraft pose estimation is vital for guiding rendezvous maneuvers between spacecraft~\cite{PAULY2023339,pasqualetto2019review}. The low SWaP-C requirements of optical sensors make them attractive for space-borne pose estimation. However, these sensors can be challenged even if the geometry of the target object is known (\eg, via CAD models): the thin atmosphere means nearly no light diffusion, resulting in extreme image contrasts, pitch-black shadows, low SNR and adverse optical effects like glare and blooming.
Moreover, the relative scarcity of space-derived visual data motivates the usage of computer-generated (synthetic) images for training, resulting in the need to deal with domain gap~\cite{PAULY2023339, park2023satellite,wang2023bridging,park2024robust}.

ESA's satellite pose estimation competitions (SPEC)~\cite{kisantal2020satellite, park2023satellite} featuring the SPEED and SPEED+ datasets~\cite{Park_2022} were the first to study purely vision-based approaches for spacecraft pose estimation and are by now the most important benchmark. 
Unlabeled images featuring a mock-up model of the TANGO spacecraft in a lab environment under harsh lighting are provided together with a large number of labeled synthetic images. The task is to develop a system to infer the 6DoF pose of the model from this data. Other datasets inspired by SPEED+ have followed with a focus on generating sequences rather than images or improving the quality of synthetic datasets~\cite{musallam2022cubesat, rathinam2024spades,Gallet_2024_CVPR}.

Considerable work has been proposed looking into spacecraft pose estimation motivated by the SPEED+ data and SPEC~\cite{yu2024comprehensive,renaut2025deep,bechini2025robust, bechini2024robust, dixon2025uncertaintyevidental}. These approaches explore techniques such as UDA, TTA and incorporating measures of uncertainty. The availability of real testing data especially under challenging conditions which present in orbit is crucial for testing and validating such algorithms.

However, for benchmarking both standalone sensor approaches and sensor fusion approaches the SPEED+ dataset has some limitations. The most obvious is that it only consists of standalone RGB images in non-sequential poses. The SHIRT~\cite{park2023adaptiveshirt} dataset aimed to address this limitation by introducing sequences of images but does not contain any sequences with the harsh lighting conditions of the \texttt{sunlamp} dataset of SPEED+ and similarly, only contains RGB data of a single satellite of object of a relatively simplistic geometrical structure. The SPEED+ dataset authors also filtered out samples with severe overexposure and glare in the post-processing stage whereas in the same overexposure scenario a COTS event sensor can still recover the structure of the spacecraft as seen in Fig.~\ref{fig:teaser}

\subsection{Event sensors for space applications}
Event sensors are bio-inspired devices measuring per-pixel brightness differences asynchronously at high temporal rates~\cite{eventsensor128, gallego2020event}. While event sensing has seen significant interest in robotics and artificial intelligence, its application to space environments remains relatively underexplored~\cite{izzo2023neuromorphic}. Existing work has studied applications to star tracking, attitude estimation, navigation in landing scenarios, more reliable visual odometry for planetary robots and debris detection~\cite{chin2019star, ng2022asynchronous, sofiamcleod2022eccv, azzalini2023on, NICHOLAS2023177}.

Event-based spacecraft pose estimation is relatively new~\cite{jawaid2023towards, rathinam2024spades, jawaid2024iros} and as alluded to in Sec.~\ref{sec:intro}, existing works have not considered genuinely harsh lighting. More recent works such as \cite{liu2025stereo} and \cite{malik2025evsat3d} have proposed spacecraft pose estimation via a stereo-event rig and simultaneous 3D reconstruction alongside pose estimation. However, similar to \cite{jawaid2023towards}, they do not consider severe harsh lighting conditions exhibiting glare and bloom which appear as noise surrounding the structure of the satellite. Furthermore, as discussed in Sec.~\ref{sec:intro}, event sensors are prone to a reduction in signal-to-noise ratio during low-motion conditions, which has also not been addressed by these methods.

Datasets specifically for spacecraft pose estimation like SEENIC have been previously proposed by Jawaid~\etal~\cite{jawaid2023towards} which encapsulate multiple lighting scenarios but not under extreme lighting matching the solar irradiance. Moreover, they capture only a single texture-less object. Rathinam~\etal~\cite{rathinam2024spades} also proposed the SPADES dataset consisting of event and RGB data of a single satellite object. However, the sensors to capture this dataset were not optically aligned and the real data was captured with background clutter of a lab environment which does not realistically represent a space scenario. Furthermore, there are no instances of extreme lighting detailed in their work, and as of submitting this work the dataset has not been made publicly available.

\subsection{Sensor fusion for space applications}
Sensor fusion has received attention for space operations due to the limitations posed by purely monocular optical systems. Recent works such as~\cite{singh2023deep, fusionplussatellite2025, tarasiewicz2023multitemporal} have explored multi-sensor fusion for remote sensing. The redundancy offered by multi-instrument or multi-sensor systems has been leveraged by~\cite{galante2016fast} to provide accurate navigation guidance in conditions of partial or full shadowing during eclipse as well as other challenging conditions. The recent works by Napolano~\etal~\cite{napolano2023multi} and Jiang~\etal~\cite{jiang2024uncooperative} outline the importance of multi-sensor and multi-instruments guidance systems for close proximity operations. These works were inspired by earlier works such as ~\cite{zhang2019fusion, peng2018pose, galante2016fast, tzschichholz2015relativefusion} which also aimed to leverage fusion of sensors and instruments for more robust navigation.

Other works have explored 2D object tracking and position estimation relative to asteroids using combinations of infrared (IR) and visible sensors~\cite{irvisible2023, irvisible2024, fusionposition2024}. However, IR sensors, like the one investigated in~\cite{irvisible2023}, have a lower dynamic range of 60dB~\cite{irsensordr} compared to event sensors which have a dynamic range of greater than 120dB~\cite{eventsensor128}. This makes event sensors a more compelling choice in the sensor fusion combination to tackle the challenges produced by harsh lighting.

The fusion of event-based data has been studied for visual odometry where event-streams are fused with conventional RGB images, RGBD images~\cite{weikersdorfer2014event}, inertial measurement units (IMUs)~\cite{mueggler2018continuous} or any combination of those modalities, to estimate ego-motion and structure. For example, Vidal \emph{et~al.}~\cite{vidal2018ultimate} formulate a joint optimization problem containing the reprojection errors of a conventional camera, an event-camera and an inertial error term from an IMU and integrate them using non-linear optimization~\cite{leutenegger2015keyframe}.

The fusion of event-based data with other sensors or instruments for close-proximity operations has been largely unexplored. To the best of our knowledge only Yishi~\etal~\cite{yishi2025crossast} have attempted to study the event-RGB sensor combination for pose estimation in a non-cooperative close-proximity scenario. However, as pointed out in Sec.~\ref{sec:intro}, this work has not considered the domain gap due to overfitting self-attention and cross-attention modules on synthetic data training data. Multiple studies have demonstrated the already large domain gap between simulated and real data especially in harsh lighting conditions~\cite{kelvinsspec21results, jawaid2023towards, park2024robust} and studies exploring attention for fusion have had to add specific regularization modules to mitigate overfitting~\cite{Yuan_2024_WACV} or perform UDA to address the domain gap problem~\cite{crossattentionuda}.

\section{Event-RGB dataset for spacecraft pose estimation}\label{sec:realdata}
In this section we detail the data acquisition and calibration procedures used for simulating the satellite rendezvous data on-ground which is used to evaluate the proposed event-RGB fusion algorithm.

\subsection{Target objects}\label{sec:targetobjects}
\begin{figure}[ht]\centering
\includegraphics[width=0.9\columnwidth]{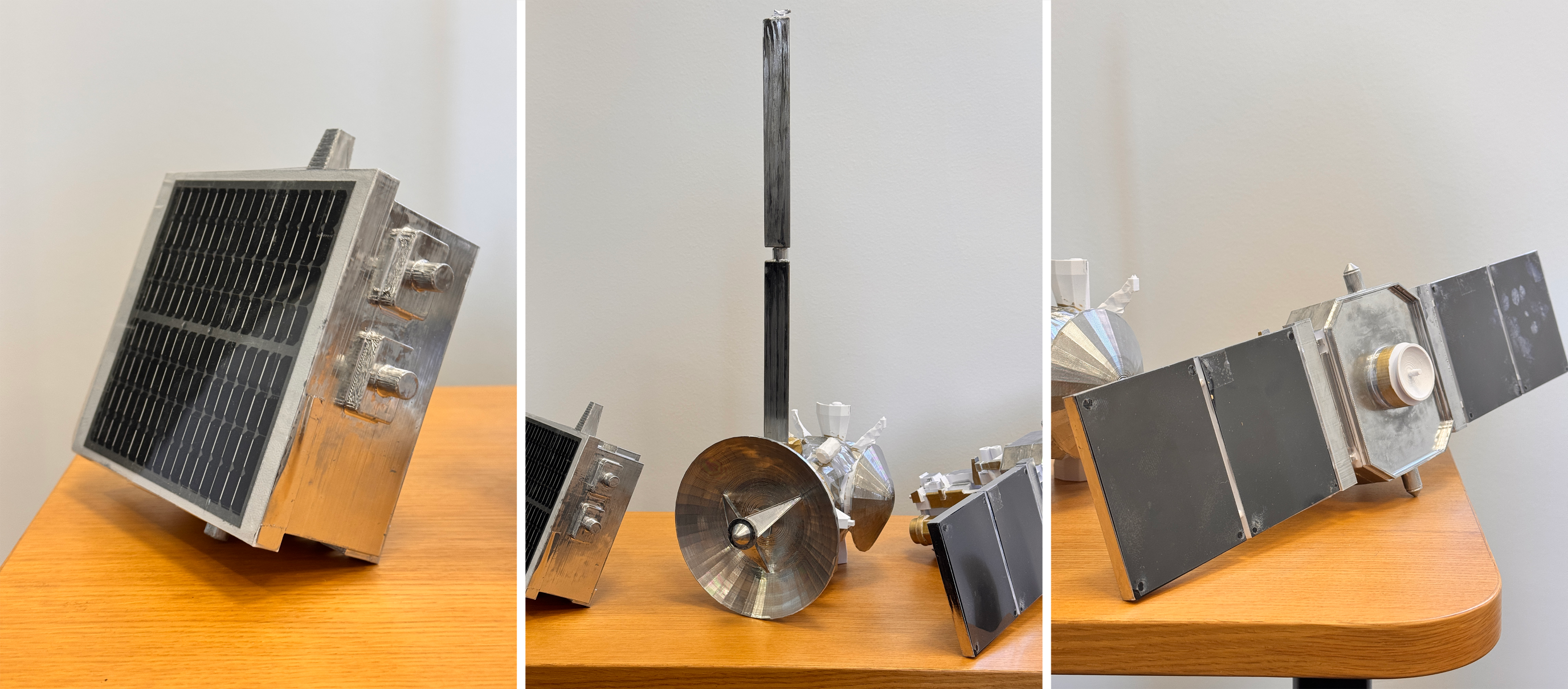}
\caption{Fabricated real models of the target objects: Satty, Cassini and SOHO from left to right.}
\label{fig:targetobjects}
\end{figure}

\begin{figure}[ht]\centering
\includegraphics[width=0.9\columnwidth]{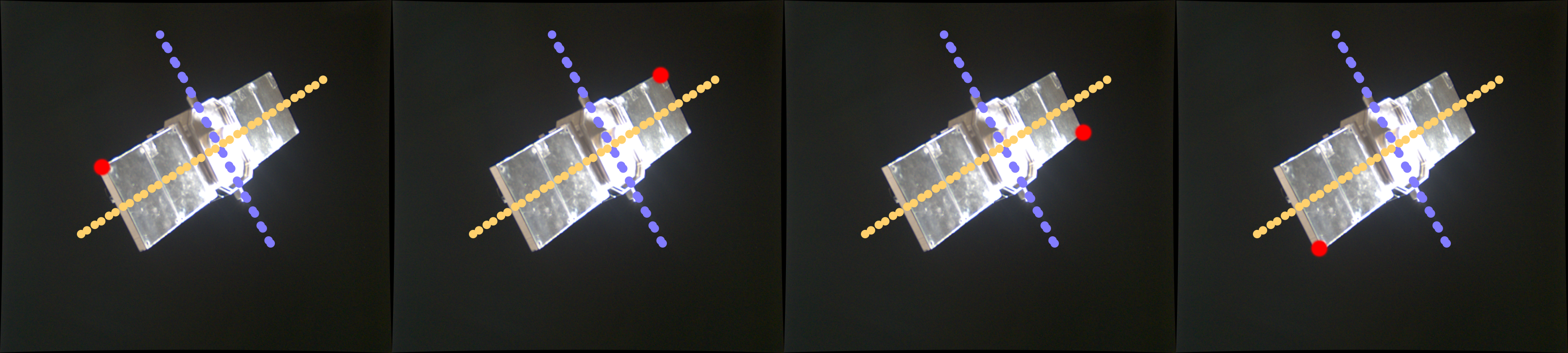}
\caption{Real frame from our dataset showing 4 possible ambiguous locations of the same keypoint in red due to the view of the object. The keypoint can be reflected along two axes of symmetry shown by dotted lines.}
\label{fig:sohosymmetry}
\end{figure}

To facilitate comprehensive testing of the proposed pose estimation approach and to provide a benchmark for future research, a diverse set of satellite models was employed. The selection included publicly available models from NASA's 3D repository~\cite{nasa3drepo}, alongside a custom model referred to as "Satty". The models were chosen to represent a range of surface geometries, material properties and features commonly observed in real-world satellites.
\begin{itemize}
    \item \textbf{Satty:} The custom satellite-like object Satty (approx. dimensions: $18 \times 16 \times 6 cm^{3}$) was 3D-modeled and printed. The fabrication included attaching the solar panel texture printed on paper and layered with glossy plastic film to simulate the reflective properties of a solar panel. Furthermore, liquid chrome paint was applied to simulate highly reflective metallic surfaces on satellites. The design incorporated a combination of flat and angled surfaces to allow for varying angle of incident light. The simplified geometry compared to the other two satellite models in this work offers a controlled baseline of the pose estimation algorithm's behavior.
    \item \textbf{Cassini:} This model was sourced from NASA's public repository of 3D-printable models~\cite{cassinimodel} and 3D printed to approx. dimensions: $45 \times 23 \times 15 cm^{3}$. The model exhibits uneven surfaces and articulated components such as the rotating dish and a long boom pole structure. Similar to SOHO, this model is painted with the liquid chrome and some parts in metallic gold.
    \item \textbf{SOHO:} The Solar and Heliospheric Observatory (SOHO) was also sourced from NASA's 3D public repository~\cite{sohomodel}. The model was 3D printed to approx. dimensions: $44 \times 15 \times 14 cm^{3}$ and possesses a symmetrical structure compared to Satty and Cassini which have more unique structures. As seen in Fig.~\ref{fig:sohosymmetry}, the estimated keypoint position can be ambiguous for a standalone instance. The model also has reflective surfaces and distinct dark panels representing solar arrays. SOHO was painted with the reflective liquid chrome paint as well as covering some surfaces with a comparatively less reflective metallic gold color, providing a more balanced material property paired with the symmetry challenge.
\end{itemize}
See Fig.~\ref{fig:targetobjects} for a close-up view of the target objects under diffused lighting.

\subsection{Capture system}
\begin{figure}[ht]
    \centering
    \begin{subfigure}[b]{.29\columnwidth}
    \centering
        \includegraphics[height=.9\columnwidth]{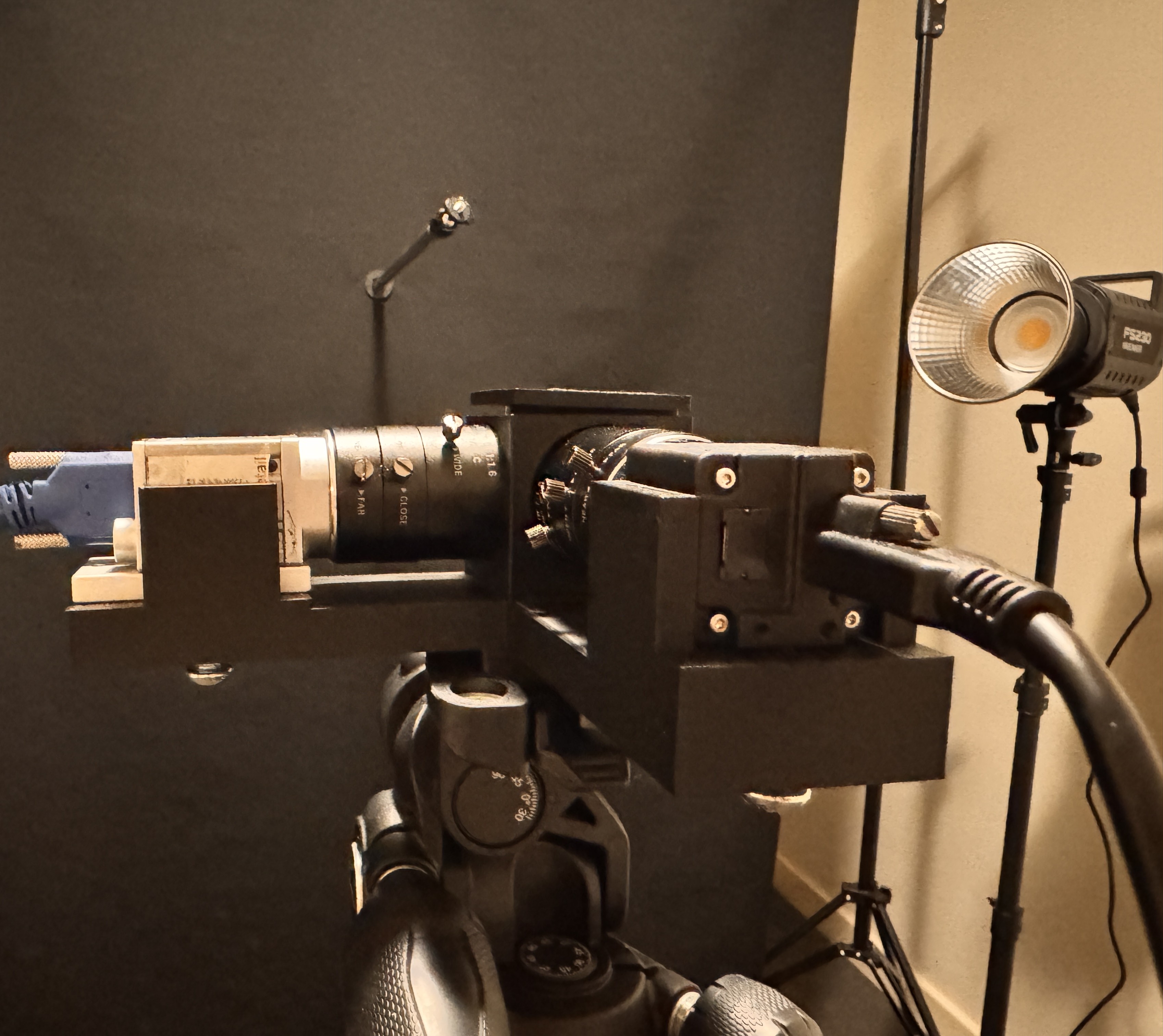}
        \caption{}
    \end{subfigure}
    \begin{subfigure}[b]{.29\columnwidth}
    \centering
    \includegraphics[height=.9\columnwidth]{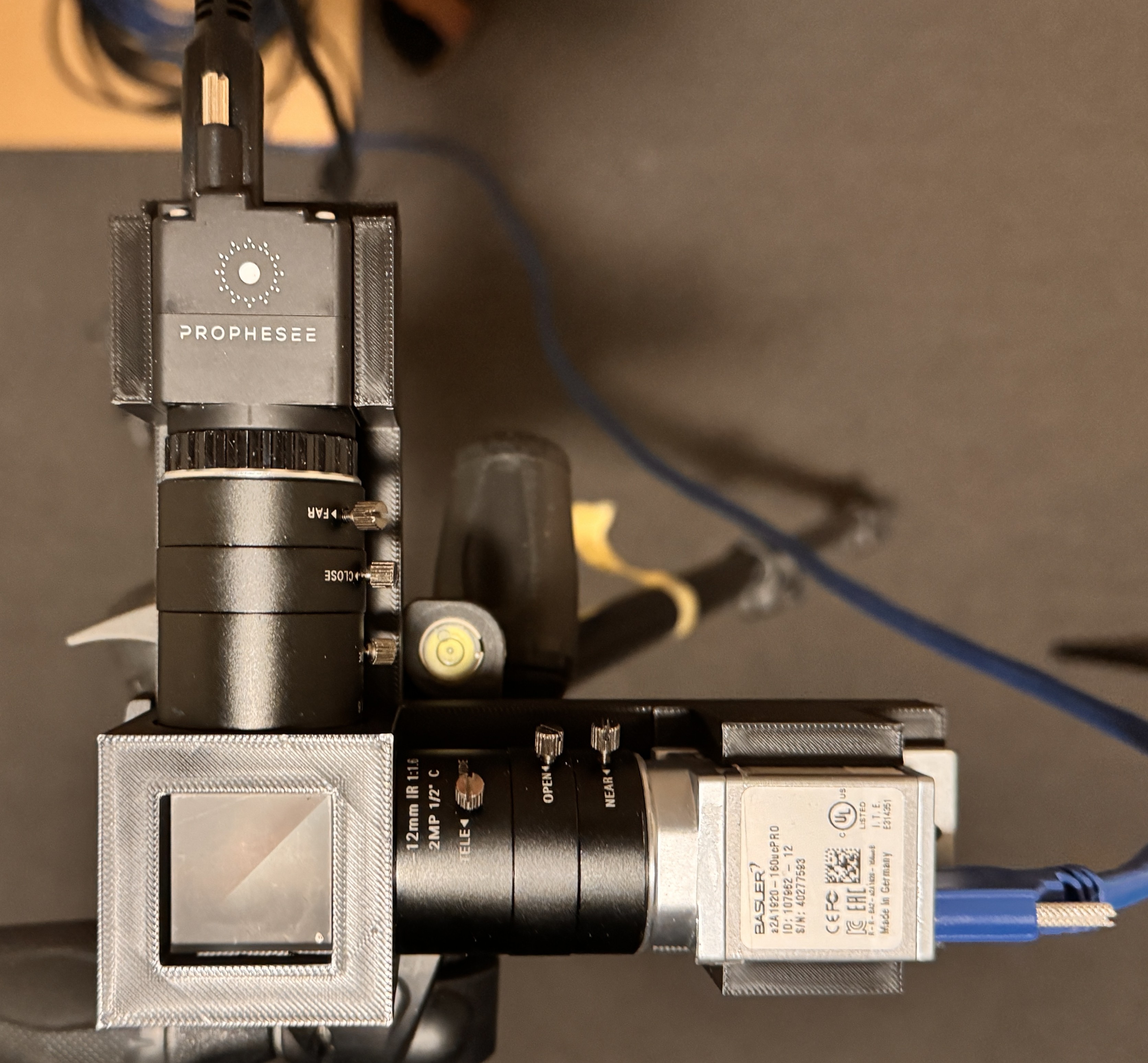}
         \caption{}
     \end{subfigure}
    \begin{subfigure}[b]{.29\columnwidth}
    \centering
    \includegraphics[height=.9\columnwidth]{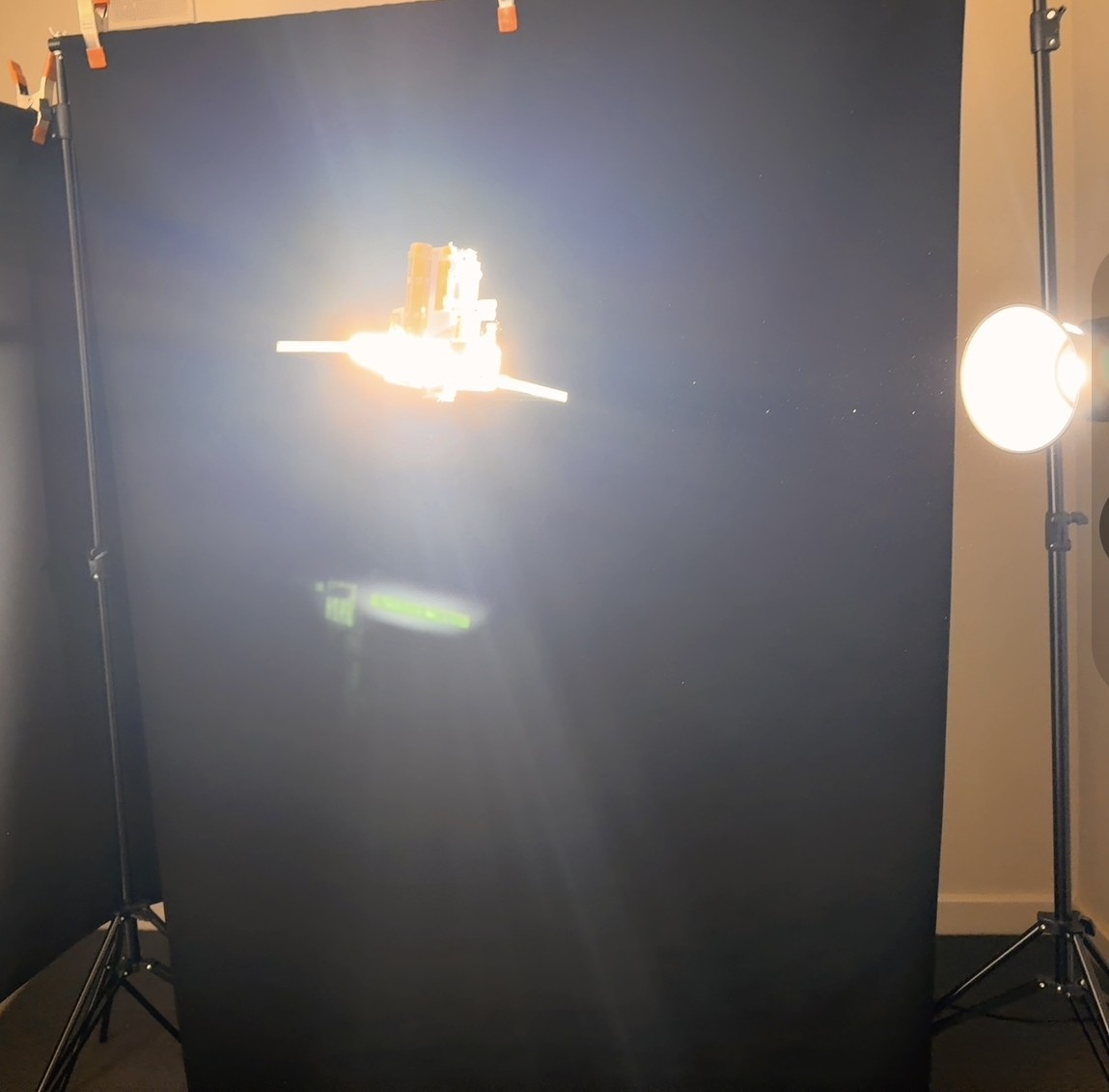}
         \caption{}
     \end{subfigure} 
     \caption{(a) Side view of the event-RGB capture setup. (b) Top-down view of the capture setup. (c) View of the satellite object, backdrop and light source.}
\label{fig:beamsplitter_setup}
\end{figure}

We developed a system incorporating a Basler ace 2 pro RGB camera equipped with a Sony IMX636 color sensor and a Prophesee EVK4 event camera. We considered COTS packages offering simultaneous event and RGB output, like the Inivation DAVIS346, but these packages have drawbacks like: lower spatial resolution for both event and frames, low frame rate for frame output, grayscale-only frame output and lower optical frame dynamic range compared to dedicated popular machine vision RGB cameras like the Basler ace 2 pro. A custom package like ours allows us and future work to fuse but also perform a like-to-like comparison of the best available sensor for multiple modalities without relying on COTS devices from sensor package manufacturers. Similar to works like~\cite{Liu_2024_CVPR, Liang_2023_ICCV}, we deployed a $25 \times 25 \times 25mm^{3}$ beamsplitter glass prism to physically align the viewpoints of the two sensors (See (a) and (b) in Fig.~\ref{fig:beamsplitter_setup}). This, together with a custom-designed and 3D printed rig, fabricated from black PLA filament with tight tolerances to minimize light leakage, centers the two cameras and their respective Arducam 4-12mm focal length lenses. The data from both sensors is cropped around the center of the frame to $800 \times 720$ resolution to remove the reflection of the beamsplitter prism which occurs at the far horizontal edges. This cropping step can be mitigated by using a larger beamsplitter prism. The lenses were initially adjusted to the maximum telephoto focal length and then subsequently fine-tuned on the RGB camera to achieve near-identical viewpoints relative to the event camera, reducing the baseline introduced by the separate lenses. The focus setting was tuned on both cameras using a rotating Siemens star pattern positioned at a distance matching the distance to the 3D satellite models for the data capture. The focal settings are then fixed for the rest of the experimental procedure.

\subsection{Data acquisition}
We captured the data using the default settings for the event camera and adjusted the RGB exposure period to 10,000 microseconds for reducing motion blur. To simulate in-orbit satellite observation conditions, an on-ground environment with controlled object trajectories and lighting was established. Each satellite model was mounted on a tripod ballhead attached to a long boom arm. The opposite end of the arm was affixed to a rotating panhead concealed behind a matte black photography backdrop (Fig.~\ref{fig:beamsplitter_setup}(c)) to minimize scene clutter as well as stray light reflections. The illumination consisted of a Neewer FS230 LED light of a 5600K manufacture set spectral response matching daylight. The light is set to an output of 170,000 lux incident on the satellite surface when $1m$ away (derived using the conversion provided by~\cite{solarconstantconversion}, together with the roughly $1400W/m^2$ solar irradiance specified in the \texttt{sunlamp} dataset of~\cite{Park_2022}). This setting creates challenging harsh lighting and induces glare. The satellite model is also oriented and positioned with the ballhead to produce reflections from the light source. The following parameters were then adjusted to obtain the data sequences:
\begin{itemize}
    \item \textbf{Trajectories:} The satellite was rotated using the panhead, capturing a portion of its trajectory encompassing both glare and non-glare conditions. Three unique orientations were captured using rotational trajectory programmed on the panhead interleaved by a brief pausing and deceleration moment of the satellite occurring after the glare subsides. The capture for the first orientation was repeated for each target object to have the glare and deceleration moment occur at the same time in the trajectory to produce RGB frames and event data which suffers from glare and low-motion together. The trajectories were designed to intentionally incorporate a variety of harsh lighting situations and are not modeled after realistic mission profiles. Although our actuating capabilities are limited, the motions may nonetheless visually approximate certain orbital maneuvers. For instance, low-motion intervals could represent attitude control or detumbling procedures, while steady rotation may simulate a tumbling target. Moreover, our proposed pose estimation method is capable of handling arbitrary motion, as it does not assume any specific underlying motion model.
    \item \textbf{Distances:} The capture for all four trajectories was repeated for two distances \texttt{close} and \texttt{far} which are $\approx=0.8m$ and $\approx1.2m$ respectively for Cassini and SOHO. For Satty, we adjusted \texttt{close} and \texttt{far} to $\approx0.3m$ and $\approx0.7m$ to maintain a similar view of the smaller model in the frame as the other two larger models.
\end{itemize}  
Following the permutations of the above \ie 3 target objects $\times$ 4 trajectories $\times$ 2 distances, we obtained $\ell=\{1,\dots,24\}$ sequences of data.

\subsection{Temporal synchronization}\label{sec:temporalsynchronisation}
The RGB and event data are temporally synchronized by capturing the data in a multi-threaded computer process. We followed a producer-consumer model for the process. The producer thread for the RGB camera operates at a fixed time period, $\Delta{\tau}$ equal to $30$ frames per second \ie producing a frame $I^{rgb}_n$ for every time step $\tau_n = \tau_1 + (\Delta{\tau}\times n)$ and $n\in [1,2,\dots, N_{\ell}]$ (where $N_{\ell}$ is the number of frames in the sequence $\ell$). The RGB frame $I^{rgb}_n$ is added to a global buffer $\cR$ which only contains the most recent RGB frame at all times. Simultaneously, the producer thread for the event data, asynchronously captures event data to the global event buffer $\cE_{\ell}$ which has the form of $\cE_{\ell} = \{ \be_1, \be_2, \dots \}$, where each event $\be_i = \{x_i,y_i,p_i,t_i\}$ is a tuple with image coordinates $(x_i,y_i)$, polarity $p_i \in \{-1,1\}$ and timestamp $t_i$.

The consumer thread also operates at $\Delta{\tau}$ and produces an event frame using event-to-frame conversion~\cite{jawaid2023towards} every $\tau_n$. Specifically, an event frame $I^{event}_n \in [0,\Gamma]^{800 \times 720}$ is generated from an event batch $\cB_n \subset \cE_{\ell}$ corresponding to the time window $W_n = [\tau_n - \Delta{\tau}, \tau_n]$. $\Gamma$ is the highest possible pixel intensity and $I^{event}_n$ is a 2D histogram with $800\times720$ cells of the image coordinates $(x_n,y_n)$ of the events in $\cB_n$. The values of the histogram are then normalized ($\Gamma = 1$).

In this capture process, we initially write the RGB frames, event frames and the full event stream to the fast system memory to reduce read/write cycle overheads and only export this data to the disk when the program exits. Following this routine and initializing the timesteps for all threads according to the system clock, intrinsically ensures temporal synchronization of the data. 

\subsection{Optical alignment}\label{sec:opticalalignment}
\begin{figure}[ht]
    \centering
    \begin{subfigure}[b]{.24\columnwidth}
    \centering
        \includegraphics[height=.9\columnwidth]{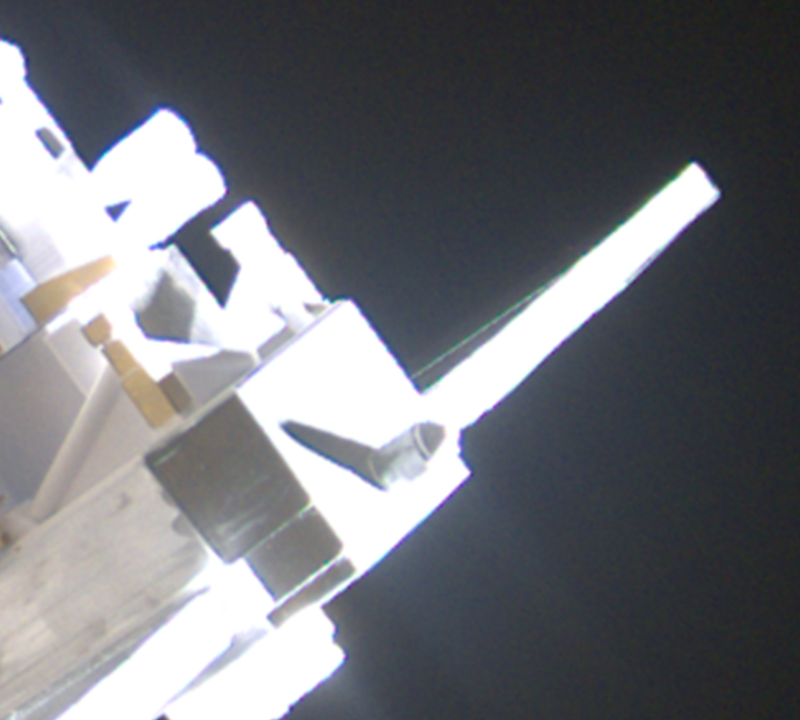}
        \caption{pre-alignment RGB frame}
    \end{subfigure}
    \begin{subfigure}[b]{.24\columnwidth}
    \centering
    \includegraphics[height=.9\columnwidth]{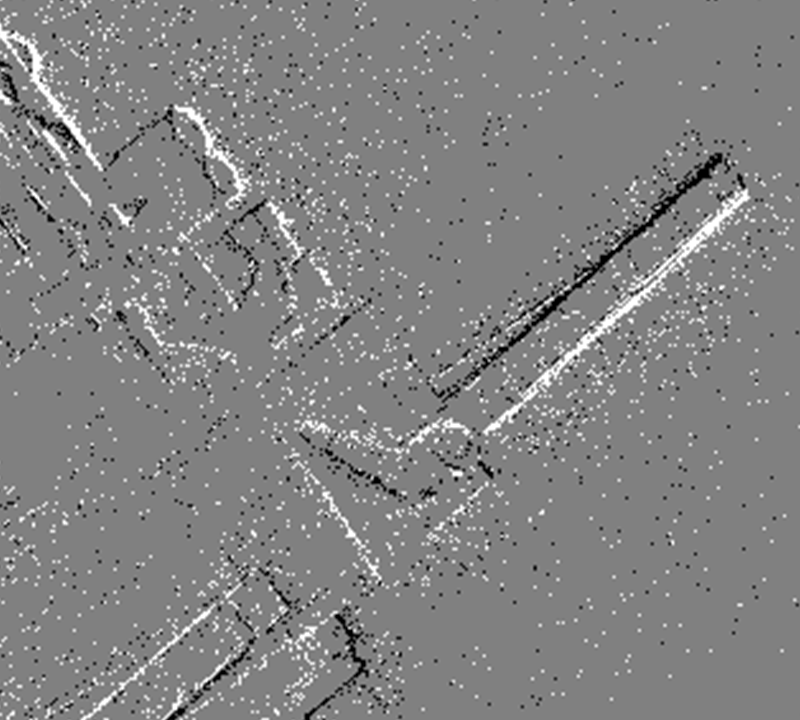}
         \caption{pre-alignment event frame}
     \end{subfigure}
    \begin{subfigure}[b]{.24\columnwidth}
    \centering
    \includegraphics[height=.9\columnwidth]{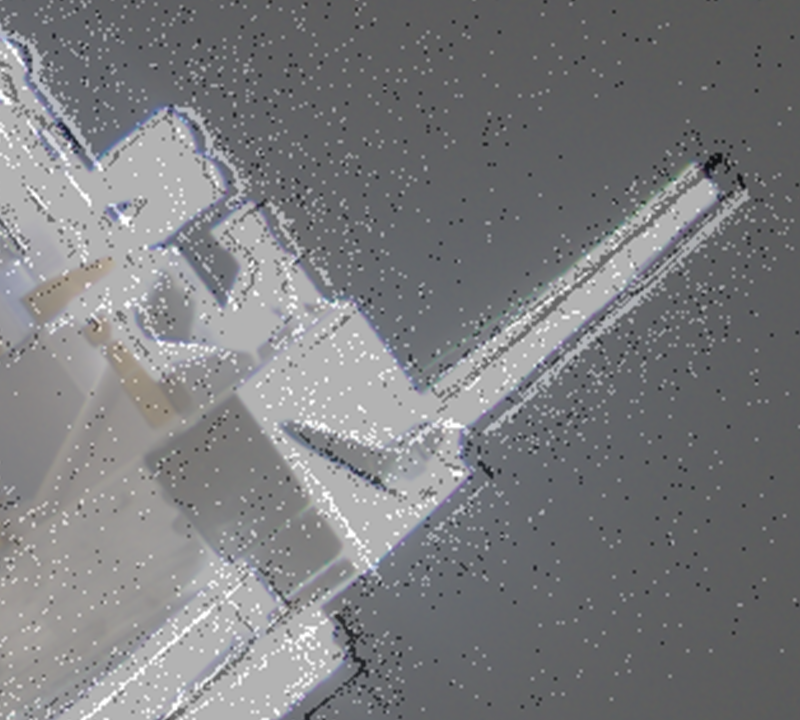}
         \caption{pre-alignment}
     \end{subfigure}
    \begin{subfigure}[b]{.24\columnwidth}
    \centering
    \includegraphics[height=.9\columnwidth]{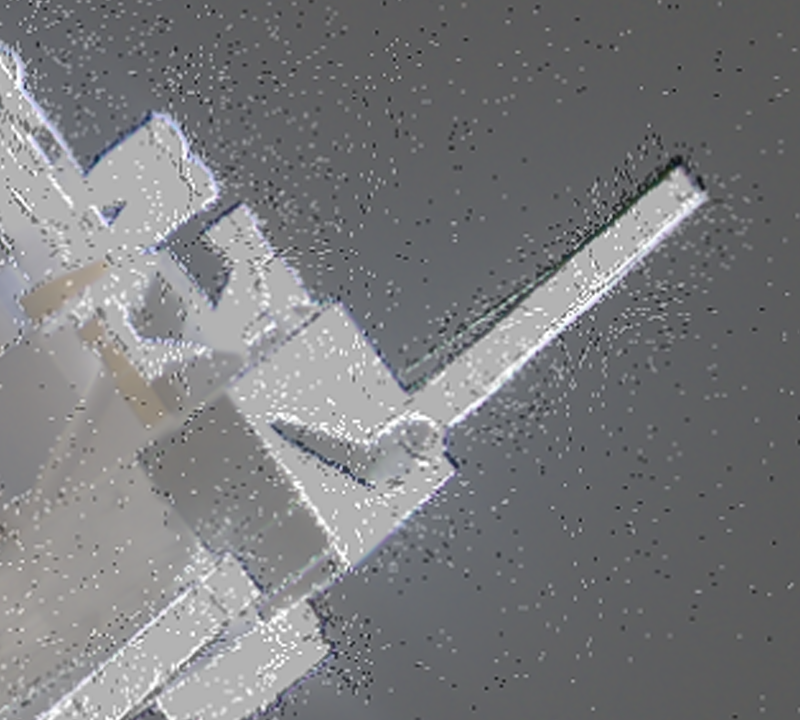}
         \caption{post-alignment}
     \end{subfigure} 
     \caption{Zoomed in view of a pair of misaligned RGB and event frame obtained from the consumer thread outlined in Sec.~\ref{sec:temporalsynchronisation}. (c) shows the two misaligned frame overlaid on top of each other and (d) finally show the overlay following the alignment in Sec.~\ref{sec:opticalalignment}.}
\label{fig:opticalalignment}
\end{figure}

We observed that despite using a beamsplitter the actual viewpoints of the two sensors were still slightly misaligned as seen in Fig.~\ref{fig:opticalalignment}. This misalignment occurs because of the different camera body constructions, physical sensor dimensions and slightly different focal lengths of the individual cameras employed in our capture setup. Therefore, calibration is necessary to reduce the remaining misalignment of the output frames from the two cameras.

We used a checkerboard pattern to calibrate and determine the camera intrinsics matrix and the distortion coefficients as follows. A $11\times6$ checkerboard pattern was displayed on a screen and captured simultaneously with both sensors. To capture an RGB calibration frame the displayed checkerboard was kept static and briefly flickered to obtain events to form an event frame of the same orientation. The calibration RGB frames and event frames were processed separately using the MRPT camera calibration tool~\cite{mrptcalibration} for each sensor to obtain the respective camera matrix and distortion coefficients. Using these parameters, the calibration RGB frames and event frames were undistorted. Following the undistortion, we used the detected checkerboard corners in a single pair of the undistorted RGB and event frames to compute a 2D transformation matrix to align the event view to the RGB view (See Fig.~\ref{fig:opticalalignment} (c) and (d) for a before and after). The transformation matrix is an affine warp with a translation and scale component to account for the different RGB and event sensor positions and sizes. Finally, we translated and scaled the event frames using the computed affine warp (see OpenCV \texttt{warpAffine} implementation for details) to align them to the RGB frame's coordinate space. This process culminated in the paired RGB frames and event frames aligned to the RGB frame space: $\{\bI^{rgb}, \bI^{event}\}$. We performed calibration on the undistorted RGB checkerboard frames to obtain the camera matrix $\mathbf{K}_{real}$ which describes the pinhole camera parameters for both $\bI^{rgb}$ and $\bI^{event}$. This works because the pixels in $\bI^{event}$ are warped to align with the pixels in $\bI^{rgb}$.

\subsection{Data labeling}\label{sec:manuallabeling}
Following Sec.~\ref{sec:opticalalignment}, we derived the overlaid frames sequence $\bI^{overlay}_{\ell}=\{0.5\times\bI^{rgb}_{\ell,n}+0.5\times\bI^{event}_{\ell,n}\}^{N}_{n=1}$ to aid in labeling during the harsh and low-motion conditions. We imported the frame sequence $\bI^{overlay}_{\ell}$ into Blender and aligned the CAD model to the target satellite object within the frame at a small number of keyframes within the frame timeline. Blender's interpolation capabilities were used to estimate the translation and rotation across the unlabeled frames. The event frame component of the overlay provides knowledge of the satellite structure to the labeler during harsh-lighting. Similarly, the structure of the satellite can be inferred from the RGB component of the overlay during low-motion event frames. During periods where there are both harsh lighting and low-motion, the pose change is not drastic, therefore, interpolating between event frames with motion just before and just after the low-motion period yielded poses in the conditions which challenge both sensors. For each frame, the corresponding pose label was exported from Blender. This culminated in obtaining the labeled real sequence of aligned RGB and event frames as well as the following key ground-truth metadata for each frame $\{\bI^{rgb}_{\ell,n}, \bI^{event}_{\ell,n}\}^{N}_{n=1}$:
\begin{itemize}
  \item 6DoF pose label $(\mathbf{R}_{\ell,n},\bt_{\ell,n}) \in SE(3)$, of the target satellite's CAD model specified in the frame of the chaser camera, where $\mathbf{R}_{\ell,n} \in SO(3)$ and $\bt_{\ell,n} \in \mathbb{R}^3$ are respectively the rotation and translation of the object.
  \item 2D locations $\cM_{\ell,n}$ of the 3D landmarks $\cL$ projected using the camera pose.
  \item Bounding box $B_{\ell,n}$ derived using the maximum and minimum values of $\cM_{\ell,n}$. The width and height of the bounding boxes are expanded by 10\% to match the bounding boxes of the synthetic data. Refer to Sec.~\ref{sec:syntheticdata} for details of adding this tolerance to the bounding boxes.
\end{itemize}
Fig.~\ref{fig:datasamples} demonstrates the accuracy of the pose labeling. In each sample frame of the figure, the wireframe of the satellite CAD model is reprojected and plotted on the frame using the labeled pose $\{\mathbf{R}_{\ell,n},\bt_{\ell,n}\}$. Due to rigor of the pose labeling process, the wireframe aligns with the object in each frame and is also consistent sequentially. The specific error introduced by this labeling cannot be precisely determined due to the unavailability of ground-truth poses from an additional source such as a motion capture system. However, there is strong indication that the error is unlikely to exceed approximately $3mm$ and $3^{\circ}$: Appendix~\ref{sec:labelingerrors} provides an analysis of the labeling accuracy of our technique by inferring the error ranges that can be expected based on a dataset with verified ground-truth poses.

\subsection{Dataset availability, statistics and samples}
Our labeled dataset dubbed FRESH (Fusion with RGB and Events for Spacecraft pose estimation under Harsh lighting) has been publicly released at~\cite{jawaidnoveldataset25}. 

\begin{figure}[ht]\centering
\includegraphics[width=0.95\columnwidth]{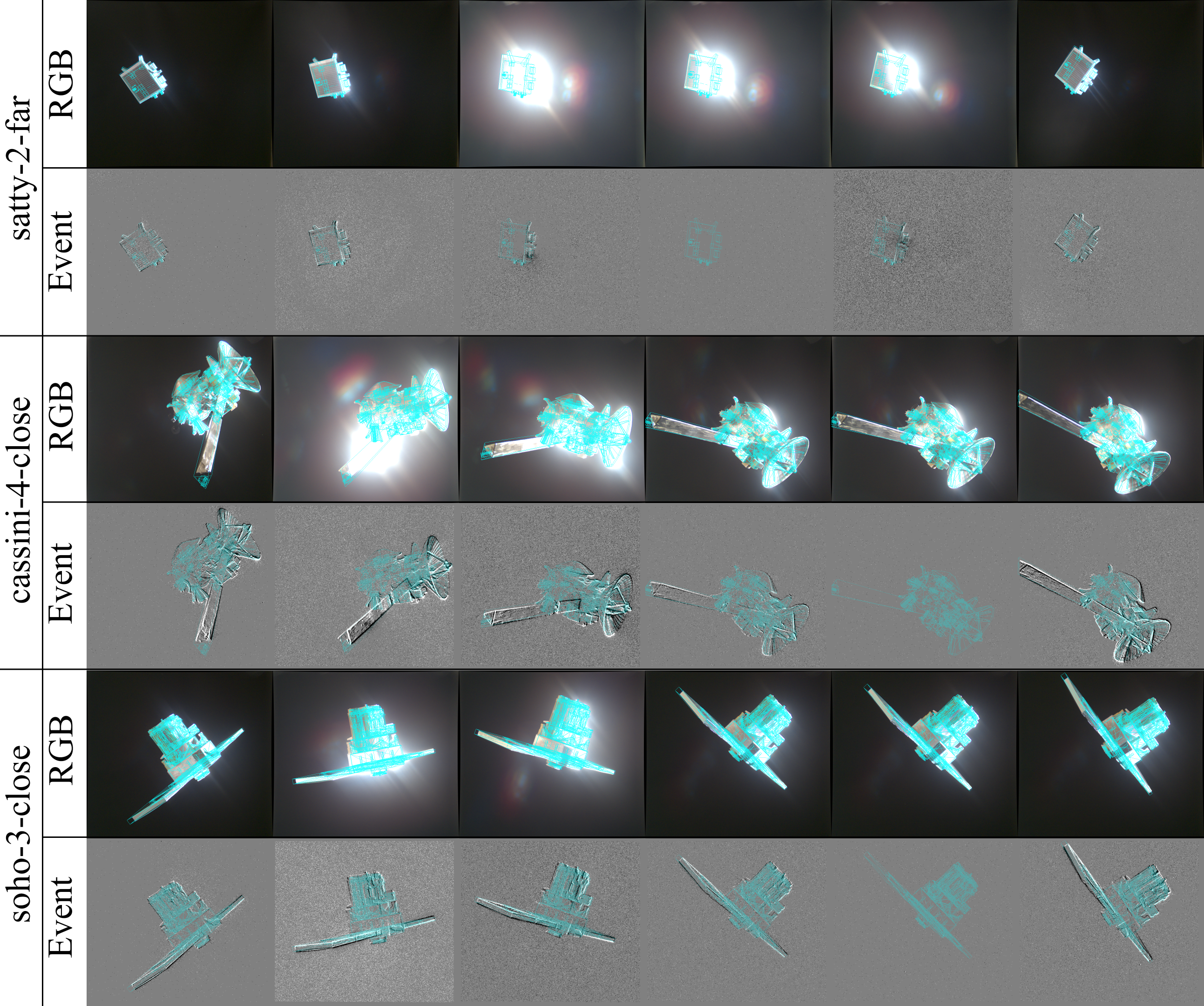}
\caption{Samples from the real dataset. Each set of two rows from top to bottom corresponds to a sequence $\ell$ shown across time from left to right. The CAD wireframe of each satellite is projected using the ground-truth pose and overlaid in cyan.}
\label{fig:datasamples}
\end{figure}

The data sequences are named in the format: \texttt{<object>-<trajectory\_index>-<distance>} where:
\begin{itemize}
    \item $\texttt{object}\in\{satty, cassini, soho\}$,
    \item $\texttt{trajectory\_index}\in\{1,2,3,4\}$,
    \item $\texttt{distance}\in\{close,far\}$
\end{itemize}
The event count in each sequence range from $\approx20\;Mil.$ events to $\approx60\;Mil.$. The frame count ranges from $\approx250$ to $\approx450$ where roughly one-third of the frames are affected by harsh lighting and roughly another third of the frames affected by low-motion. For $trajectory\_index=2$ the harsh lighting period and low-motion period overlap whereas for all other sequences the harsh lighting period is followed by the low-motion period. Frame ranges for these periods are included in the metadata to aid in reproducing our results. In the results section~\ref{sec:results}, we plot these regions on the error charts. See Fig.~\ref{fig:datasamples} for qualitative samples of the data.

\section{Event-RGB fusion for spacecraft pose estimation}\label{sec:ourmethod}
For safe close proximity, rendezvous and on-orbit servicing operations we need to know the 6DoF pose of the target object with respect to the chaser to maneuver and align with the docking port for example. Assuming that the CAD model of the target object is known, we propose using an RGB and event sensor package on-board the chaser as part of the visual-based navigation system (VBS) and built our method around these two input channels.
To emulate operational conditions, our method to estimate the pose is entirely trained on synthetic RGB and event data. The algorithm accesses the real data solely to estimate poses during test-time and never for training. Our proposed method starts by performing object detection on the event-frame (see Sec.~\ref{sec:objectdetectionresults} for performance comparison RGB and event-based object detection) and obtaining a bounding box to isolate the target object within the captured frames. Since the RGB and event channels are aligned after Sec.~\ref{sec:opticalalignment}, we use the same bounding for the RGB channel in the subsequent steps. For both channels, we then estimate the 2D positions (keypoints) corresponding to pre-selected 3D positions on the surface of the target object's CAD model. This keypoint estimation is performed within the boundaries of initially detected bounding box. Finally, our method fuses the keypoint estimations from the RGB and event channels to output one combined pose result for both channels to leverage the strengths of both sensors. More details are given in the following sections.

\subsection{Proposed pipeline}\label{sec:basepipeline}
\begin{figure}[H]\centering
\includegraphics[width=0.9\columnwidth]{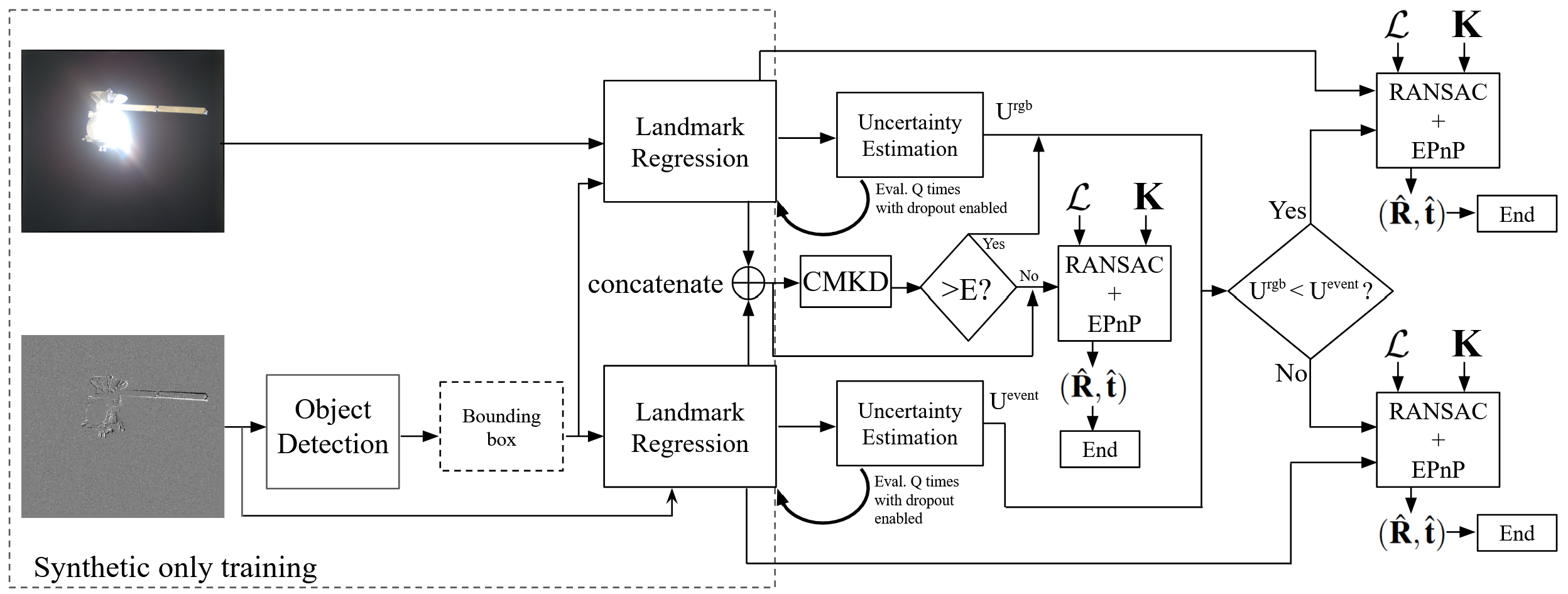}
\caption{Our proposed event-RGB fusion pipeline. Dotted box indicates section trained only on synthetically generated data. Refer to Sec.~\ref{sec:ourmethod} for implementation details of specific modules \eg CMKD.}
\label{fig:pipeline}
\end{figure}

Our method takes a pair of RGB and event frames, $\bI^{rgb}_{\ell,n}$ and $\bI^{event}_{\ell,n}$ as input and outputs a single pose $(\hat{\mathbf{R}}_{\ell,n},\hat{\bt}_{\ell,n}) \in SE(3)$ of the target satellite object in the frame of the chaser's camera rig. The pipeline is divided into two parts: a learnable portion consisting of DNN object detection and landmark regression models which are trained on synthetically generated data, and a non-learnable test-time fusion routine. Fig.~\ref{fig:pipeline} shows the learnable portion in a dotted box and the test-time fusion outside the dotted box.

For both channels, we used the object detector from the event source which outputs a 2D bounding box $\hat{B}_{\ell,n}$ of the satellite in $\bI^{rgb}_{\ell,n}$ and $\bI^{event}_{\ell,n}$ (results shown in Sec.~\ref{sec:objectdetectionresults} justify this approach). Subsequently, $\bI^{rgb}_{\ell,n}$ and $\bI^{event}_{\ell,n}$ are cropped, resized and padded to a resolution of $512\times512$ centered around the center of the respective bounding box and then input to the landmark regressor which once again is separate for each modality. Following this, the landmark regressor for the RGB channel computes heatmaps of 2D spatial coordinates within the $4\times4$ downsampled input space of $Z$ pre-selected 3D landmark positions $\cL$ on the CAD model of the object. The event channel's landmark regressor simultaneously computes the heatmaps for the event channel. Finally, the heatmaps from each channel are separately discretized to yield 2D coordinates $\hat{\cM}^{rgb}_{\ell,n}$ and $\hat{\cM}^{event}_{\ell,n}$ respectively. For implementation details of the single channel pipeline, we refer the reader to~\cite{bospec2019}.

\subsection{Synthetic training data}\label{sec:syntheticdata}
Labeled synthetic datasets were created to train the learnable portion of the method (dotted box in Fig.~\ref{fig:pipeline}). The rendering process was conducted within a Blender~\cite{blender} environment, where the digital camera of the chaser viewing the target satellite object is configured to match the calibrated camera intrinsic matrix $\mathbf{K}_{real}$ corresponding to the real-world RGB camera setup used during physical data collection. The CAD model of each target object was separately rendered against a black background under a diffused lighting ensuring a mostly uniform illumination across the scene. The simple setup allowed flexibility in creating variations of the frames using data augmentation techniques during training (See Sec.~\ref{sec:dataaugmentations}).

For each satellite model $\cC \in \{satty,cassini,soho\}$, the material and texture properties were selected to replicate, as closely as possible, the appearance of the hand-painted, 3D-printed mock-ups (as shown in Fig.~\ref{fig:targetobjects}). Furthermore, we manually marked $Z$ number of 3D salient positions on the surface of the CAD model. These $Z$ points form the set $\cL$ which are the landmark positions used to train the landmark regression module. The manual landmark selection follows the procedure conducted in~\cite{bospec2019}, whereby $Z$ number of points are distributed on the surface of the CAD model to encompass the corners as well pronounced structures such as boom poles and antennae. For Satty and Cassini, we selected $Z=18$ and for SOHO $Z=24$ landmarks to assess the effect of landmark count on method performance.
Then, the chaser camera within Blender is kept static and pointed towards the satellite, which is centered in the frame. The satellite is rotated about the center of volume, along the $x$-axis, incrementally by $1^{\circ}$ per frame, until a $20^{\circ}$ interval is covered. Subsequently, a full $360^{\circ}$ rotation is performed, once again $1^{\circ}$ per frame along the $y$-axis. This process is repeated iteratively for the further $20^{\circ}$ intervals along the $x$-axis, culminating in a compete $360^{\circ}$ rotation. Translations in all axes as well as rotation in the $z$-axis are excluded during rendering, as data augmentations described in Sec.~\ref{sec:dataaugmentations} account for such variations. This routine prevents generating redundant views of the object while ensuring a smooth inter-frame motion to aids in converting frames to synthetic event data. This routine is conducted for two distances \texttt{close} and \texttt{far} representative of the real dataset's distances between the capture system and the satellite model further detailed in Sec.~\ref{sec:realdata}. The process yielded $\cS^{rgb}_{\cC} = \{\cS^{rgb}_{\cC,a}\}^{A}_{a=1}$ frames of $800 \times 720$ resolution, where $A=13,644$ and additionally the key ground-truth metadata is exported for each frame which consists of:
\begin{itemize}
  \item 6DoF pose $(\mathbf{R}^s_{\cC,a},\bt^s_{\cC,a}) \in SE(3)$.
  \item 2D locations $\cM^s_{\cC,a}$ of the 3D landmarks $\cL$ projected using the camera pose.
  \item Bounding box $B^s_{\cC,a}$ derived using $\cM^s_{\cC,a}$ in the same way as the real data bounding boxes. The bounding boxes are expanded by $10\%$ width and height, which ensures the keypoints are not at the exact edge of the tight bounding box and offers better generalization in harsh conditions, as shown by Team TangoUnchained's method in~\cite{kelvinsspec21results}.
\end{itemize}

Finally, the rendered RGB frames are subjected to the V2E~\cite{hu2021v2e} algorithm to generate simulated event data as well as one-to-one paired event frames corresponding to the RGB frames \ie $\cS^{event}_{\cC}=\{\cS^{event}_{\cC,a}\}^A_{a=1}$ and $\cS^{event}_{\cC,a}=V2E(\cS^{rgb}_{\cC,a})$

\begin{figure}[h]\centering
\includegraphics[width=0.8\columnwidth]{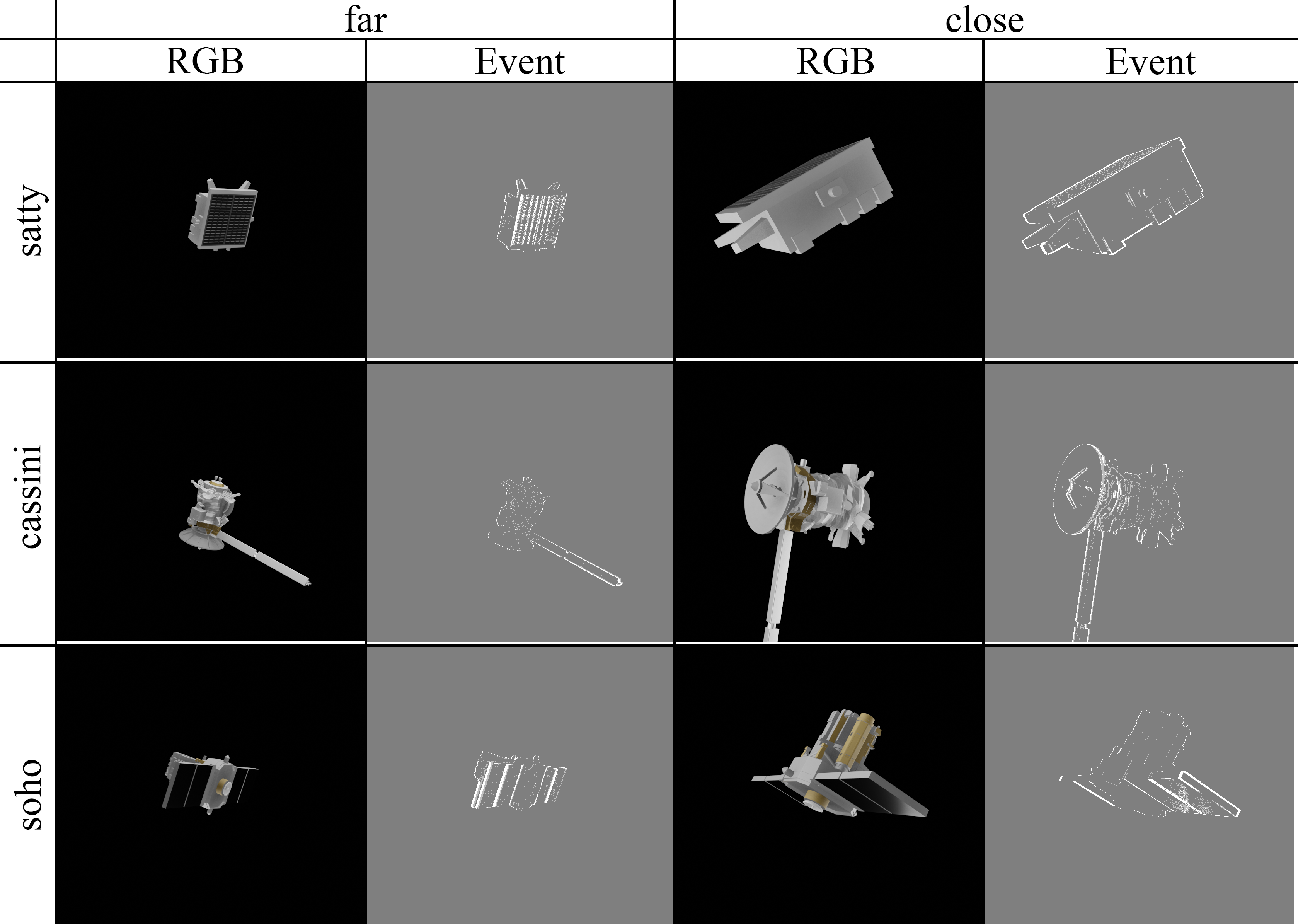}
\caption{Samples from the synthetic dataset.}
\label{fig:syntheticdatasamples}
\end{figure}

Sample synthetic RGB and event frames are visualized in Fig.~\ref{fig:syntheticdatasamples}.

\subsection{Synthetic-only training}\label{sec:synthetictraining}
The object detector which utilizes a \texttt{Faster R-CNN} configuration from \texttt{detectron2}~\cite{wu2019detectron2} for the RGB only methods in Sec.~\ref{sec:results} is trained on $(\cS^{rgb}_{\cC},B^s_{\cC})$ where $B^s_{\cC}=\{B^s_{\cC,a}\}^{A}_{a=1}$. Similarly, the object detector for the event channel is trained on $(\cS^{event}_{\cC},B^s_{\cC})$.

The landmark regressor utilizes an extended \texttt{HRNet}~\cite{wang2020hrnet} backbone with a dropout layer inserted after each activation layer during the encoding phase. The landmark regressor for the RGB channel is trained on $(\cS^{rgb}_{\cC},B^s_{\cC},\cM^s_{\cC})$ where $\cM^s_{\cC}=\{\cM^s_{\cC,a}\}^{A}_{a=1}$. Similarly, the landmark regressor for the event channel is trained on $(\cS^{event}_{\cC},B^s_{\cC},\cM^s_{\cC})$.

\subsubsection{Data augmentations}\label{sec:dataaugmentations}
\begin{figure}[H]
    \centering
    \includegraphics[width=0.95\columnwidth]{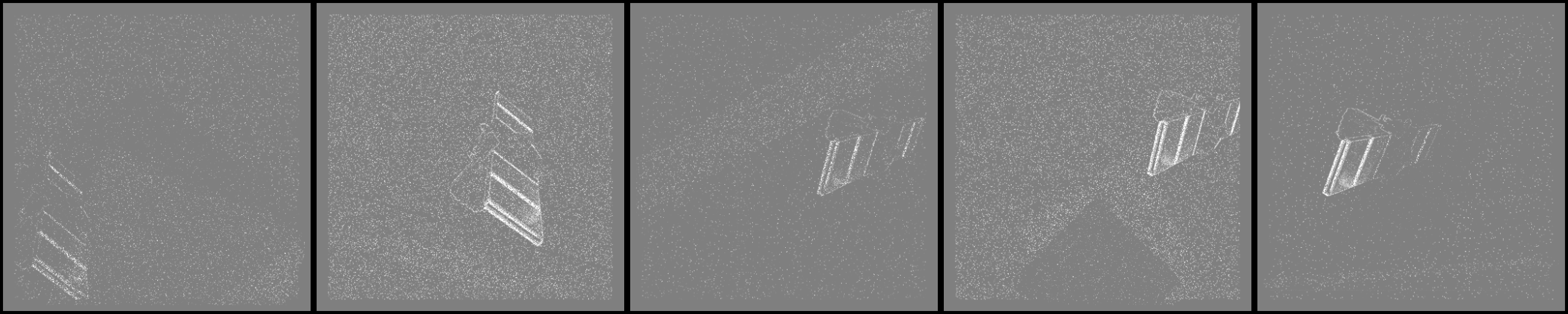}
\caption{A sample of data augmentations applied on the event frames. \texttt{RandomRotation} and \texttt{RandomTranslation} are visualized here as well to demonstrate how we generalized training while reducing data redundancy in the generated synthetic frames.}
\label{fig:eventaugmentations}
\end{figure}

During training of the object detector and landmark regression, we apply a number of data augmentations aid in generalization. We deliberately excluded computer-generated harsh lighting effects during synthetic data rendering as well as the data augmentations. This technique allowed us to quantify and output the uncertainty during unseen conditions, which is used to detect harsh lighting and low-motion in Sec.~\ref{sec:cmkd} and \ref{sec:uncertaintydecision}. \texttt{RandomRotation} and \texttt{RandomTranslation} augmentations are applied to both RGB and event channels to regularize training to positioning of the target object in the frame. Gaussian blur augmentation is also randomly applied to both channels as edges in the synthetic data are sharper compared to the real data. Furthermore, collections of augmentations specific to each channel are also applied to $50\%$ of samples at random. For the RGB channel we apply the following:
\begin{itemize}
    \item \texttt{ColorJitter} from the \texttt{Pytorch} library (with brightness$=0.5$, contrast$=0.75$ and hue$=0.2$)
    \item \texttt{UniformNoise}
    \item \texttt{ColorNoise}
\end{itemize}
And for the event channel we apply:
\begin{itemize}
    \item \texttt{IgnorePolarity} which inverts the black pixels to create a polarity-invariant event frame. This is also applied to the testing images to unify the appearance of the event frame.
    \item \texttt{EventNoise} which adds or removes events according to a Gaussian distribution from the whole event frame.
    \item \texttt{EventPatchNoise} acts like \texttt{EventNoise} but within a randomly generated quadrilateral shape.
\end{itemize}
A sample of these augmentations is shown in Fig.~\ref{fig:eventaugmentations}. The RGB augmentations are trivial and available in popular libraries like \texttt{OpenCV} and \texttt{Pytorch}.

\subsection{Object detection without fusion}\label{sec:objectdetection}
Our testing revealed that object detection trained only on synthetic data with simple data augmentations (refer to Sec.~\ref{sec:dataaugmentations}) is able to detect bounding boxes in the real data to a high degree of accuracy. This was also demonstrated previously by Jawaid~\etal~\cite{jawaid2023towards} on event data. Results in Sec.~\ref{sec:objectdetectionresults} demonstrate the robustness of event-based object detection through the synthetic-to-real domain gap. Though in low or no-motion conditions the performance of the event-based detection drops, it can be accounted for by post-processing bounding-box detections using Alg.~\ref{alg:objectdetectionextension}. Intuitively, Alg.~\ref{alg:objectdetectionextension} performs smoothening of bounding box predictions across time by filtering bounding boxes that drastically change from the last known good prediction (indicated by a >0.98 detection score and a large overlap area with the current bounding box). In this way, the filtered bounding box is replaced with the last known good bounding box. 

\begin{algorithm}
\caption{Event-based object detection extension}\label{alg:objectdetectionextension}
\begin{algorithmic}
\Require $\Theta_{\cC}$ as the object detector trained on $\cS^{event}_{\cC}$ and $B^s_{\cC,a}$ where $\cC$ is the satellite corresponding to the sequence $\ell$
\Require $\bI^{event}=\{\bI^{event}_{\ell,n}\}^{N}_{n=1}$
\State $lastGoodBox,lastGoodScore \gets \Theta_{\cC}\left(\bI^{event}_{\ell,1}\right)$
\State $n \gets 1$
\While{$n \leq N$}
\State $\hat{B}_{\ell,n},score_{\ell,n} \gets \Theta_{\cC}\left(\bI^{event}_{\ell,n}\right)$
\If{$score_{\ell,n} > 0.98$ and $\texttt{IoU}(\hat{B}_{\ell,n}, lastGoodBox) > 0.6$}
    \State $lastGoodBox \gets \hat{B}_{\ell,n}$
    \State $lastGoodScore \gets score_{\ell,n}$
\Else
    \State $\hat{B}_{\ell,n} \gets lastGoodBox$
    \State $score_{\ell,n} \gets lastGoodScore$
\EndIf
\State $n \gets n + 1$
\EndWhile
\end{algorithmic}
\end{algorithm}

\subsection{Test-time fusion techniques}\label{sec:testtimefusion}
As mentioned in Sec.~\ref{sec:intro}, conducting in-situ training or test-time refinement of deep learning models is limited by the SWaP-C requirements of space missions. Therefore, we extended the synthetically trained pipeline in Sec.~\ref{sec:basepipeline} with test-time fusion techniques that do not require further training or learning-based refinement during the mission.

\subsubsection{Cross-modal RANSAC}\label{sec:crossmodalransac}
\begin{figure}[h]
    \centering
    \includegraphics[width=0.9\columnwidth]{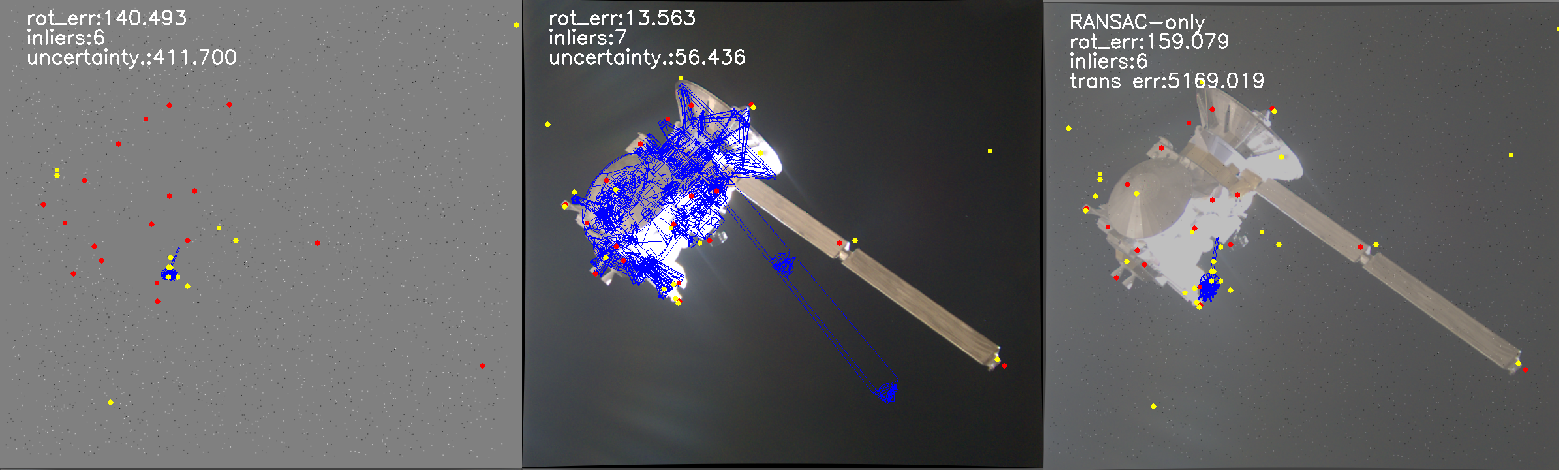}
\caption{An example of the degenerate case where the cross-modal RANSAC favors the points from the event frame (left) as inliers over the RGB observations (middle) to compute the final fused result (right) because more of $\cM^{event}_{n}$ fit the reprojection error threshold. In each image, the wireframe reprojected using the estimated pose is plotted in blue, the yellow dots are the predicted keypoints and red dots are the ground-truth keypoints.}
\label{fig:degeneratecase}
\end{figure}

Established spacecraft pose estimation pipelines utilizing sparse correspondences such as the one proposed by team \texttt{TangoUnchained}~\cite{kelvinsspec21results} typically use a form of Perspective-N-Points(PnP) solver along with RANSAC to filter outlier keypoint predictions. Similarly, other major works including~\cite{wang2023bridging} and~\cite{park2024robust} are two-stage methods. The first stage regresses 2D keypoints (either directly or with intermediate heatmap regression) before solving for the pose using 2D-3D correspondences. The RANSAC algorithm filters out outlier points and maximizes the number of inlier points that fit a given model. We build upon the theoretical foundation of RANSAC in filtering out outlier observations and exploit it for sensor fusion at the keypoint prediction level. The temporal and optical alignment conducted in Sec.~\ref{sec:temporalsynchronisation} and \ref{sec:opticalalignment} allowed us to double the observations which RANSAC can filter for outliers before solving for the final pose using the \texttt{PnP} algorithm (our method uses the EPnP implementation~\cite{lepetit2009epnp}). The input to RANSAC for our method is $\left(\left[\hat{\cM}^{rgb}_{n}, \hat{\cM}^{event}_{n}\right], \left[\cL, \cL\right]\right)$ where $[]$ is the concatenation operator. The output of RANSAC is the inlier points from both sources $\hat{\cM}^{'}_{n}$ and the corresponding 3D landmarks $\cL^{'}$ which then serves as the input to \texttt{PnP} for the final pose output. While RANSAC offers robustness against outlier points across the two channels, it can still be perturbed by false-positives in some edge cases where a large number of outlier points are in a cluster. An example of this is shown in Fig.~\ref{fig:degeneratecase}. In order to account for such edge cases, we developed a heuristic approach to detecting when predicted keypoints from the two channels are not in agreement and then selecting the channel, which would result in a better pose estimation. We detail in Sec.~\ref{sec:cmkd} and \ref{sec:uncertaintydecision} how we conduct this decision-making.

\subsubsection{Cross-modal keypoint distance}\label{sec:cmkd}
We developed a cross-modal keypoint distance (CMKD) criterion to decide if we have an edge case where the two channels differ significantly in their keypoint estimates. As discussed in Sec.~\ref{sec:opticalalignment} and Sec.~\ref{sec:temporalsynchronisation}, we have achieved alignment of the RGB and event channels. If the output 2D keypoints from the two channels, $\hat{\cM}^{rgb}_{n}$ and $\hat{\cM}^{event}_{n}$ are consistent, then both sources can be used with RANSAC. However, if $\hat{\cM}^{rgb}_{n}$ and $\hat{\cM}^{event}_{n}$ are inconsistent, then we can assume that at least one of the channels will not be able to contribute to an optimal pose output. With this reasoning, we can derive this consistency criterion of CMKD as follows:
\begin{equation}\label{eq:cmkd}
CMKD = \texttt{median}\left(\left\{\left|\hat{\cM}^{rgb}_{n,z} - \hat{\cM}^{event}_{n,z}\right|_{2}\right\}^{Z}_{z=1}\right)
\end{equation}

If the CMKD is greater than a threshold $E_{\ell,n}$, we forego the cross-modal RANSAC and select the channel with lower uncertainty as described in the following section (\ref{sec:uncertaintydecision}). For each testing instance $n$, we computed $E_{\ell,n}=\alpha\times |diagonal(B_{\ell,n})|_2$. This is because the keypoints are always predicted within the object detection bounding box. Therefore, the highest possible value of the CMKD is the length of the diagonal of $B_{\ell,n}$. We can then infer that when the two sources are not in agreement, the CMKD will be greater than a percentage of $|diagonal(B_{\ell,n})|_2$. We use the scaling hyperparameter $\alpha$ to decide this percentage. To determine $\alpha$, we used the synthetic data which has been generated based on similar chaser to target distances. Given the synthetic 2D keypoint labels $\{\cM^s_{\cC,a}\}^{A}_{a=1}$ we calculate $\alpha_{\cC}$ for one satellite model as
\begin{equation*}
\alpha^{*}_{\cC}=\sum^{A}_{a=1}max\left(\left\{\frac{\left|NearestNeighbor(\cM_{\cC,a,z}, \cM^s_{\cC,a})-\cM_{\cC,a,z}\right|_2}{|diagonal(B^s_{\cC,a})|_2}\right\}^{Z}_{z=1}\right)~~~,~~~~\alpha_{\cC}=\frac{\alpha^{*}_{\cC}}{A}
\end{equation*}
Where $NearestNeighbor(\cM_{\cC,a,z}, \cM^s_{\cC,a})$ is a function to retrieve the nearest-neighbor of a 2D keypoint $\cM_{\cC,a,z}\in\cM^s_{\cC,a}$ among the other 2D keypoints in $\cM^s_{\cC,a}$. Since the satellites are at similar distances from the sensor rig, we determined $\alpha$ to be 0.2, which is the mean of the $\alpha_{\cC}$ of the three different satellites. Intuitively, this heuristic determines the threshold at which a point is clearly deviating into the space of another keypoint, as a fraction of the maximal deviation. 

In practice, this $\alpha$ will need to be tuned on synthetic data generated according to the expected relative distances between the chaser and the target. In most cases, these distances are determined in the mission planning phase and can be simulated in mission design software such as~\cite{muralidharan2022autonomous}.

\subsubsection{Uncertainty-based decision-making}\label{sec:uncertaintydecision}
Once we have decided that the two sources are not in agreement using the CMKD, we can proceed to select the more reasonable source based on an uncertainty measurement. At test-time, the dropout layers in the landmark regressor can be used to approximate evaluating an ensemble of models~\cite{gal2016dropout} trained on the same data, but initialized differently. These layers can be enabled at test-time and evaluated $Q$ times to produce $\hat{\cM}^{rgb}_{\ell,n,Q}=\{\hat{\cM}^{rgb}_{\ell,n,q}\}^Q_{q=1}$ and $\hat{\cM}^{event}_{\ell,n,Q}=\{\hat{\cM}^{event}_{\ell,n,q}\}^Q_{q=1}$. If the source input is reliable, then the $Q$ samples will be similar and thus produce a low uncertainty measurement. Given that $Q$ is a large enough sample size for the Central Limit Theorem to apply~\cite{islam2018samplesize},we computed the uncertainty measurement $U^{source}_{\ell,n}\;\forall\;source \in\{rgb, event\}$ as follows:
\begin{equation}
U^{source}_{\ell,n,ord} = stdev(\hat{\cM}^{source}_{\ell,n,Q,ord})\times3 \;\forall\; ord \in \{x, y\}   
\end{equation}
\begin{equation}
U^{source}_{\ell,n} = \frac{U^{source}_{\ell,n,x} + U^{source}_{\ell,n,y}}{2}
\end{equation}
Where $stdev$ indicates the standard deviation. We used 3 standard deviations as the uncertainty measurement as $99.7\%$ of normally distributed points would lie within that boundary. Then if $U^{rgb}_{\ell,n}<U^{event}_{\ell,n}$, we conduct PnP and RANSAC on $\hat{\cM}^{rgb}_{\ell,n}$ and otherwise on $\hat{\cM}^{event}_{\ell,n}$.

\section{Results}\label{sec:results}
\subsection{On the accuracy object detection without fusion}\label{sec:objectdetectionresults}
As previously stated in Sec.~\ref{sec:objectdetection} the results below demonstrate the robust performance of object detection through harsh conditions and low-motion using Alg.~\ref{alg:objectdetectionextension}. To measure the performance, we computed the intersection-over-union (IoU) score $\xi_{\ell,n}$ of the predicted bounding box $\hat{B}_{\ell,n}$ and the ground-truth bounding box $B_{\ell,n}$ $\forall n\in[1,N_{\ell}]$ on all sequences $\ell$. Following this, we computed the mean IoU $\bar{\xi}$ across all sequences $\ell$ as the first metric and also the success rate of the IoU, $\Xi$, similar to the mean average precision score commonly used for object detection tasks~\cite{wu2022iouloss}. An IoU score of greater than 0.5 generally indicates a successful detection~\cite{neuripsiou2021}, therefore, we calculate $\Xi$ according to the number of instances greater than the threshold $\lambda=0.5$ in Eq.~\ref{eq:detectionsuccess}:
\begin{equation}
    \Xi_{\ell} = \frac{\sum_{n=1}^{N_{\ell}} \mathbb{I}\left(\xi_{\ell,n}~>~\lambda\right)}{N_{\ell}}, \;where\; \mathbb{I}(condition)=1\;if\;condition\;is\;true\;else\;0
    \label{eq:detectionsuccess}
\end{equation}
\begin{equation}
    \Xi = \frac{\sum_{\ell=1}^{24} \Xi_{\ell}}{24}
\end{equation}

We quantify these two metrics in Table~\ref{tab:objectdetectionresults} for three variants: RGB-only detector trained on $(\cS^{rgb}_{\cC},B^s_{\cC})$ and tested on $\bI^{rgb}_{\ell}$, event-only detector trained on $(\cS^{event}_{\cC},B^s_{\cC})$ and tested on $\bI^{event}_{\ell}$, and lastly, the proposed method of using event-only with the Alg.~\ref{alg:objectdetectionextension} extension. The quantified results show that the proposed event-only fusion-less method performs the best with $\Xi=0.91$ and is sufficiently able to detect the satellite object in the frame for most instances.
The only case where the algorithmic extension of Alg.~\ref{alg:objectdetectionextension} will suffer is when the low or no motion is at the start of the sequence. In that scenario, the object detection score for the event channel will be low and the chaser can default to $\bI^{rgb}_{\ell}$, which according to the RGB-only column in Table~\ref{tab:objectdetectionresults} still shows a high $\Xi$ of 0.8.
\begin{table}[H]\centering
{\footnotesize
\begin{NiceTabular}{clll}[hvlines,rules/color=[gray]{0.3}]
 & RGB-only & Event-only & Event-only w. Alg~\ref{alg:objectdetectionextension}  \\
$\bar{\xi}$ & 0.63 & 0.65 & 0.73 \\
$\Xi$ & 0.80 & 0.77 & 0.91 \\
\end{NiceTabular}}
\caption{Object detection performance demonstrating the robustness of the event-only detection especially with the Alg.~\ref{alg:objectdetectionextension} extension.}
\label{tab:objectdetectionresults}
\end{table}  

We visualize some of the qualitative results of object detection in Fig.~\ref{fig:qualitativeresults}, which demonstrate the efficacy of the object detection trained and tested as per Sec.~\ref{sec:objectdetectionresults}.

\subsection{Pose Estimation Metrics}\label{sec:successratemetric}
For the $\ell$-th sequence $D_{\ell}$, a pose estimation method produced the predicted object poses $\{(\hat{\mathbf{R}}_{\ell,n},\hat{\bt}_{\ell,n})\}_{n=1}^{N_\ell}$, which we compared the results against the ground-truth object poses $\{(\mathbf{R}_{\ell,n},\bt_{\ell,n})\}_{n=1}^{N_\ell}$ using the following metrics~\cite{park2023satellite}:
\begin{equation}
\omega_{\ell,n} = \left\| \bt_{\ell,n} - \hat{\bt}_{\ell,n} \right\|_2,\;\; and \;\;
\theta_{\ell,n} = 2\arccos \left( | \langle \phi_{\ell,n}, \hat{\phi}_{\ell,n} \rangle | \right),
\label{eq:posemetrics}
\end{equation}
which respectively measure translation error in meters (m) and rotation error in degrees ($^{\circ}$), and $\phi_{\ell,n}$ and $\hat{\phi}_{\ell,n}$ are the quaternion form of $\mathbf{R}_{\ell,n}$ and $\hat{\mathbf{R}}_{\ell,n}$. $|\langle\rangle|$ indicates the absolute inner product. To account for potential inaccuracies introduced by our manual labeling process conducted in~\ref{sec:manuallabeling}, we evaluated the methods using a success rate metric. We define the \emph{success rate} for a method on the $\ell$-th testing sequence as
\begin{equation}
    \Omega_{\ell} = \frac{\sum_{n=1}^{N_{\ell}} \mathbb{I}\left(\omega_{\ell,n}~<~\rho\right)}{N_{\ell}}, \;\; \text{and} \;\; \Theta_{\ell} = \frac{\sum_{n=1}^{N_{\ell}} \mathbb{I}\left(\theta_{\ell,n}~<~\sigma\right)}{N_{\ell}}, \;\; where\; \mathbb{I}(condition)=1\;if\;condition\;is\;true\;else\;0,
\end{equation}
This evaluation tolerates ground-truth inaccuracies up to $\rho$ in translation and $\sigma$ in rotation. We set $\rho=10mm$ and $\sigma=10^{\circ}$ to evaluate the translation and rotation success rates respectively. While a minimal error is desirable for successful operations, even state-of-the-art methods (such as the top-performing approach of the Kelvin's challenge~\cite{kelvinsspec21results}) are not able to achieve these success bounds in about half of the cases (see Table~\ref{tab:successratestable}), despite relying on UDA. 

Sec.~\ref{sec:resultsdiscussion} gives a detailed analysis of the results, including success rates based solely on the periods with adverse effects (harsh lighting, low-motion) (provided in the metadata for the dataset). The subset of data corresponding to the periods of adverse effects is denoted by subscript $\Psi$, \eg, $D_{\ell,\Psi} \subset D_{\ell}$ in Table~\ref{tab:successratestable}.

\subsection{Benchmarks}
We trained the methods below separately for each satellite object $\cC$, including event-only and RGB-only HRNet baselines~\cite{bospec2019}, on (RGB and/or event parts of) the synthetic data $D^S_{\cC} = \{ \cS^{rgb}_{\cC},\cS^{event}_{\cC},B^s_{\cC},\cM^s_{\cC}\}$ and tested them on (RGB and/or event parts of) the real data $D_{\ell} = \{\bI^{rgb}_{\ell}, \bI^{event}_{\ell}\}$ $\forall \ell\in\ell_{\cC}$ relevant to the satellite $\cC$.
\begin{itemize}
    \item RGB-only: HRNet trained on $\{\cS^{rgb}_{\cC}, B^s_{\cC}, \cM^s_{\cC}\}$ and tested on $\{\bI^{rgb}_{\ell}\}\forall\ell\in\ell_{\cC}$.
    \item RGB-UDA: Unsupervised domain adaptation (UDA) method proposed by TangoUnchained~\cite{kelvinsspec21results} using the RGB-only model refined on $\{\cS^{rgb}_{\cC}, B^s_{\cC},\cM^s_{\cC}, \bI^{rgb}_{\ell}\}\forall\ell\in\ell_{\cC}$ and tested on $\{\bI^{rgb}_{\ell}\}\forall\ell\in\ell_{\cC}$.
    \item Event-only: HRNet trained on $\{\cS^{event}_{\cC}, B^s_{\cC}, \cM^s_{\cC}\}$ and tested on $\{\bI^{event}_{\ell}\}\forall\ell\in\ell_{\cC}$.
    \item Event-$\zeta$: HRNet trained on $\{ \cS^{event}_{\cC}, B^s_{\cC},\cM^s_{\cC}\}$ then self-supervised separately for each satellite object $\cC$ using the certification method of \cite{jawaid2024iros} and tested on $\{\bI^{event}_{\ell}\}\forall\ell_{\cC}$.
    \item RANSAC-fusion': Our method proposed in Sec.~\ref{sec:crossmodalransac} with RANSAC-only filtering without any edge-case handling; trained on $D^S_{\cC}$ and tested on $D_{\ell}\;\forall\ell\in\ell_{\cC}$.
    \item RANSAC-fusion: Our method with the CMKD and uncertainty-based edge case handling proposed in Sec.~\ref{sec:cmkd} and Sec.~\ref{sec:uncertaintydecision}; trained on $D^S_{\cC}$ and tested on $D_{\ell}\;\forall\ell\in\ell_{\cC}$.
\end{itemize}
Training was conducted on an Nvidia RTX 3090 24GB GPU till convergence on the relevant (RGB or event) subset $D^S_{\cC}$ to each method and testing was conducted on the same GPU.

\subsection{Hyperparameters}\label{sec:hyperparameters}
We outline the hyperparameters of each tested method. The learning rates were selected according to~\cite{kelvinsspec21results, jawaid2023towards, jawaid2024iros}. The training batch sizes were optimized to maximize GPU utilization. Specific values are as follows:
\begin{itemize}
    \item RGB-only: Trained for 6 epochs with an initial learning rate of 0.001 reduced by a factor of 0.1 at the 4th epoch with a batch size of 20.
    \item RGB-UDA: Batch size of 12 for $\cS^{rgb}_{\cC}$ and batch size of 4 for $\bI^{rgb}_{\ell}$. Refined for 2 epochs (epoch 1 at a learning rate of $0.001$ reduced by a factor of $0.1$ the following epoch) as no performance improvement was observed after those epochs.
    \item Event-only: Trained for 6 epochs with an initial learning rate of 0.001 reduced by a factor of 0.1 at the 4th epoch with a batch size of 20.
    \item Event-$\zeta$: The pre-trained event-only model is self-supervised for 10 epochs with a batch-size of 12. Initial learning rate of $0.0001$ reduced by a factor of 0.1 at the 5th epoch. Events noise filter threshold set to 0.96. The certification threshold was relaxed to 100 and reduced by a factor of 0.975 every epoch to adapt the values indicated by~\cite{jawaid2024iros} to the more complex satellite models in our dataset. 
    \item RANSAC-fusion: Dropout probability of 0.1 following~\cite{gal2016dropout}. We selected $Q=32$ to ensure that our data approximates a normal distribution and have a testing batch size in accordance with CUDA best practices~\cite{nvidiapowerof2batch}.
\end{itemize}
All methods use 10,000 RANSAC iterations to calculate the final pose with a reprojection error threshold of 20 pixels.

\subsubsection{Error plots and success rates}\label{sec:errorplots}
\begin{figure}[h]
    \centering
    \includegraphics[width=0.95\linewidth]{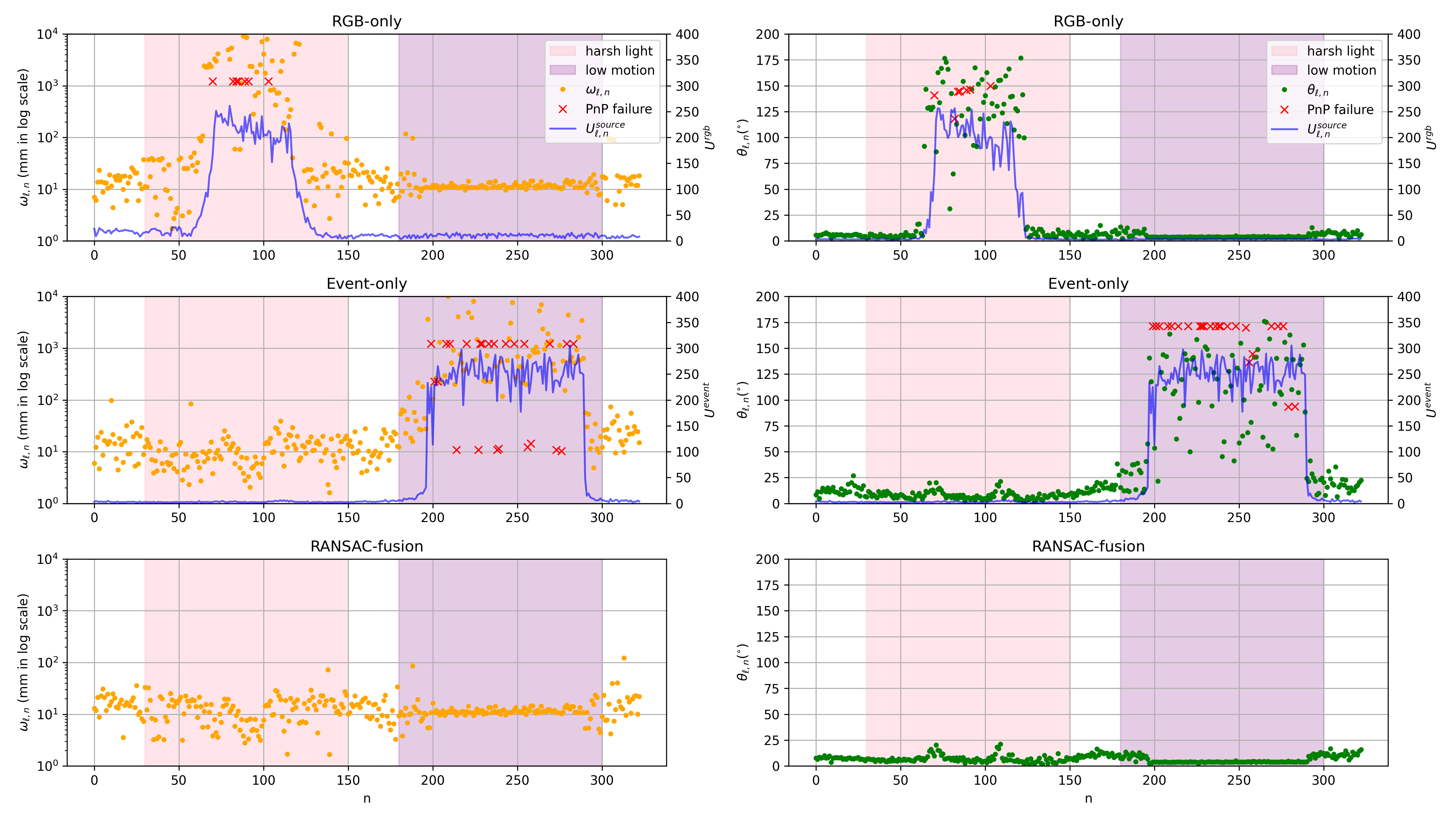}
    \caption{Error and uncertainty plots of the sequence \texttt{cassini-3-far}.}
    \label{fig:good-cassini-plot}
\end{figure}

\begin{figure}[h]
    \centering
    \includegraphics[width=0.95\linewidth]{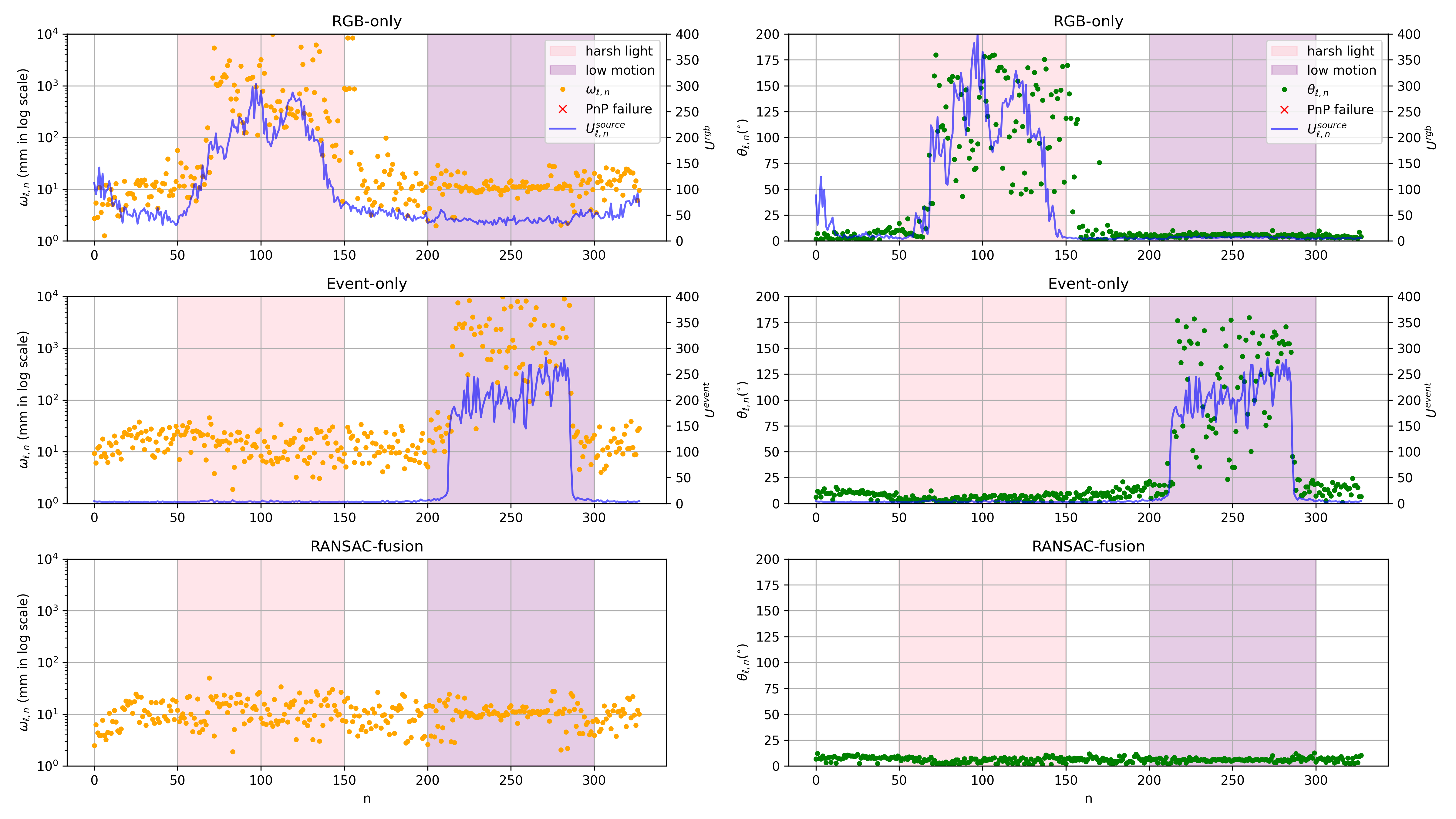}
    \caption{Error and uncertainty plots of the sequence \texttt{soho-3-close}.}
    \label{fig:good-soho-plot}
\end{figure}

\begin{figure}[h]
    \centering
    \includegraphics[width=0.95\linewidth]{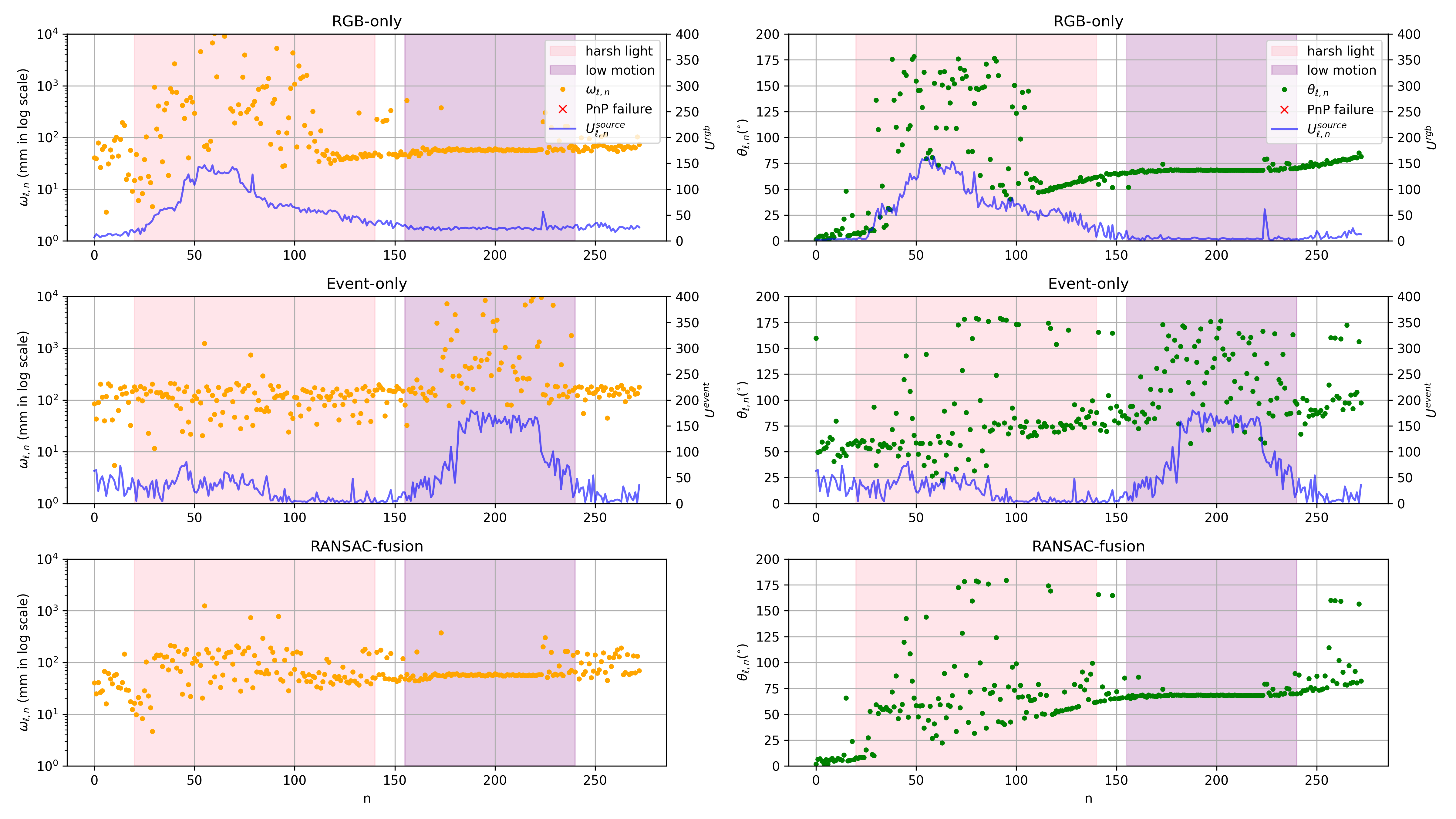}
    \caption{Error and uncertainty plots of the sequence \texttt{satty-4-far}.}
    \label{fig:bad-satty-plot}
\end{figure}

\begin{figure}[h]
    \centering
    \includegraphics[width=0.95\linewidth]{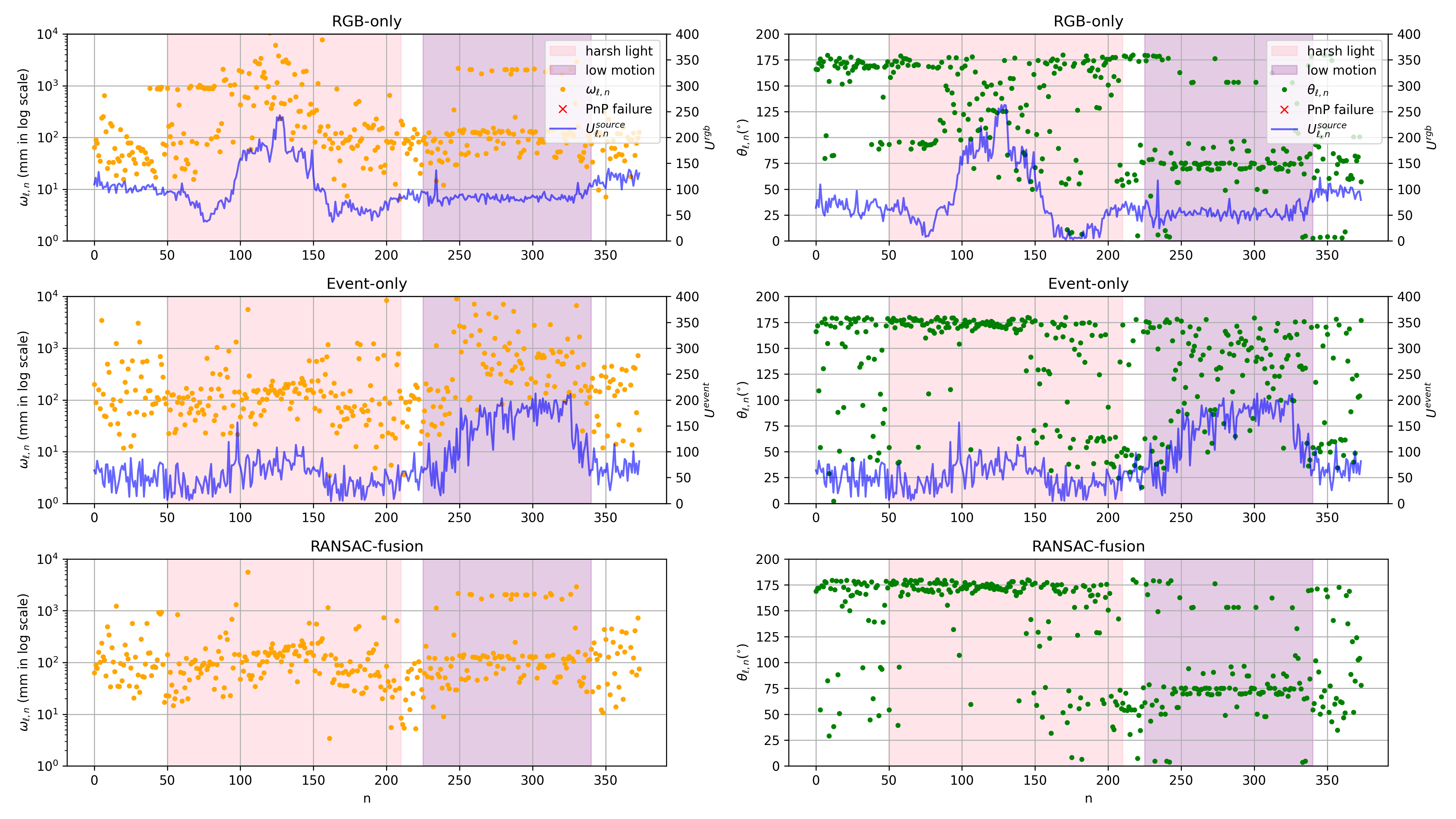}
    \caption{Error and uncertainty plots of the sequence \texttt{soho-4-far}.}
    \label{fig:bad-soho-plot}
\end{figure}

\begin{figure}[h]
    \centering
    \includegraphics[width=0.95\linewidth]{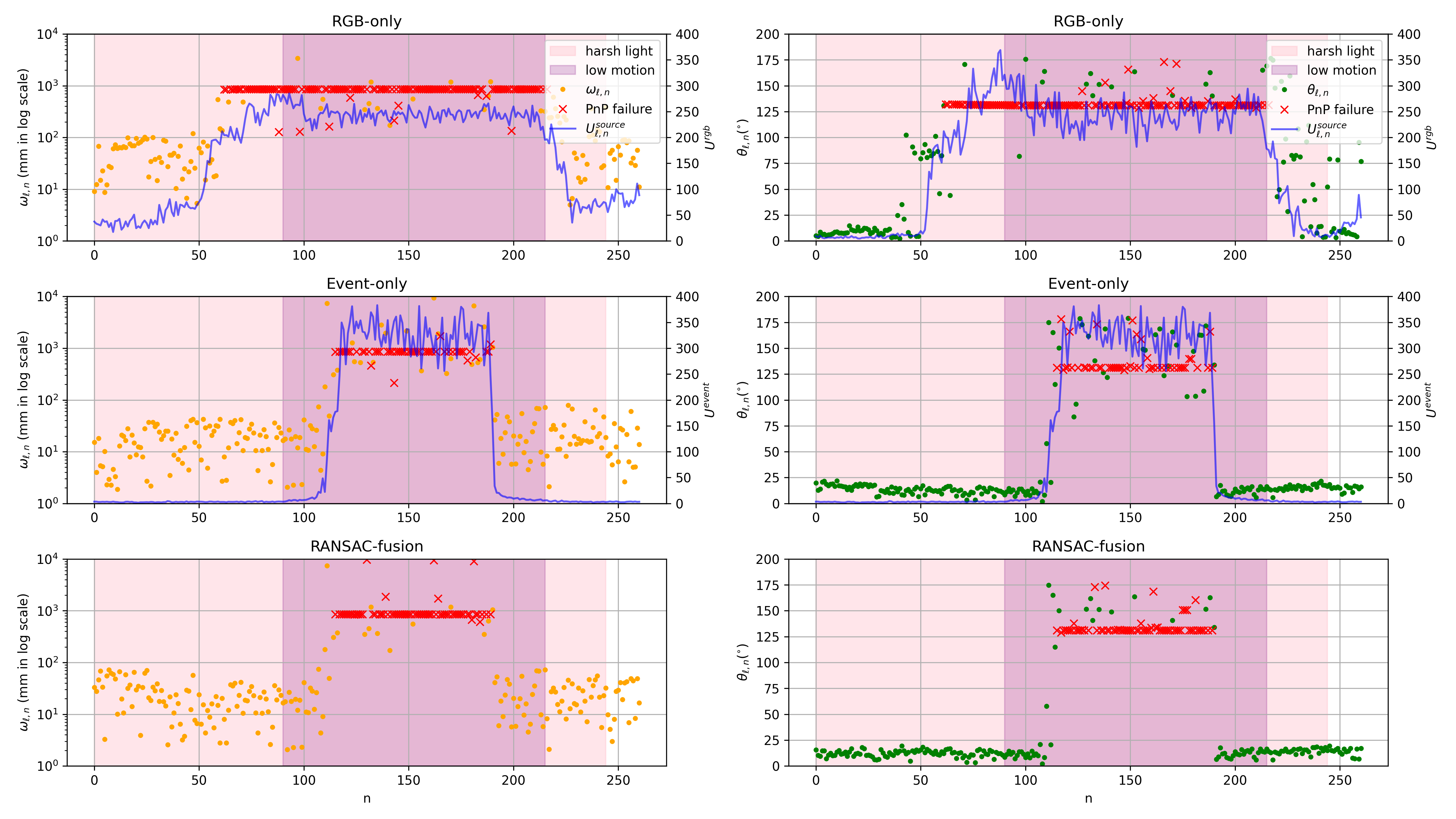}
    \caption{Error and uncertainty plots of the sequence \texttt{cassini-2-close} where the harsh lighting and low-motion periods occur together.}
    \label{fig:confounding-plot}
\end{figure}

We show the success rates for each sequence in Table~\ref{tab:successratestable} which quantitatively demonstrate the efficacy of our fusion method. For each row, we calculated the average success rate of two columns corresponding to all samples (subscript $\ell$) and harsh/low-motion periods (subscript $\ell,\Psi$) for each method. The harsh and low-motion periods correspond to the pink and purple shaded areas in the error plots: Figs.~\ref{fig:good-cassini-plot}, \ref{fig:good-soho-plot}, \ref{fig:bad-satty-plot}, \ref{fig:bad-soho-plot} and \ref{fig:confounding-plot} (for plots of other sequences, refer to Appendix~\ref{sec:appendix}). If the average success rate for the method for that row is the maximum in that row, we highlight that in \textbf{bold}. Furthermore, we also illustrate that our method only sees the real samples at test-time and does not conduct any online self-training as opposed to the RGB-UDA and Event-$\zeta$ methods. To show this, we highlight in green for each row the columns for the method which has the maximum average success rate among the RGB-only, Event-only, RANSAC-fusion' and RANSAC-fusion methods only, \ie only the methods entirely trained on synthetic data and tested on real data without the algorithm having any prior knowledge of the real data or refining the deep learning model on it at test-time.

Moreover, we also plotted per-frame errors $\omega_{\ell,n}$ and $\theta_{\ell,n}$ to show how the rotation and translation errors change over time in each sequence. In several instances, the PnP solver is unable to output a unique result due to the low count(<4) of inliers found by RANSAC. The specific OpenCV implementation of the EPnP version of this solver outputs a default $\mathbf{0}$ vector for both translation and rotation in these cases. For cases like these, we have indicated in the plots with a red cross that the error is based on the default result. We also shade the areas corresponding to harsh lighting in light pink and areas corresponding to low-motion in purple. The uncertainty for each modality is also plotted to demonstrate the correlation of the uncertainty value with harsh lighting and low-motion periods. To indicate the best-case and worst-case performance of our method, we plot two sequences of the highest success rates and two sequences of the lowest success rates(refer to Table~\ref{tab:successratestable} for the success rate values). These plots are given in Figs.~\ref{fig:good-cassini-plot}, \ref{fig:good-soho-plot}, \ref{fig:bad-satty-plot} and \ref{fig:bad-soho-plot}. The plots for all other sequences are provided in the Appendix.

\begin{table*}[ht]\centering
{\footnotesize	
\begin{NiceTabular}{lcccccccccccc}[vlines,rules/color=[gray]{0.3},color-inside]
\hline
Method & \multicolumn{2}{c}{RGB-UDA} & \multicolumn{2}{c}{Event-$\zeta$} & \multicolumn{2}{c}{RGB-only}& \multicolumn{2}{c}{Event-only} & \multicolumn{2}{c}{RANSAC-fusion'} & \multicolumn{2}{c}{RANSAC-fusion} \\
\hline
 Testing subset & $\bI^{rgb}_{\ell}$ & $\bI^{rgb}_{\ell,\Psi}$ & $\bI^{event}_{\ell}$ & $\bI^{event}_{\ell,\Psi}$ & $\bI^{rgb}_{\ell}$ & $\bI^{rgb}_{\ell,\Psi}$ & $\bI^{event}_{\ell}$ & $\bI^{event}_{\ell,\Psi}$ & $D_{\ell}$ & $D_{\ell,\Psi}$ & $D_{\ell}$ & $D_{\ell,\Psi}$ \\ \hline
 \multicolumn{13}{c}{$\Omega_{\ell}$} \\
 \hline
 satty-1-close & 0.90 & 0.80 & 0.82 & 0.66 & 0.85 & 0.72 & 0.71 & 0.57 & \cellcolor{green}\textbf{0.99} & \cellcolor{green}\textbf{0.98} & \cellcolor{green}\textbf{0.99} & \cellcolor{green}\textbf{0.98} \\
satty-1-far & 0.74 & 0.65 & 0.73 & 0.49 & 0.65 & 0.53 & 0.28 & 0.11 & 0.78 & 0.67 & \cellcolor{green}\textbf{0.80} & \cellcolor{green}\textbf{0.69} \\
satty-2-close & \textbf{0.89} & \textbf{0.84} & 0.81 & 0.74 & 0.60 & 0.44 & 0.65 & 0.62 & \cellcolor{green}0.80 & \cellcolor{green}0.72 & 0.79 & 0.70 \\
satty-2-far & 0.53 & 0.38 & \textbf{0.67} & \textbf{0.52} & 0.30 & 0.10 & 0.22 & 0.07 & \cellcolor{green}0.44 & \cellcolor{green}0.20 & 0.42 & 0.20 \\
satty-3-close & 0.92 & 0.86 & 0.77 & 0.60 & 0.88 & 0.78 & 0.77 & 0.59 & 0.99 & 0.97 & \cellcolor{green}\textbf{0.99} & \cellcolor{green}\textbf{0.98} \\
satty-3-far & \textbf{0.90} & \textbf{0.80} & 0.79 & 0.57 & 0.86 & 0.71 & 0.30 & 0.12 & \cellcolor{green}0.87 & \cellcolor{green}0.75 & 0.86 & 0.74 \\
satty-4-close & 0.72 & 0.63 & \textbf{0.76} & \textbf{0.68} & 0.35 & 0.19 & 0.38 & 0.32 & \cellcolor{green}0.63 & \cellcolor{green}0.51 & 0.57 & 0.43 \\
satty-4-far & 0.04 & 0.04 & \textbf{0.20} & \textbf{0.17} & \cellcolor{green}0.04 & \cellcolor{green}0.03 & 0.01 & 0.00 & 0.03 & 0.02 & 0.03 & 0.03 \\
cassini-1-close & 0.42 & 0.38 & 0.44 & 0.43 & 0.10 & 0.11 & 0.45 & 0.43 & \cellcolor{green}\textbf{0.51} & \cellcolor{green}\textbf{0.53} & 0.51 & 0.52 \\
cassini-1-far & 0.26 & 0.23 & 0.33 & 0.20 & 0.18 & 0.15 & \cellcolor{green}\textbf{0.29} & \cellcolor{green}\textbf{0.31} & 0.24 & 0.28 & 0.26 & 0.29 \\
cassini-2-close & 0.26 & 0.22 & 0.27 & 0.24 & 0.06 & 0.06 & \cellcolor{green}\textbf{0.40} & \cellcolor{green}\textbf{0.37} & 0.33 & 0.32 & 0.35 & 0.34 \\
cassini-2-far & \textbf{0.29} & \textbf{0.27} & 0.00 & 0.00 & 0.09 & 0.08 & \cellcolor{green}0.28 & \cellcolor{green}0.27 & 0.21 & 0.22 & 0.19 & 0.21 \\
cassini-3-close & \textbf{0.29} & \textbf{0.25} & 0.00 & 0.00 & 0.02 & 0.03 & 0.28 & 0.22 & \cellcolor{green}0.27 & \cellcolor{green}0.26 & 0.26 & 0.22 \\
cassini-3-far & 0.75 & 0.72 & 0.70 & 0.60 & 0.56 & 0.51 & 0.53 & 0.49 & \cellcolor{green}\textbf{0.87} & \cellcolor{green}\textbf{0.91} & \cellcolor{green}\textbf{0.87} & \cellcolor{green}\textbf{0.91} \\
cassini-4-close & 0.23 & 0.15 & \textbf{0.48} & \textbf{0.37} & 0.01 & 0.00 & 0.44 & 0.34 & 0.45 & 0.35 & \cellcolor{green}0.46 & \cellcolor{green}0.36 \\
cassini-4-far & \textbf{0.28} & \textbf{0.19} & 0.03 & 0.04 & 0.16 & 0.06 & 0.07 & 0.06 & \cellcolor{green}0.20 & \cellcolor{green}0.20 & 0.19 & 0.18 \\
soho-1-close & 0.25 & 0.23 & \textbf{0.73} & \textbf{0.63} & 0.05 & 0.03 & \cellcolor{green}0.35 & \cellcolor{green}0.27 & 0.36 & 0.23 & \cellcolor{green}0.37 & \cellcolor{green}0.25 \\
soho-1-far & 0.07 & 0.04 & \textbf{0.10} & \textbf{0.07} & 0.03 & 0.03 & 0.08 & 0.05 & 0.06 & 0.06 & \cellcolor{green}0.08 & \cellcolor{green}0.09 \\
soho-2-close & 0.06 & 0.03 & \textbf{0.70} & \textbf{0.60} & 0.02 & 0.00 & 0.33 & 0.19 & \cellcolor{green}0.40 & \cellcolor{green}0.27 & 0.36 & 0.22 \\
soho-2-far & 0.03 & 0.02 & \textbf{0.11} & \textbf{0.08} & 0.01 & 0.00 & 0.07 & 0.06 & 0.06 & 0.07 & \cellcolor{green}0.06 & \cellcolor{green}0.08 \\
soho-3-close & 0.71 & 0.65 & 0.38 & 0.37 & 0.61 & 0.50 & 0.55 & 0.45 & \cellcolor{green}\textbf{0.92} & \cellcolor{green}\textbf{0.92} & 0.90 & 0.88 \\
soho-3-far & 0.32 & 0.30 & 0.04 & 0.06 & 0.32 & 0.27 & 0.16 & 0.17 & 0.36 & 0.35 & \cellcolor{green}\textbf{0.40} & \cellcolor{green}\textbf{0.41} \\
soho-4-close & 0.04 & 0.02 & 0.08 & 0.05 & 0.05 & 0.04 & 0.08 & 0.09 & \cellcolor{green}\textbf{0.12} & \cellcolor{green}\textbf{0.08} & \cellcolor{green}\textbf{0.11} & \cellcolor{green}\textbf{0.09} \\
soho-4-far & \textbf{0.10} & \textbf{0.10} & 0.09 & 0.03 & 0.03 & 0.02 & 0.04 & 0.04 & \cellcolor{green}0.09 & \cellcolor{green}0.06 & 0.05 & 0.03 \\
\hline
Avg. across all $D_{\ell}$ & 0.42 & 0.37 & 0.42 & 0.34 & 0.29 & 0.22 & 0.32 & 0.26 & \cellcolor{green}\textbf{0.46} & \cellcolor{green}\textbf{0.41} & 0.45 & 0.41 \\
 \hline
 \multicolumn{13}{c}{$\Theta_{\ell}$} \\
 \hline
 satty-1-close & 0.85 & 0.70 & 0.82 & 0.65 & 0.83 & 0.67 & 0.60 & 0.46 & 0.96 & 0.92 & \cellcolor{green}\textbf{0.96} & \cellcolor{green}\textbf{0.94} \\
satty-1-far & 0.79 & 0.64 & 0.73 & 0.48 & 0.77 & 0.63 & 0.18 & 0.04 & \cellcolor{green}\textbf{0.82} & \cellcolor{green}\textbf{0.68} & \cellcolor{green}\textbf{0.82} & \cellcolor{green}\textbf{0.68} \\
satty-2-close & \textbf{0.82} & \textbf{0.75} & 0.81 & 0.74 & 0.46 & 0.24 & 0.52 & 0.52 & \cellcolor{green}0.73 & \cellcolor{green}0.61 & 0.72 & 0.60 \\
satty-2-far & 0.40 & 0.13 & \textbf{0.69} & \textbf{0.55} & 0.36 & 0.10 & 0.16 & 0.02 & 0.38 & 0.10 & \cellcolor{green}0.38 & \cellcolor{green}0.11 \\
satty-3-close & 0.87 & 0.78 & 0.77 & 0.60 & 0.84 & 0.71 & 0.75 & 0.56 & \cellcolor{green}\textbf{0.98} & \cellcolor{green}\textbf{0.97} & \cellcolor{green}\textbf{0.98} & \cellcolor{green}\textbf{0.97} \\
satty-3-far & \textbf{0.86} & \textbf{0.73} & 0.79 & 0.57 & 0.85 & 0.69 & 0.12 & 0.03 & \cellcolor{green}0.86 & \cellcolor{green}0.71 & 0.84 & 0.68 \\
satty-4-close & 0.72 & 0.63 & \textbf{0.76} & \textbf{0.68} & 0.32 & 0.17 & 0.22 & 0.21 & \cellcolor{green}0.49 & \cellcolor{green}0.42 & 0.45 & 0.36 \\
satty-4-far & \textbf{0.10} & \textbf{0.04} & 0.09 & 0.02 & \cellcolor{green}0.08 & \cellcolor{green}0.03 & 0.00 & 0.00 & 0.07 & 0.02 & 0.08 & 0.02 \\
cassini-1-close & 0.45 & 0.45 & \textbf{0.74} & \textbf{0.70} & 0.22 & 0.22 & 0.14 & 0.12 & \cellcolor{green}0.44 & \cellcolor{green}0.41 & 0.32 & 0.28 \\
cassini-1-far & 0.87 & 0.85 & 0.68 & 0.66 & 0.80 & 0.72 & 0.47 & 0.48 & 0.95 & 0.93 & \cellcolor{green}\textbf{0.95} & \cellcolor{green}\textbf{0.94} \\
cassini-2-close & 0.29 & 0.30 & \textbf{0.48} & \textbf{0.44} & 0.16 & 0.16 & 0.13 & 0.13 & \cellcolor{green}0.21 & \cellcolor{green}0.21 & 0.15 & 0.15 \\
cassini-2-far & \textbf{0.94} & \textbf{0.94} & 0.22 & 0.19 & 0.30 & 0.22 & 0.47 & 0.47 & 0.54 & 0.50 & \cellcolor{green}0.55 & \cellcolor{green}0.50 \\
cassini-3-close & \textbf{0.66} & \textbf{0.64} & 0.57 & 0.59 & 0.01 & 0.01 & 0.38 & 0.39 & \cellcolor{green}0.47 & \cellcolor{green}0.47 & 0.43 & 0.42 \\
cassini-3-far & 0.75 & 0.68 & 0.63 & 0.58 & 0.76 & 0.71 & 0.33 & 0.39 & \cellcolor{green}\textbf{0.81} & \cellcolor{green}\textbf{0.88} & \cellcolor{green}\textbf{0.81} & \cellcolor{green}\textbf{0.88} \\
cassini-4-close & 0.27 & 0.21 & \textbf{0.54} & \textbf{0.47} & 0.01 & 0.00 & 0.44 & 0.37 & \cellcolor{green}0.53 & \cellcolor{green}0.45 & 0.48 & 0.39 \\
cassini-4-far & 0.74 & 0.61 & \textbf{0.81} & \textbf{0.69} & 0.38 & 0.23 & 0.48 & 0.28 & 0.64 & 0.43 & \cellcolor{green}0.70 & \cellcolor{green}0.53 \\
soho-1-close & 0.06 & 0.02 & \textbf{0.77} & \textbf{0.67} & 0.03 & 0.00 & 0.42 & 0.34 & 0.48 & 0.33 & \cellcolor{green}0.48 & \cellcolor{green}0.34 \\
soho-1-far & 0.30 & 0.20 & \textbf{0.69} & \textbf{0.58} & 0.10 & 0.01 & 0.33 & 0.16 & 0.39 & 0.20 & \cellcolor{green}0.39 & \cellcolor{green}0.21 \\
soho-2-close & 0.03 & 0.02 & \textbf{0.81} & \textbf{0.76} & 0.01 & 0.00 & 0.42 & 0.28 & \cellcolor{green}0.44 & \cellcolor{green}0.29 & 0.43 & 0.28 \\
soho-2-far & 0.07 & 0.03 & \textbf{0.74} & \textbf{0.68} & 0.10 & 0.00 & 0.25 & 0.13 & 0.27 & 0.13 & \cellcolor{green}0.28 & \cellcolor{green}0.14 \\
soho-3-close & 0.80 & 0.68 & 0.76 & 0.61 & 0.65 & 0.51 & 0.52 & 0.53 & \cellcolor{green}\textbf{0.97} & \cellcolor{green}\textbf{0.99} & 0.96 & 0.98 \\
soho-3-far & 0.80 & 0.73 & 0.75 & 0.64 & 0.65 & 0.57 & 0.49 & 0.49 & \cellcolor{green}\textbf{0.93} & \cellcolor{green}\textbf{0.91} & 0.92 & 0.90 \\
soho-4-close & \textbf{0.02} & \textbf{0.01} & 0.00 & 0.00 & \cellcolor{green}0.01 & \cellcolor{green}0.01 & 0.00 & 0.00 & 0.01 & 0.00 & 0.01 & 0.00 \\
soho-4-far & \textbf{0.10} & \textbf{0.09} & 0.00 & 0.00 & 0.02 & 0.01 & 0.00 & 0.00 & \cellcolor{green}0.03 & \cellcolor{green}0.03 & 0.02 & 0.03 \\
\hline
Avg. across all $D_{\ell}$ & 0.52 & 0.45 & \textbf{0.61} & \textbf{0.52} & 0.36 & 0.28 & 0.33 & 0.27 & \cellcolor{green}0.56 & \cellcolor{green}0.48 & 0.55 & 0.47 \\
 \hline
\end{NiceTabular}}
\caption{Success rate per sequence of estimating translation (top) and rotation (bottom). Bold font indicates best per-method success rate (average of subscript $\ell$ and $\ell,\Psi$) across all methods. Green highlight indicates best per-method success rate among methods trained only on synthetic data.} 
\label{tab:successratestable}
\end{table*}

\subsection{Qualitative results of our method}
\begin{figure}[h]
    \centering
    \includegraphics[width=0.95\linewidth]{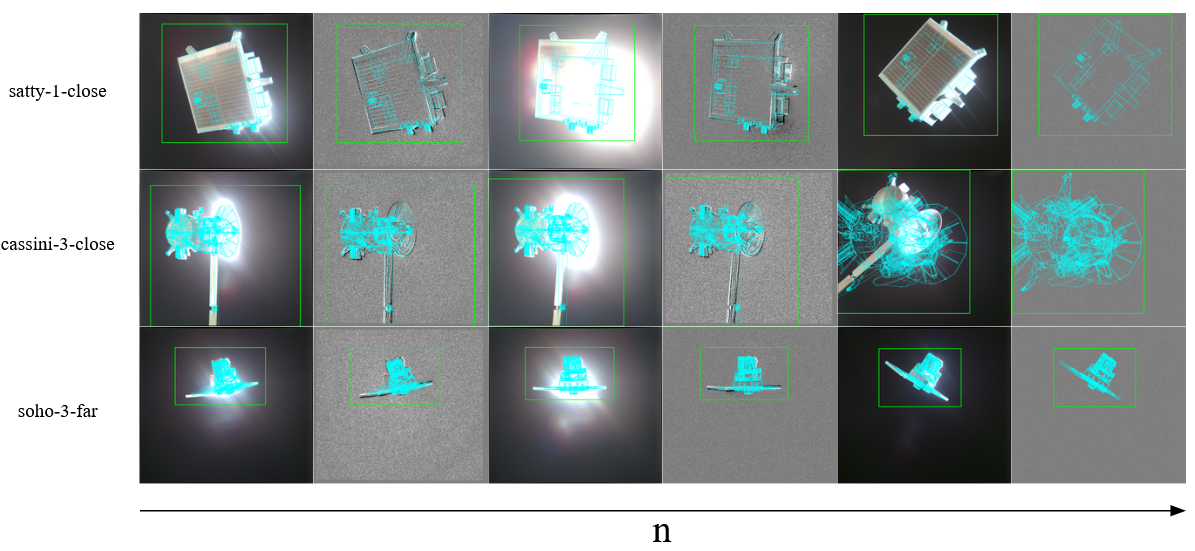}
    \caption{Visualization of the estimated bounding box in green and the wireframe of the CAD model of each satellite reprojected using the final estimated pose given by the RANSAC-fusion method in cyan.}
    \label{fig:qualitativeresults}
\end{figure}

\begin{figure}[h]
    \centering
    \includegraphics[width=0.6\linewidth]{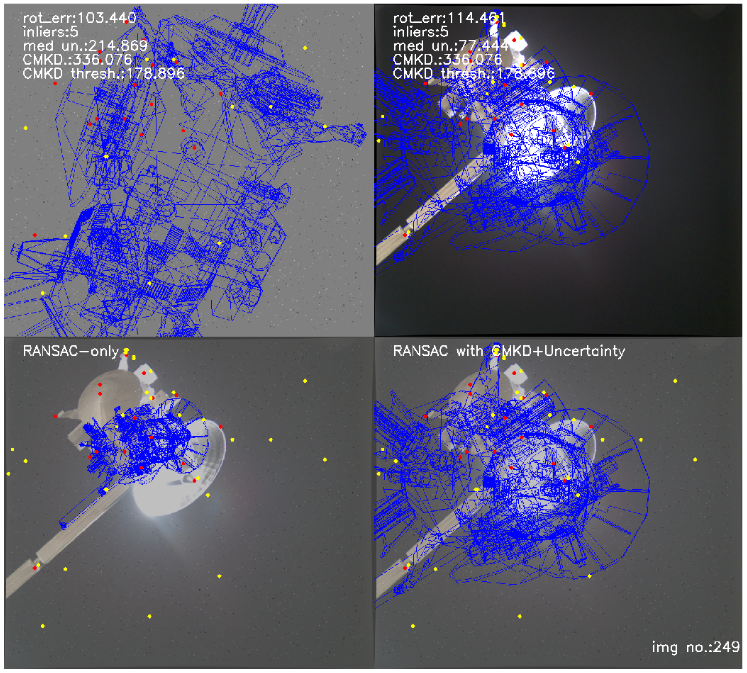}
    \caption{Visualization of the Event-only and RGB-only results (top row) followed by the RANSAC-fusion method with and without the CMKD and uncertainty-based decision making (bottom row). Neither the cross-modal RANSAC nor the CMKD and uncertainty-based decision making are able to optimize the result in cases like these likely due to the remaining domain gap in the two input channels. In each image, the wireframe reprojected using the estimated pose is plotted in blue, the yellow dots are the predicted keypoints and red dots are the ground-truth keypoints.}
    \label{fig:bothchannelsbadexample}
\end{figure}

We visualize a subset of results from a sequence from each satellite to demonstrate that the estimated pose using our method correctly lines up the satellite CAD model with the testing frames even through harsh lighting and low event SNR (last column Fig.~\ref{fig:qualitativeresults}). Note the last result of the \texttt{cassini-3-close} row in Fig.~\ref{fig:qualitativeresults} which is due to both channels suffering due to a potential domain gap as shown specifically in Fig.~\ref{fig:bothchannelsbadexample}. For such cases, we present ideas in Sec.~\ref{sec:future} which can drive future work.

\subsection{Effect of the CMKD and uncertainty-based decision making}
\begin{figure}[h]
    \centering
    \includegraphics[width=0.8\linewidth]{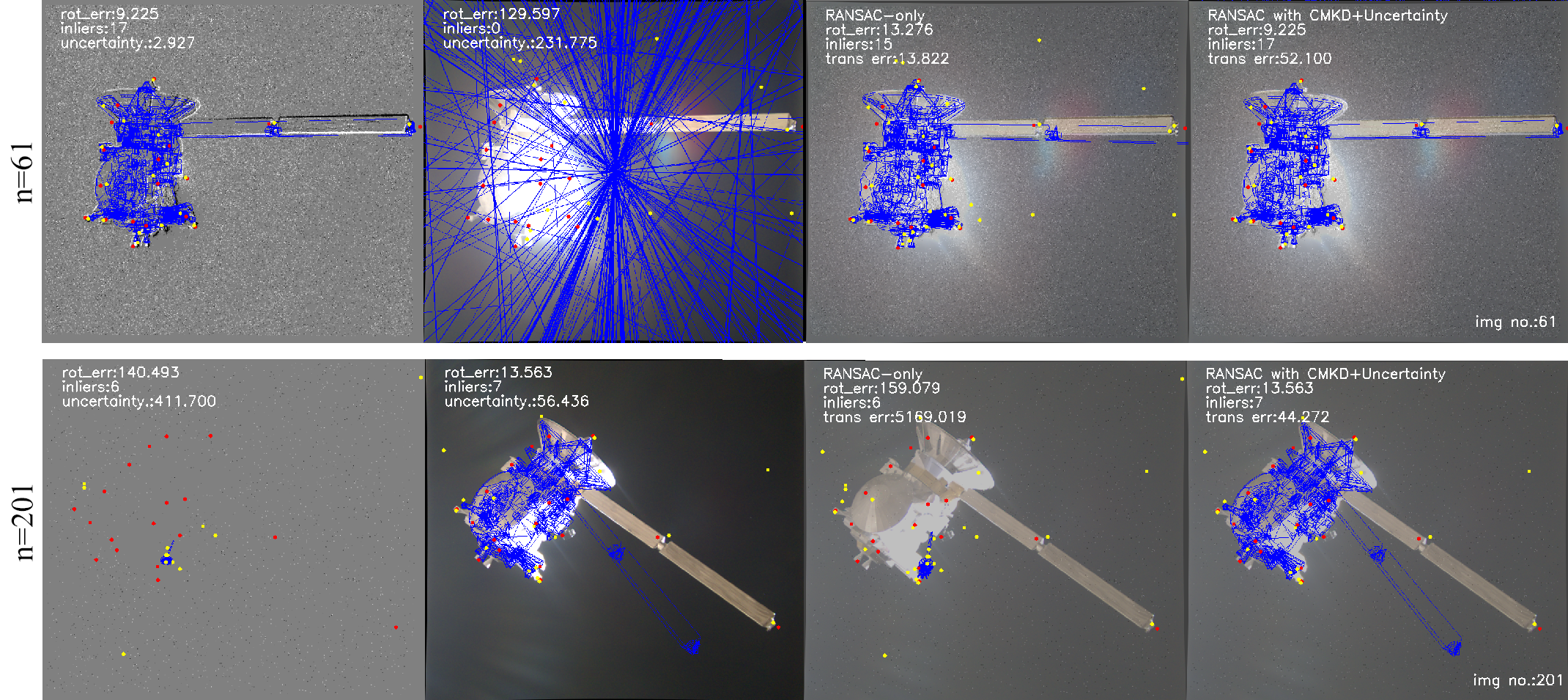}
    \caption{Effect of the CMKD and uncertainty-based decision-making on the translation and rotation error. The decision-making process is able to select the channel with lower uncertainty when the RANSAC-only filtering favors false-positive inlier points. We show two examples from the \texttt{cassini-1-close} sequence. The bottom row shows how the edge case presented in Fig.~\ref{fig:degeneratecase} is handled by our method in comparison to just using RANSAC to filter points from the two channels. In each image, the wireframe reprojected using the estimated pose is plotted in blue, the yellow dots are the predicted keypoints and red dots are the ground-truth keypoints.}
    \label{fig:cmkdsolvesedgecases}
\end{figure}

The extended method to handle the edge cases due to clustered predictions as described in Sec.~\ref{sec:cmkd} and ~\ref{sec:opticalalignment} has roughly on par performance compared to the RANSAC-only filtering, offering only a very slight gain in performance in some scenes as seen in Table~\ref{tab:successratestable}. This is because the edge cases are a rare occurrence. In most cases, with enough iterations, RANSAC is able to find the correct inlier points from the larger set of the observations provided by the two combined channels. In Fig.~\ref{fig:cmkdsolvesedgecases} we show two examples of how incorporating CMKD and uncertainty is able to improve on the RANSAC-only result. Moreover, the plots in Figs.~\ref{fig:good-soho-plot} and \ref{fig:good-cassini-plot} show that the uncertainty measure is well correlated with the periods of harsh lighting and low-motion. The uncertainty also correlates well with the error values. Even in Fig.~\ref{fig:bad-soho-plot} where our method does not perform well, the uncertainty measure still correlates well with the challenging periods. This indicates that even if the method is not performing the pose estimation well, we can use the uncertainty measure to determine and relay challenging conditions to on-ground operators.

\subsection{Computation times}
We determine per-instance computation times by measuring the actual time during inference of a single batch and diving by its batch size. The per-instance time per batch is then averaged over all the batches in a sequence. We selected the \texttt{cassini-1-close} sequence as a representative scenario, as it requires the full method (CMKD + uncertainty-based decision making) on 112 out of 277 test instances. Including object detection and landmark regression, average inference time per instance are as follows:
\begin{itemize}
    \item Single channel (RGB-only and event-only have comparable times): 0.111s
    \item RANSAC-fusion': 0.258s
    \item RANSAC-fusion: 0.757s
\end{itemize}
We observe that the RANSAC-fusion' which does not incorporate the CMKD and uncertainty is roughly double the time taken for the single channel method. This is expected as we are evaluating two different DNNs and the forward propagation is conducted twice for both the object detection as well as the landmark regression. Given the above results, we can safely assume that the full fusion method, including CMKD and uncertainty-based decision, takes at most 1 frame per second on modern workstation GPUs, like the RTX 3090 in this case.

For onboard deployment, our pipeline needs to be adapted to ensure compatibility with low-power hardware accelerators such as FPGAs~\cite{fpgaposeestimation}. For orbital missions, which typically operate over several hours~\cite{rendezvousmissiontime}, we anticipate that even at a reduced frame rate below 1 fps, our pipeline might still deliver valuable pose estimates and information, provided the scene remains relatively stable and does not exhibit high levels of dynamism.

\subsection{Discussion}\label{sec:resultsdiscussion}
Results in Table~\ref{tab:successratestable} demonstrate the efficacy of using RANSAC-based fusion on RGB and event data that are optically aligned and temporally synchronized. On average, we see a two-fold performance gain when using both channels together as compared to the singular channel only methods, as illustrated by the last rows of the $\Omega_{\ell}$ and $\Theta_{\ell}$ sub-tables of Table~\ref{tab:successratestable}. In some scenes the RGB-UDA and Event-$\zeta$ methods are comparable or offer slightly better performance than our fusion method; however these methods utilize test-time learning or require hundreds of unlabeled data samples for offline refinement which for future rendezvous scenarios is often infeasible. We see that in general the \texttt{far} sequences are challenged more as there are fewer texture details in the event data. The RGB channel fares better in the \texttt{far} and is able to aid the fusion method into obtaining a better result than the event channel alone can estimate as seen in Fig.~\ref{fig:bad-satty-plot}. The most challenging scene is \texttt{soho-4} at both distances due to the symmetric geometry of SOHO at certain poses. Even with the additional $Z=24$ landmarks compared to the other two target objects, the symmetry problem creates a great challenge for the landmark regressor. We do not consider the symmetry problem in our pose estimation pipeline but our dataset provides an opportunity to drive future work in this area. Moreover, in the sequences where the harsh lighting and low-motion occur together, the method only performs as well as the better individual channel. We see this phenomenon in the error plots for the sequence \texttt{cassini-2-close} in Fig.~\ref{fig:confounding-plot}. In this sequence, the performance of our method is roughly the same as the event-only channel because the harsh lighting, which affects the RGB channel more, occurs over a larger period than the low-motion, which is a greater challenge for the event channel. Finally, the plots of $\omega_{\ell,n}$ and $\theta_{\ell,n}$ show that in harsh-lighting and low-motion conditions using only a single channel results in many cases where the PnP solver cannot converge to a unique solution. Our fusion method is effective in reducing such cases across all sequences. 

\section{Conclusions}\label{sec:conclusion}
In this work, we proposed an event-RGB fusion method which addressed the shortcomings of individual sensor channels. Specifically, we fused the intermediate keypoint predictions by exploiting the outlier-rejection robustness of RANSAC. We proposed the FRESH dataset consisting of event-RGB sequences captured in a real lab setting under harsh-lighting to evaluate the performance of our proposed method against existing single channel methods. Our results demonstrate the efficacy of this fusion scheme especially under harsh-lighting and low-motion scenarios where the method reduced PnP failures and converges on a more accurate pose using the inlier 2D-3D correspondences from the two input sensor channels. Our CMKD and uncertainty-based decision making components were able to solve the degenerate case scenarios under which RANSAC alone cannot detect inlier correspondences. We hope that this work drives the utilization of event-RGB sensor rigs for pose estimation in an orbital setting.

\subsection{Limitations and future considerations}\label{sec:future}
\begin{itemize}
\item The proposed RANSAC fusion method is not able to \emph{simultaneously} deal with harsh lighting and low relative motion. A potential solution is introducing deliberate jitter to the event sensor~\cite{latif2023high} to generate data for spacecraft pose estimation.
\item While the event-RGB sensor combination offers compelling performance in harsh-lighting scenarios, a more comprehensive study needs to be conducted to determine the best sensor combination for our RANSAC-based fusion method. There is also a potential to use multiple beamsplitters, such as in the work of~\cite{Zou_2021_CVPR}, to fuse more than two sensors. Our RANSAC-fusion can incorporate an arbitrary number of 2D-3D correspondences, therefore is already compatible with extension to greater than two sensor channels.
\item Due to the focus on harsh lighting, the ability of our pose estimation pipeline to handle clutter and occlusions has not been evaluated. However, similar to other sparse correspondence methods~\cite[Sec.~3.1.1]{liu2024deep}, training on data with clutter and occlusions should produce some tolerance.
\item Consider object symmetry for pose estimation on sequential frames/event data. Especially as many space-borne objects like SOHO, Soyuz and cube satellites observe a symmetric geometry in some or all axes.
\item The labeling process for the FRESH dataset was conducted by a single annotator. Therefore, the possibility of inaccuracies in the labels resulting from subjective bias cannot be entirely excluded. For the purpose of this study, the manual labels are sufficient as generous thresholds used for the success rates (refer to Sec.~\ref{sec:successratemetric}), mitigate any major impact on the performance evaluation. The performance gain of our method is also apparent due to the reduction of PnP failure cases which are not determined by the ground-truth labels (such as in Fig.~\ref{fig:appendixsatty1}). In future studies, where more fine-grained performance gains might need to be considered, we will incorporate technologies such as industrial robots and motion tracking sensors to obtain more accurate ground-truth labels.
\end{itemize}

\section*{Funding sources}
Tat-Jun Chin is SmartSat CRC Chair of Sentient Satellites.

\clearpage
\bibliographystyle{elsarticle-num}
\bibliography{references}

\begin{thebibliography}{10}
\expandafter\ifx\csname url\endcsname\relax
  \def\url#1{\texttt{#1}}\fi
\expandafter\ifx\csname urlprefix\endcsname\relax\def\urlprefix{URL }\fi
\expandafter\ifx\csname href\endcsname\relax
  \def\href#1#2{#2} \def\path#1{#1}\fi

\bibitem{renaut2025deep}
L.~Renaut, H.~Frei, A.~N{\"u}chter, Deep learning on 3d point clouds for fast pose estimation during satellite rendezvous, Acta Astronautica 232 (2025) 231--243.

\bibitem{bechini2025robust}
M.~Bechini, M.~Lavagna, Robust and efficient single-cnn-based spacecraft relative pose estimation from monocular images, Acta Astronautica (2025).

\bibitem{pasqualetto2019review}
L.~{Pasqualetto Cassinis}, R.~Fonod, E.~Gill, \href{https://www.sciencedirect.com/science/article/pii/S0376042119300302}{Review of the robustness and applicability of monocular pose estimation systems for relative navigation with an uncooperative spacecraft}, Progress in Aerospace Sciences 110 (2019) 100548.
\newblock \href {https://doi.org/https://doi.org/10.1016/j.paerosci.2019.05.008} {\path{doi:https://doi.org/10.1016/j.paerosci.2019.05.008}}.
\newline\urlprefix\url{https://www.sciencedirect.com/science/article/pii/S0376042119300302}

\bibitem{cavaciuti2022}
A.~J. Cavaciuti, J.~H. Heying, J.~Davis, In-space servicing, assembly, and manufacturing for the new space economy, Tech. rep., Center for Space Policy and Strategy, Aerospace Corporation (2022).

\bibitem{reed2016restorel}
B.~B. Reed, R.~C. Smith, B.~J. Naasz, J.~F. Pellegrino, C.~E. Bacon, The restore-l servicing mission, in: AIAA space 2016, 2016, p. 5478.

\bibitem{activedebrisremoval}
M.~Poozhiyil, M.~H. Nair, M.~C. Rai, A.~Hall, C.~Meringolo, M.~Shilton, S.~Kay, D.~Forte, M.~Sweeting, N.~Antoniou, et~al., Active debris removal: A review and case study on leopard phase 0-a mission, Advances in space research 72~(8) (2023) 3386--3413.

\bibitem{soyuzdockingteaser}
NASA, Soyuz ms-14 aborted docking, \url{https://www.youtube.com/watch?v=8ipeqxNxaKc} (August 2019).

\bibitem{issdockingteaser1}
NASA, One-year crew docking to the international space station, \url{https://www.youtube.com/watch?v=pRaZgSVnsNs} (May 2015).

\bibitem{issdockingteaser2}
NASA, Crew-7 approach and docking, \url{https://www.youtube.com/watch?v=bEsyo5WLFkQ} (August 2023).

\bibitem{sharma2018pose}
S.~Sharma, C.~Beierle, S.~D'Amico, Pose estimation for non-cooperative spacecraft rendezvous using convolutional neural networks, in: 2018 IEEE Aerospace Conference, IEEE, 2018, pp. 1--12.

\bibitem{kisantal2020satellite}
M.~Kisantal, S.~Sharma, T.~H. Park, D.~Izzo, M.~M{\"a}rtens, S.~D’Amico, Satellite pose estimation challenge: Dataset, competition design, and results, IEEE Transactions on Aerospace and Electronic Systems 56~(5) (2020) 4083--4098.

\bibitem{park2024robust}
T.~H. Park, S.~D’Amico, Robust multi-task learning and online refinement for spacecraft pose estimation across domain gap, Advances in Space Research 73~(11) (2024) 5726--5740.

\bibitem{Park_2022}
T.~H. Park, M.~M\"artens, G.~Lecuyer, D.~Izzo, S.~D’Amico, \href{http://dx.doi.org/10.1109/AERO53065.2022.9843439}{Speed+: Next-generation dataset for spacecraft pose estimation across domain gap}, in: 2022 IEEE Aerospace Conference (AERO), IEEE, 2022.
\newblock \href {https://doi.org/10.1109/aero53065.2022.9843439} {\path{doi:10.1109/aero53065.2022.9843439}}.
\newline\urlprefix\url{http://dx.doi.org/10.1109/AERO53065.2022.9843439}

\bibitem{ursoesaunreal}
P.~Proenca, \href{https://doi.org/10.5281/zenodo.3279632}{Urso: Unreal rendered spacecrafts on-orbit datasets} (2019).
\newblock \href {https://doi.org/10.5281/zenodo.3279632} {\path{doi:10.5281/zenodo.3279632}}.
\newline\urlprefix\url{https://doi.org/10.5281/zenodo.3279632}

\bibitem{kelvinsspec21}
\href{https://kelvins.esa.int/pose-estimation-2021/challenge/}{Kelvins satellite pose estimation competition 2021} (2021).
\newline\urlprefix\url{https://kelvins.esa.int/pose-estimation-2021/challenge/}

\bibitem{park2023adaptiveshirt}
T.~H. Park, S.~D’Amico, Adaptive neural-network-based unscented kalman filter for robust pose tracking of noncooperative spacecraft, Journal of Guidance, Control, and Dynamics 46~(9) (2023) 1671--1688.

\bibitem{dixon2025uncertaintyevidental}
T.~O. Dixon, S.~A. Giles, A.~A. Gorodetsky, Predicting uncertainty in vision-based satellite pose estimation using deep evidential regression, Aerospace Science and Technology (2025) 110055.

\bibitem{bao2021rendering}
Z.~Bao, W.~Xiao, C.~Xue, Rendering a light source based on visual experience-influence of radial streaks on self-luminance perception, in: Advances in Ergonomics in Design: Proceedings of the AHFE 2021 Virtual Conference on Ergonomics in Design, July 25-29, 2021, USA, Springer, 2021, pp. 761--768.

\bibitem{delavennat2021physically}
J.~Delavennat, Physically-based real-time glare (2021).

\bibitem{chen2021self}
Y.~Chen, F.~Liu, K.~Pei, Self-supervised sun glare detection cnn for self-aware autonomous driving, in: Proceedings of the Machine Learning for Autonomous Driving Workshop at the 35th Conference on Neural Information Processing System, Virtual, 2021, pp. 6--14.

\bibitem{sun2021end}
Q.~Sun, C.~Wang, F.~Qiang, D.~Xiong, H.~Wolfgang, End-to-end complex lens design with differentiable ray tracing, ACM Trans. Graph 40~(4) (2021) 1--13.

\bibitem{csurka2017comprehensive}
G.~Csurka, A comprehensive survey on domain adaptation for visual applications, Domain adaptation in computer vision applications (2017) 1--35.

\bibitem{Peng_2018_CVPR_Workshops}
X.~Peng, B.~Usman, N.~Kaushik, D.~Wang, J.~Hoffman, K.~Saenko, Visda: A synthetic-to-real benchmark for visual domain adaptation, in: Proceedings of the IEEE Conference on Computer Vision and Pattern Recognition (CVPR) Workshops, 2018.

\bibitem{kelvinsspec21results}
\href{https://kelvins.esa.int/pose-estimation-2021/leaderboard/lightbox-final-result}{Kelvins spec2021 lightbox final result (closed)} (2022).
\newline\urlprefix\url{https://kelvins.esa.int/pose-estimation-2021/leaderboard/lightbox-final-result}

\bibitem{wang2023bridging}
Z.~Wang, M.~Chen, Y.~Guo, Z.~Li, Q.~Yu, Bridging the domain gap in satellite pose estimation: A self-training approach based on geometrical constraints, IEEE transactions on aerospace and electronic systems (2023).

\bibitem{jawaid2024iros}
M.~Jawaid, R.~Talak, Y.~Latif, L.~Carlone, T.-J. Chin, Test-time certifiable self-supervision to bridge the sim2real gap in event-based satellite pose estimation, in: IEEE/RSJ International Conference on Intelligent Robots and Systems (IROS), 2024.

\bibitem{eventsensor128}
P.~Lichtsteiner, C.~Posch, T.~Delbruck, A 128$\times$128 120 db 15$\mu$ s latency asynchronous temporal contrast vision sensor, IEEE journal of solid-state circuits 43~(2) (2008) 566--576.

\bibitem{eventsensor640}
B.~Son, Y.~Suh, S.~Kim, H.~Jung, J.-S. Kim, C.~Shin, K.~Park, K.~Lee, J.~Park, J.~Woo, et~al., 4.1 a 640$\times$ 480 dynamic vision sensor with a 9$\mu$m pixel and 300meps address-event representation, in: 2017 IEEE International Solid-State Circuits Conference (ISSCC), IEEE, 2017, pp. 66--67.

\bibitem{jawaid2023towards}
M.~Jawaid, E.~Elms, Y.~Latif, T.-J. Chin, Towards bridging the space domain gap for satellite pose estimation using event sensing, in: 2023 IEEE International Conference on Robotics and Automation (ICRA), IEEE, 2023, pp. 11866--11873.

\bibitem{rathinam2024spades}
A.~Rathinam, H.~Qadadri, D.~Aouada, Spades: A realistic spacecraft pose estimation dataset using event sensing, in: 2024 IEEE International Conference on Robotics and Automation (ICRA), IEEE, 2024, pp. 11760--11766.

\bibitem{yishi2025crossast}
W.~Yishi, M.~Maestrini, Z.~Zexu, M.~Massari, P.~Di~Lizia, Cross-modal fusion of monocular images and neuromorphic streams for 6d pose estimation of non-cooperative targets, Aerospace Science and Technology (2025) 110338.

\bibitem{crossattentionuda}
C.~Xie, W.~Gao, R.~Guo, Cross-modal learning for event-based semantic segmentation via attention soft alignment, IEEE Robotics and Automation Letters 9~(3) (2024) 2359--2366.
\newblock \href {https://doi.org/10.1109/LRA.2024.3355648} {\path{doi:10.1109/LRA.2024.3355648}}.

\bibitem{Yuan_2024_WACV}
D.~Yuan, F.~Maire, F.~Dayoub, Cross-attention between satellite and ground views for enhanced fine-grained robot geo-localization, in: Proceedings of the IEEE/CVF Winter Conference on Applications of Computer Vision (WACV), 2024, pp. 1249--1256.

\bibitem{fischler1981ransac}
M.~A. Fischler, R.~C. Bolles, Random sample consensus: a paradigm for model fitting with applications to image analysis and automated cartography, Communications of the ACM 24~(6) (1981) 381--395.

\bibitem{jawaidnoveldataset25}
M.~Jawaid, M.~Märtens, T.-J. Chin, \href{https://zenodo.org/records/15861758}{Fresh (fusion with rgb and events for spacecraft pose estimation under harsh lighting) dataset} (Jul. 2025).
\newblock \href {https://doi.org/10.1016/j.ast.2025.111039} {\path{doi:10.1016/j.ast.2025.111039}}.
\newline\urlprefix\url{https://zenodo.org/records/15861758}

\bibitem{PAULY2023339}
L.~Pauly, W.~Rharbaoui, C.~Shneider, A.~Rathinam, V.~Gaudillière, D.~Aouada, \href{https://www.sciencedirect.com/science/article/pii/S0094576523003995}{A survey on deep learning-based monocular spacecraft pose estimation: Current state, limitations and prospects}, Acta Astronautica 212 (2023) 339--360.
\newblock \href {https://doi.org/https://doi.org/10.1016/j.actaastro.2023.08.001} {\path{doi:https://doi.org/10.1016/j.actaastro.2023.08.001}}.
\newline\urlprefix\url{https://www.sciencedirect.com/science/article/pii/S0094576523003995}

\bibitem{park2023satellite}
T.~H. Park, M.~Märtens, M.~Jawaid, Z.~Wang, B.~Chen, T.-J. Chin, D.~Izzo, S.~D’Amico, \href{https://www.sciencedirect.com/science/article/pii/S0094576523000048}{Satellite pose estimation competition 2021: Results and analyses}, Acta Astronautica 204 (2023) 640--665.
\newblock \href {https://doi.org/https://doi.org/10.1016/j.actaastro.2023.01.002} {\path{doi:https://doi.org/10.1016/j.actaastro.2023.01.002}}.
\newline\urlprefix\url{https://www.sciencedirect.com/science/article/pii/S0094576523000048}

\bibitem{musallam2022cubesat}
M.~A. Musallam, A.~Rathinam, V.~Gaudilli{\`e}re, M.~O.~d. Castillo, D.~Aouada, Cubesat-cdt: a cross-domain dataset for 6-dof trajectory estimation of a symmetric spacecraft, in: European Conference on Computer Vision, Springer, 2022, pp. 112--126.

\bibitem{Gallet_2024_CVPR}
F.~Gallet, C.~Marabotto, T.~Chambon, Exploring ai-based satellite pose estimation: from novel synthetic dataset to in-depth performance evaluation, in: Proceedings of the IEEE/CVF Conference on Computer Vision and Pattern Recognition (CVPR) Workshops, 2024, pp. 6770--6778.

\bibitem{yu2024comprehensive}
Y.~Yu, Z.~Wang, Z.~Li, Q.~Yu, A comprehensive study on pnp-based pipeline for pose estimation of noncooperative satellite, Acta Astronautica 224 (2024) 486--496.

\bibitem{bechini2024robust}
M.~Bechini, G.~Gu, P.~Lunghi, M.~Lavagna, Robust spacecraft relative pose estimation via cnn-aided line segments detection in monocular images, Acta Astronautica 215 (2024) 20--43.

\bibitem{gallego2020event}
G.~Gallego, T.~Delbr{\"u}ck, G.~Orchard, C.~Bartolozzi, B.~Taba, A.~Censi, S.~Leutenegger, A.~J. Davison, J.~Conradt, K.~Daniilidis, et~al., Event-based vision: A survey, IEEE transactions on pattern analysis and machine intelligence 44~(1) (2020) 154--180.

\bibitem{izzo2023neuromorphic}
D.~Izzo, A.~Hadjiivanov, D.~Dold, G.~Meoni, E.~Blazquez, Neuromorphic computing and sensing in space, in: Artificial Intelligence for Space: AI4SPACE, CRC Press, 2023, pp. 107--159.

\bibitem{chin2019star}
T.-J. Chin, S.~Bagchi, A.~Eriksson, A.~Van~Schaik, Star tracking using an event camera, in: CVPR Workshops, 2019.

\bibitem{ng2022asynchronous}
Y.~Ng, Y.~Latif, T.~Chin, R.~E. Mahony, Asynchronous kalman filter for event-based star tracking, in: ECCV Workshops, 2022.

\bibitem{sofiamcleod2022eccv}
S.~McLeod, G.~Meoni, D.~Izzo, A.~Mergy, D.~Liu, Y.~Latif, I.~Reid, T.-J. Chin, Globally optimal event-based divergence estimation for ventral landing, in: ECCV Workshops, 2022.

\bibitem{azzalini2023on}
L.~J. Azzalini, E.~Blazquez, A.~Hadjiivanov, G.~Meoni, D.~Izzo, On the generation of a synthetic event-based vision dataset for navigation and landing, CoRR abs/2308.00394 (2023).

\bibitem{NICHOLAS2023177}
A.~C. Nicholas, J.~M. Wolf, L.~J. Kordella, T.~T. Finne, C.~M. Brown, S.~A. Budzien, K.~D. Marr, C.~R. Englert, On-orbit optical detection of lethal non-trackable debris, Acta Astronautica 212 (2023) 177--186.
\newblock \href {https://doi.org/https://doi.org/10.1016/j.actaastro.2023.08.002} {\path{doi:https://doi.org/10.1016/j.actaastro.2023.08.002}}.

\bibitem{liu2025stereo}
Z.~Liu, B.~Guan, Y.~Shang, Y.~Bian, P.~Sun, Q.~Yu, Stereo event-based, 6-dof pose tracking for uncooperative spacecraft, IEEE Transactions on Geoscience and Remote Sensing (2025).

\bibitem{malik2025evsat3d}
S.~S. Malik, M.~Moshrefizadeh, O.~Tahri, X.~Bai, E.~Blasch, V.~Sagan, H.~Aliakbarpour, Evsat3d: Satellite pose estimation and 3d reconstruction with event camera, IEEE Access (2025).

\bibitem{singh2023deep}
A.~Singh, K.~Gaurav, Deep learning and data fusion to estimate surface soil moisture from multi-sensor satellite images, Scientific Reports 13~(1) (2023) 2251.

\bibitem{fusionplussatellite2025}
Y.~F. Hestrio, D.~Surya~Candra, R.~Prima~Brahmantara, Y.~Prabowo, A.~Dimara~Sakti, M.~M. Luthfi~Ramadhan, Y.~Loong~Yap, K.~Azizah, M.~Hafizhuddin~Hilman, S.~Phinn, S.~Liang~Lim, W.~Jatmiko, Fusion+: A multi-sensor image fusion for very high-resolution satellite imagery, IEEE Access 13 (2025) 62856--62869.
\newblock \href {https://doi.org/10.1109/ACCESS.2025.3558558} {\path{doi:10.1109/ACCESS.2025.3558558}}.

\bibitem{tarasiewicz2023multitemporal}
T.~Tarasiewicz, J.~Nalepa, R.~A. Farrugia, G.~Valentino, M.~Chen, J.~A. Briffa, M.~Kawulok, Multitemporal and multispectral data fusion for super-resolution of sentinel-2 images, IEEE Transactions on Geoscience and Remote Sensing 61 (2023) 1--19.

\bibitem{galante2016fast}
J.~M. Galante, J.~Van~Eepoel, C.~D'Souza, B.~Patrick, Fast kalman filtering for relative spacecraft position and attitude estimation for the raven iss hosted payload, in: AAS Guidance and Control Conference, no. GSFC-E-DAA-TN29232, 2016.

\bibitem{napolano2023multi}
G.~Napolano, C.~Vela, A.~Nocerino, R.~Opromolla, M.~Grassi, A multi-sensor optical relative navigation system for small satellite servicing, Acta Astronautica 207 (2023) 167--192.

\bibitem{jiang2024uncooperative}
C.~Jiang, P.~Guo, Q.~Hu, C.~Long, D.~Li, Uncooperative spacecraft pose estimation based on intensity and range images fusion, IEEE Transactions on Instrumentation and Measurement (2024).

\bibitem{zhang2019fusion}
Z.~Zhang, R.~Zhao, E.~Liu, K.~Yan, Y.~Ma, Y.~Xu, A fusion method of 1d laser and vision based on depth estimation for pose estimation and reconstruction, Robotics and Autonomous Systems 116 (2019) 181--191.

\bibitem{peng2018pose}
J.~Peng, W.~Xu, B.~Liang, A.-G. Wu, Pose measurement and motion estimation of space non-cooperative targets based on laser radar and stereo-vision fusion, IEEE sensors journal 19~(8) (2018) 3008--3019.

\bibitem{tzschichholz2015relativefusion}
T.~Tzschichholz, T.~Boge, K.~Schilling, Relative pose estimation of satellites using pmd-/ccd-sensor data fusion, Acta Astronautica 109 (2015) 25--33.

\bibitem{irvisible2023}
Z.~Zhang, D.~Zhou, G.~Sun, B.~Zhou, An image fusion algorithm based on iterative wavelet transform for space non-cooperative targets, Proceedings of the Institution of Mechanical Engineers, Part G: Journal of Aerospace Engineering 237~(10) (2023) 2228--2239.
\newblock \href {https://doi.org/10.1177/09544100221143565} {\path{doi:10.1177/09544100221143565}}.

\bibitem{irvisible2024}
Z.~Zhang, D.~Zhou, G.~Sun, Y.~Hu, R.~Deng, Dfti: Dual-branch fusion network based on transformer and inception for space noncooperative objects, IEEE Transactions on Instrumentation and Measurement 73 (2024) 1--11.
\newblock \href {https://doi.org/10.1109/TIM.2024.3403182} {\path{doi:10.1109/TIM.2024.3403182}}.

\bibitem{fusionposition2024}
I.~Hall, J.~Feng, M.~Vasile, Ai-based sensor fusion for spacecraft relative position estimation around asteroids (Oct. 2024).
\newblock \href {https://doi.org/10.5281/zenodo.13885663} {\path{doi:10.5281/zenodo.13885663}}.

\bibitem{irsensordr}
T.~I. Source, Datasheet for on semiconductor python 2000 cmos, \url{https://s1-dl.theimagingsource.com/api/2.5/packages/publication/sensor-onsemi/python2000/c8ce9f23-e551-5b47-932d-239025c2d632/python2000_1.0.en_US.pdf}.

\bibitem{weikersdorfer2014event}
D.~Weikersdorfer, D.~B. Adrian, D.~Cremers, J.~Conradt, Event-based 3d slam with a depth-augmented dynamic vision sensor, in: 2014 IEEE international conference on robotics and automation (ICRA), IEEE, 2014, pp. 359--364.

\bibitem{mueggler2018continuous}
E.~Mueggler, G.~Gallego, H.~Rebecq, D.~Scaramuzza, Continuous-time visual-inertial odometry for event cameras, IEEE Transactions on Robotics 34~(6) (2018) 1425--1440.

\bibitem{vidal2018ultimate}
A.~R. Vidal, H.~Rebecq, T.~Horstschaefer, D.~Scaramuzza, Ultimate slam? combining events, images, and imu for robust visual slam in hdr and high-speed scenarios, IEEE Robotics and Automation Letters 3~(2) (2018) 994--1001.

\bibitem{leutenegger2015keyframe}
S.~Leutenegger, S.~Lynen, M.~Bosse, R.~Siegwart, P.~Furgale, Keyframe-based visual--inertial odometry using nonlinear optimization, The International Journal of Robotics Research 34~(3) (2015) 314--334.

\bibitem{nasa3drepo}
J.-C. NASA, Nasa 3d resources, \url{https://nasa3d.arc.nasa.gov/models}.

\bibitem{cassinimodel}
N.-C. Brian~Kumanchik, Cassini, \url{https://nasa3d.arc.nasa.gov/detail/jpl-cassini}.

\bibitem{sohomodel}
D.~William~Keeter, Solar and heliospheric observatory, \url{https://nasa3d.arc.nasa.gov/detail/soho-2}.

\bibitem{Liu_2024_CVPR}
H.~Liu, S.~Peng, L.~Zhu, Y.~Chang, H.~Zhou, L.~Yan, Seeing motion at nighttime with an event camera, in: Proceedings of the IEEE/CVF Conference on Computer Vision and Pattern Recognition (CVPR), 2024, pp. 25648--25658.

\bibitem{Liang_2023_ICCV}
J.~Liang, Y.~Yang, B.~Li, P.~Duan, Y.~Xu, B.~Shi, Coherent event guided low-light video enhancement, in: Proceedings of the IEEE/CVF International Conference on Computer Vision (ICCV), 2023, pp. 10615--10625.

\bibitem{solarconstantconversion}
P.~Michael, \href{https://dx.doi.org/10.21227/mxr7-p365}{A conversion guide: Solar irradiance and lux illuminance} (2019).
\newblock \href {https://doi.org/10.21227/mxr7-p365} {\path{doi:10.21227/mxr7-p365}}.
\newline\urlprefix\url{https://dx.doi.org/10.21227/mxr7-p365}

\bibitem{mrptcalibration}
J.-C. NASA, The mobile robot programming toolkit, \url{https://github.com/MRPT/mrpt}.

\bibitem{bospec2019}
B.~Chen, J.~Cao, A.~Parra, T.-J. Chin, Satellite pose estimation with deep landmark regression and nonlinear pose refinement, in: ICCV Workshops, 2019.

\bibitem{blender}
\url{https://www.blender.org/} (2025).

\bibitem{hu2021v2e}
Y.~Hu, S.-C. Liu, T.~Delbruck, v2e: From video frames to realistic dvs events, in: CVPR, 2021, pp. 1312--1321.

\bibitem{wu2019detectron2}
Y.~Wu, A.~Kirillov, F.~Massa, W.-Y. Lo, R.~Girshick, Detectron2, \url{https://github.com/facebookresearch/detectron2} (2019).

\bibitem{wang2020hrnet}
J.~Wang, K.~Sun, T.~Cheng, B.~Jiang, C.~Deng, Y.~Zhao, D.~Liu, Y.~Mu, M.~Tan, X.~Wang, et~al., Deep high-resolution representation learning for visual recognition, IEEE transactions on pattern analysis and machine intelligence 43~(10) (2020) 3349--3364.

\bibitem{lepetit2009epnp}
V.~Lepetit, F.~Moreno-Noguer, P.~Fua, Ep n p: An accurate o (n) solution to the p n p problem, International journal of computer vision 81 (2009) 155--166.

\bibitem{muralidharan2022autonomous}
V.~Muralidharan, C.~Martinez, A.~{\v{Z}}inys, M.~Klimavi{\v{c}}ius, M.~Olivares-Mendez, Autonomous control for satellite rendezvous in near-earth orbits, in: 2022 International conference on control, automation and diagnosis (ICCAD), IEEE, 2022, pp. 1--6.

\bibitem{gal2016dropout}
Y.~Gal, Z.~Ghahramani, Dropout as a bayesian approximation: Representing model uncertainty in deep learning, in: international conference on machine learning, PMLR, 2016, pp. 1050--1059.

\bibitem{islam2018samplesize}
M.~R. Islam, Sample size and its role in central limit theorem (clt), Computational and Applied Mathematics Journal 4~(1) (2018) 1--7.

\bibitem{wu2022iouloss}
S.~Wu, J.~Yang, X.~Wang, X.~Li, Iou-balanced loss functions for single-stage object detection, Pattern Recognition Letters 156 (2022) 96--103.

\bibitem{neuripsiou2021}
J.~HE, S.~Erfani, X.~Ma, J.~Bailey, Y.~Chi, X.-S. Hua, \textbackslash alpha-iou: A family of power intersection over union losses for bounding box regression, in: M.~Ranzato, A.~Beygelzimer, Y.~Dauphin, P.~Liang, J.~W. Vaughan (Eds.), Advances in Neural Information Processing Systems, Vol.~34, Curran Associates, Inc., 2021, pp. 20230--20242.

\bibitem{nvidiapowerof2batch}
NVIDIA, Cuda c++ best practices guide, \url{https://docs.nvidia.com/cuda/cuda-c-best-practices-guide/index.html} (2025).

\bibitem{fpgaposeestimation}
J.~Posso, G.~Bois, Y.~Savaria, Real-time spacecraft pose estimation using mixed-precision quantized neural network on cots reconfigurable mpsoc, in: 2024 22nd IEEE Interregional NEWCAS Conference (NEWCAS), 2024.
\newblock \href {https://doi.org/10.1109/NewCAS58973.2024.10666317} {\path{doi:10.1109/NewCAS58973.2024.10666317}}.

\bibitem{rendezvousmissiontime}
A.~Petit, E.~Marchand, K.~Kanani, Vision-based space autonomous rendezvous: A case study, in: 2011 IEEE/RSJ International Conference on Intelligent Robots and Systems, 2011, pp. 619--624.
\newblock \href {https://doi.org/10.1109/IROS.2011.6094568} {\path{doi:10.1109/IROS.2011.6094568}}.

\bibitem{latif2023high}
Y.~Latif, P.~Anastasiou, Y.~Ng, Z.~Prime, T.-F. Lu, M.~Tetlow, R.~Mahony, T.-J. Chin, High frequency, high accuracy pointing onboard nanosats using neuromorphic event sensing and piezoelectric actuation, arXiv preprint arXiv:2309.01361 (2023).

\bibitem{Zou_2021_CVPR}
Y.~Zou, Y.~Zheng, T.~Takatani, Y.~Fu, Learning to reconstruct high speed and high dynamic range videos from events, in: Proceedings of the IEEE/CVF Conference on Computer Vision and Pattern Recognition (CVPR), 2021, pp. 2024--2033.

\bibitem{liu2024deep}
J.~{Liu}, W.~{Sun}, H.~{Yang}, Z.~{Zeng}, C.~{Liu}, J.~{Zheng}, X.~{Liu}, H.~{Rahmani}, N.~{Sebe}, A.~{Mian}, {Deep Learning-Based Object Pose Estimation: A Comprehensive Survey}, arXiv e-prints (2024) arXiv:2405.07801\href {http://arxiv.org/abs/2405.07801} {\path{arXiv:2405.07801}}, \href {https://doi.org/10.48550/arXiv.2405.07801} {\path{doi:10.48550/arXiv.2405.07801}}.

\end{thebibliography}

\clearpage

\begin{appendices}
\section{Labeling process accuracy}\label{sec:labelingerrors}
To determine the accuracy of the Blender labeling process, we labeled 100 samples of the \texttt{roe1} and \texttt{roe2} real(\texttt{lightbox}) sequences from the SHIRT~\cite{park2023adaptiveshirt} dataset. We used SHIRT as it has accurate ground-truth labels obtained using calibrated robotic arms and motion capture systems. For both sequences we compare the SHIRT labels $\{({\phi_i},{\bt_i})\}_{i=1}^{100}$, and equivalent labels from our Blender labeling process $\{({\phi^{blender}_i},{\bt^{blender}_i})\}_{i=1}^{100}$ using the pose metrics similar to eq.~\ref{eq:posemetrics}.
We used eq.~\ref{eq:posemetricslabeling} to determine the translation error $\omega_{i}$ (in mm) and rotation error $\theta_{i}$ (in degrees) per sample of our pose labeling technique where $\bt$ is the translation component and $\phi$ is the rotation component expressed as a quaternion.
\begin{equation}
\omega_{i} = \left\| \bt_{i} - \bt^{blender}_{i} \right\|_2,\;\; and \;\;
\theta_{i} = 2\arccos \left( | \langle \phi_{i}, \phi^{blender}_{i} \rangle | \right),
\label{eq:posemetricslabeling}
\end{equation}
We then computed the mean and standard deviation of $\omega$ and $\theta$ across the 100 samples for the two individual sequences as well as for both sequences together indicated by the concatenation ($[]$) operator. We report the labeling errors in Table~\ref{tab:labelingaccuracy}.

\begin{table}[H]\centering
{\footnotesize
\begin{NiceTabular}{cllll}[hvlines,rules/color=[gray]{0.3}]
 & mean $\omega$ (mm) & std $\omega$ (mm) & mean $\theta$ ($\circ$) & std $\theta$ ($\circ$) \\
\texttt{roe1} & 2.24 & 1.09 & 3.13 & 2.98 \\
\texttt{roe2} & 2.47 & 1.74 & 2.03 & 1.36 \\
\texttt{[roe1 roe2]} & 2.35 & 1.45 & 2.58 & 2.37 \\
\end{NiceTabular}}
\caption{Quantitative error of the Blender labeling process using the SHIRT dataset.}
\label{tab:labelingaccuracy}
\end{table}

The labeling errors are within the $\rho=10mm$ and $\sigma=10^{\circ}$ success rate thresholds. We anticipate that the labeling accuracy for the satellite models in the FRESH dataset will be comparable, as the motion trajectories are less complex compared to those in SHIRT. Additionally, the event data offers an extra source of information to support labeling, particularly for low-contrast RGB images.

\section{Error plots of all sequences}\label{sec:appendix}
\begin{figure}[H]
    \centering
    \includegraphics[width=0.95\linewidth]{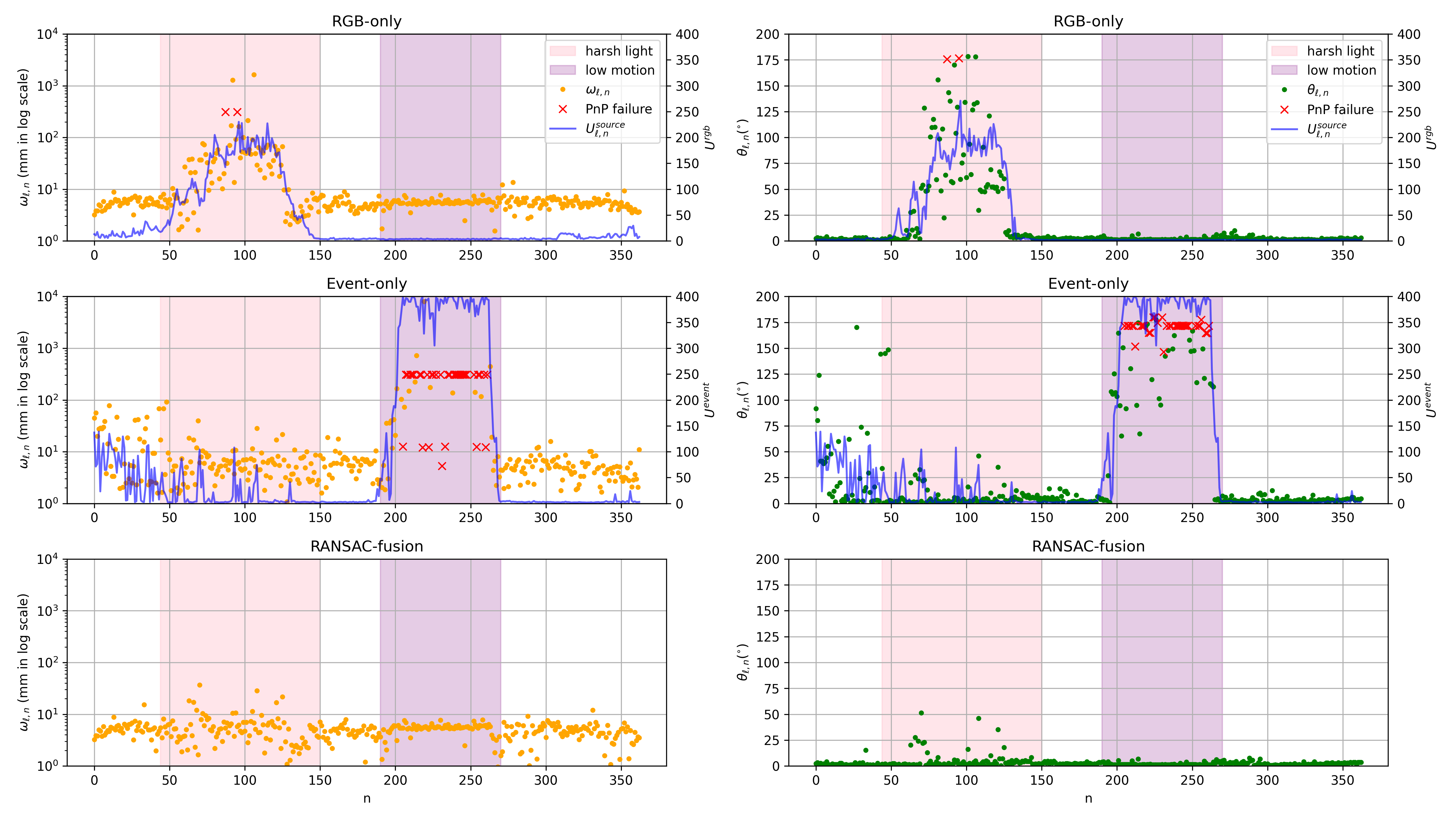}
    \caption{Error and uncertainty plots of the sequence \texttt{satty-1-close}.}
    \label{fig:appendixsatty1}
\end{figure}

\begin{figure}[H]
    \centering
    \includegraphics[width=0.95\linewidth]{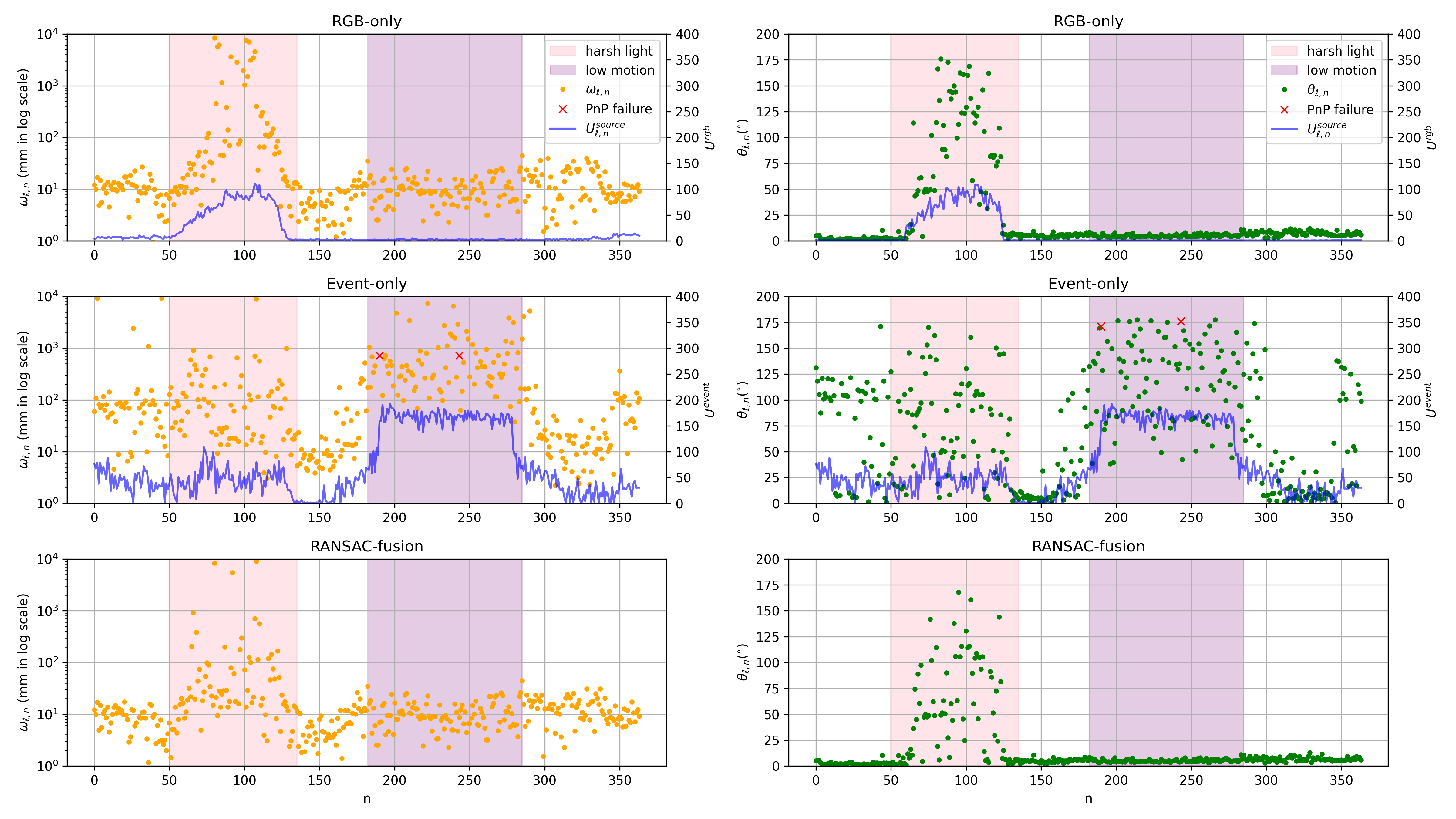}
    \caption{Error and uncertainty plots of the sequence \texttt{satty-1-far}.}
    \label{fig:appendixsatty2}
\end{figure}

\begin{figure}[H]
    \centering
    \includegraphics[width=0.95\linewidth]{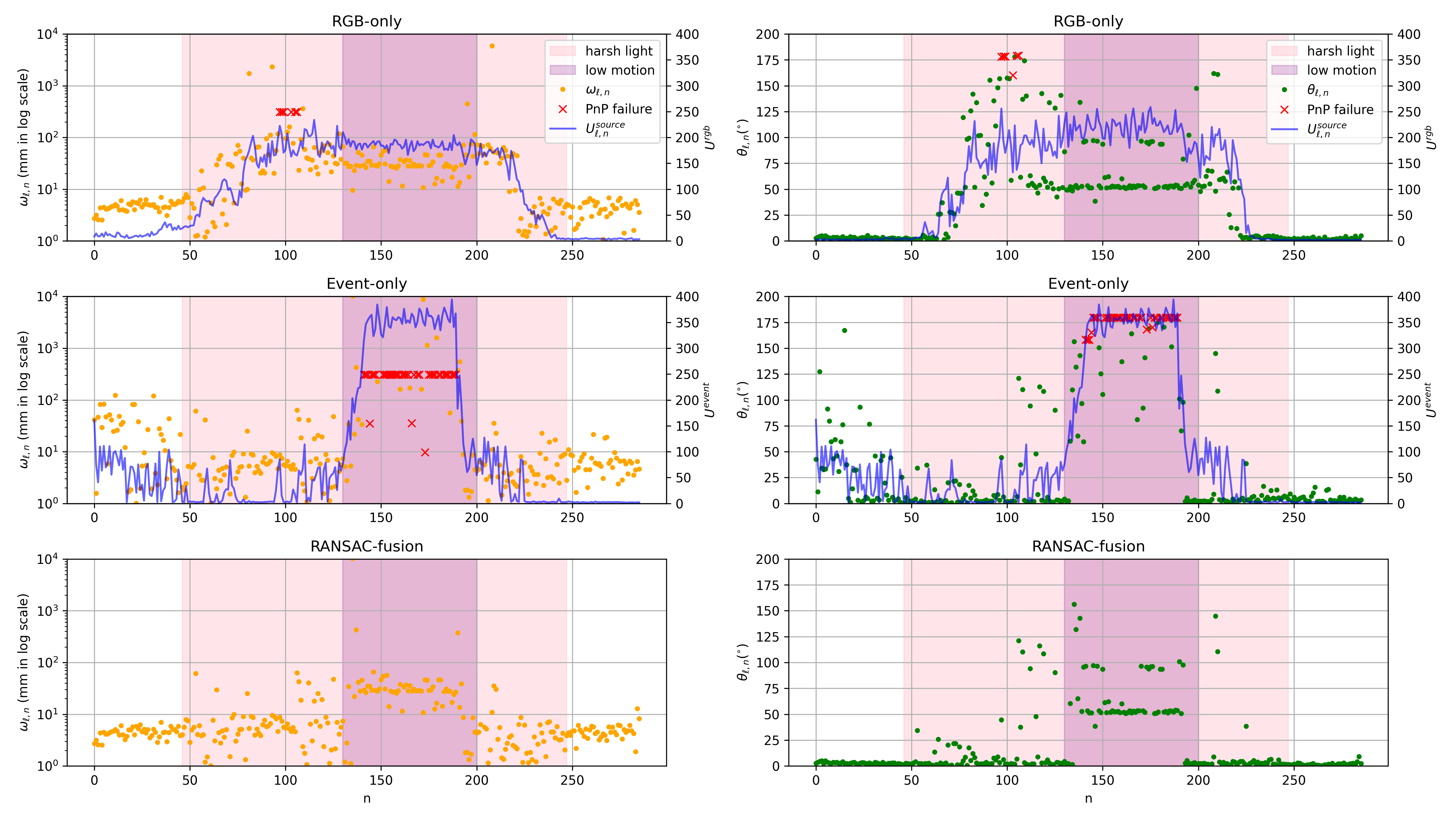}
    \caption{Error and uncertainty plots of the sequence \texttt{satty-2-close}.}
    \label{fig:appendixsatty3}
\end{figure}

\begin{figure}[H]
    \centering
    \includegraphics[width=0.95\linewidth]{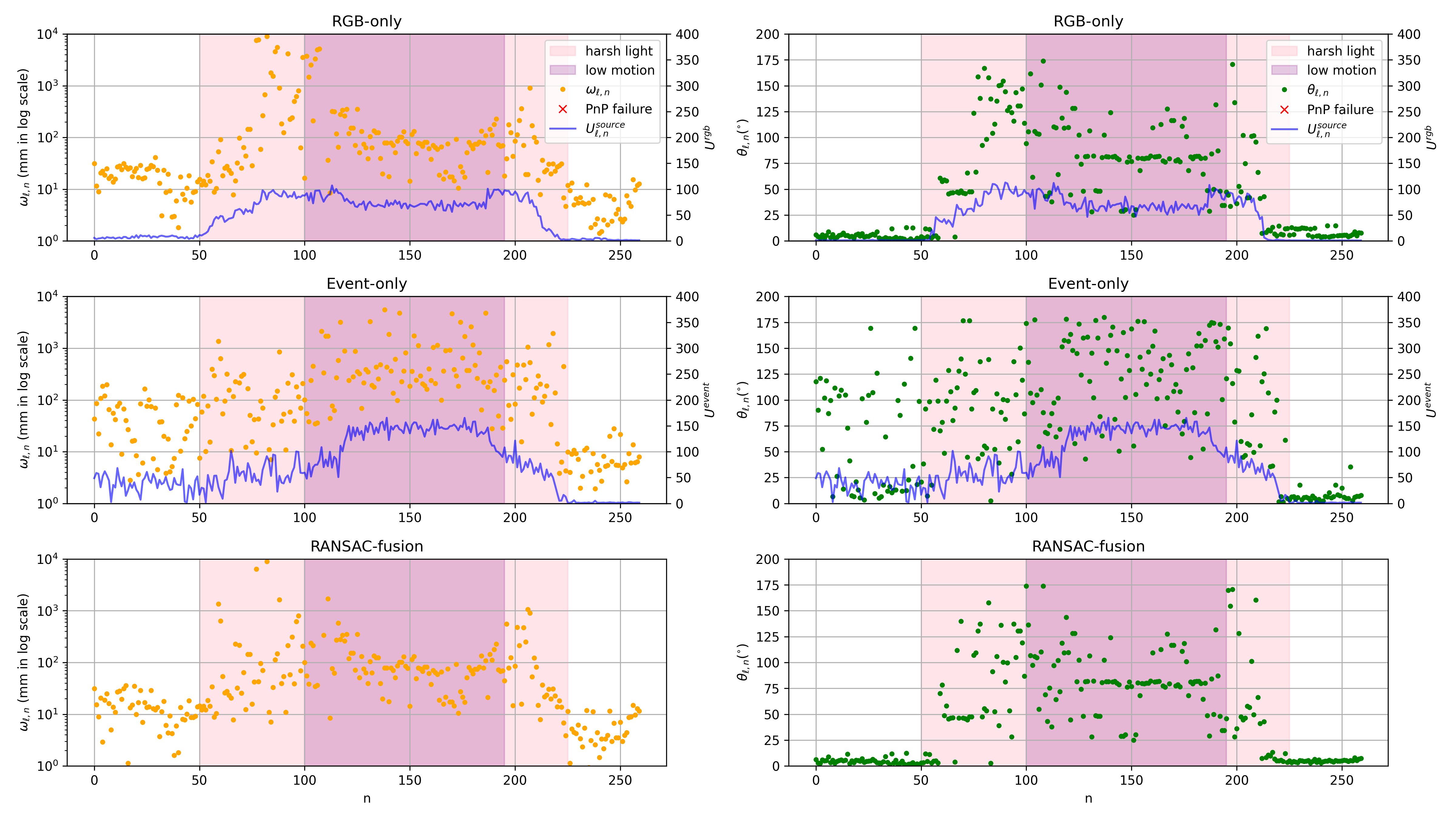}
    \caption{Error and uncertainty plots of the sequence \texttt{satty-2-far}.}
    \label{fig:appendixsatty4}
\end{figure}

\begin{figure}[H]
    \centering
    \includegraphics[width=0.95\linewidth]{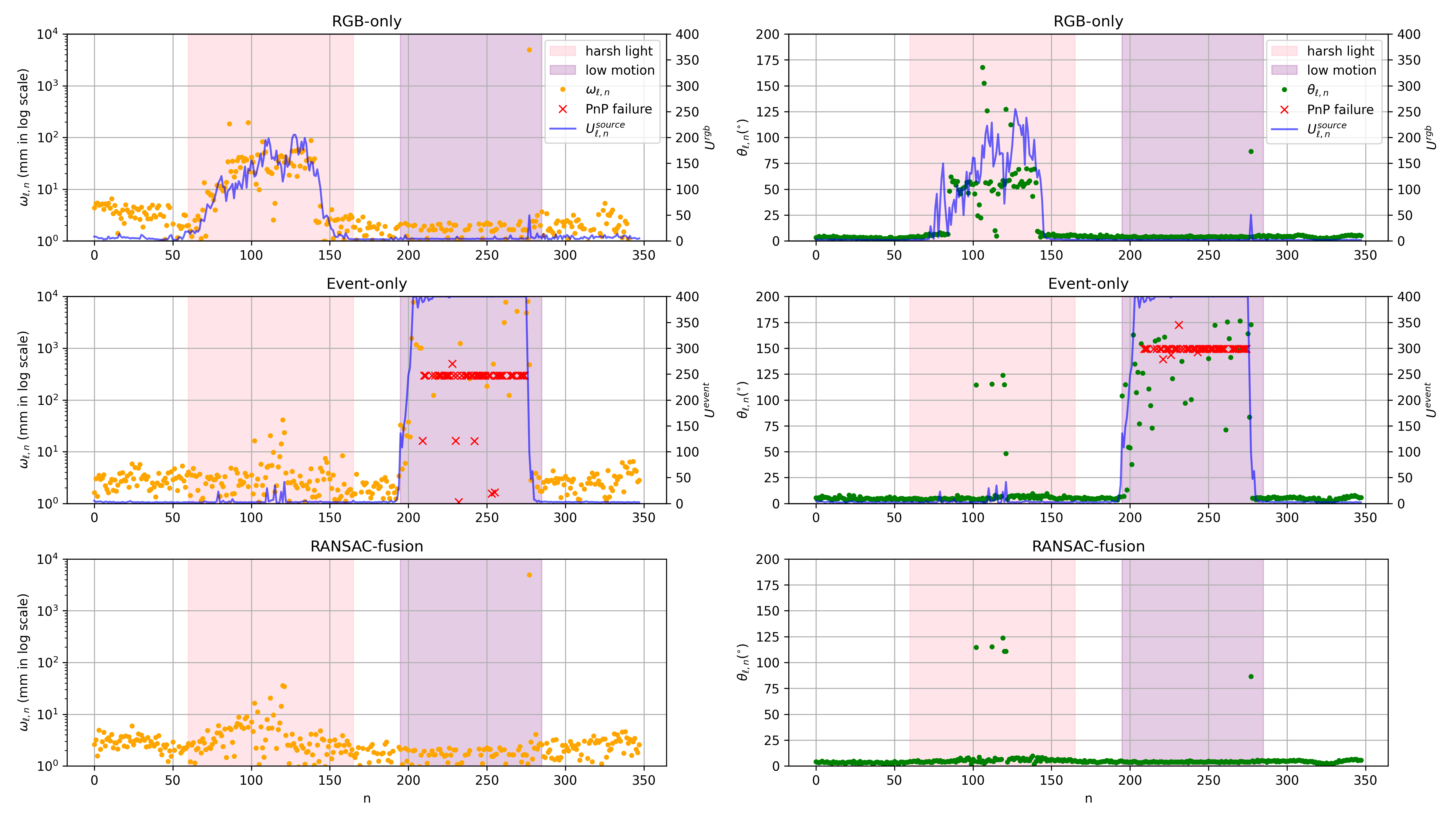}
    \caption{Error and uncertainty plots of the sequence \texttt{satty-3-close}.}
    \label{fig:appendixsatty5}
\end{figure}

\begin{figure}[H]
    \centering
    \includegraphics[width=0.95\linewidth]{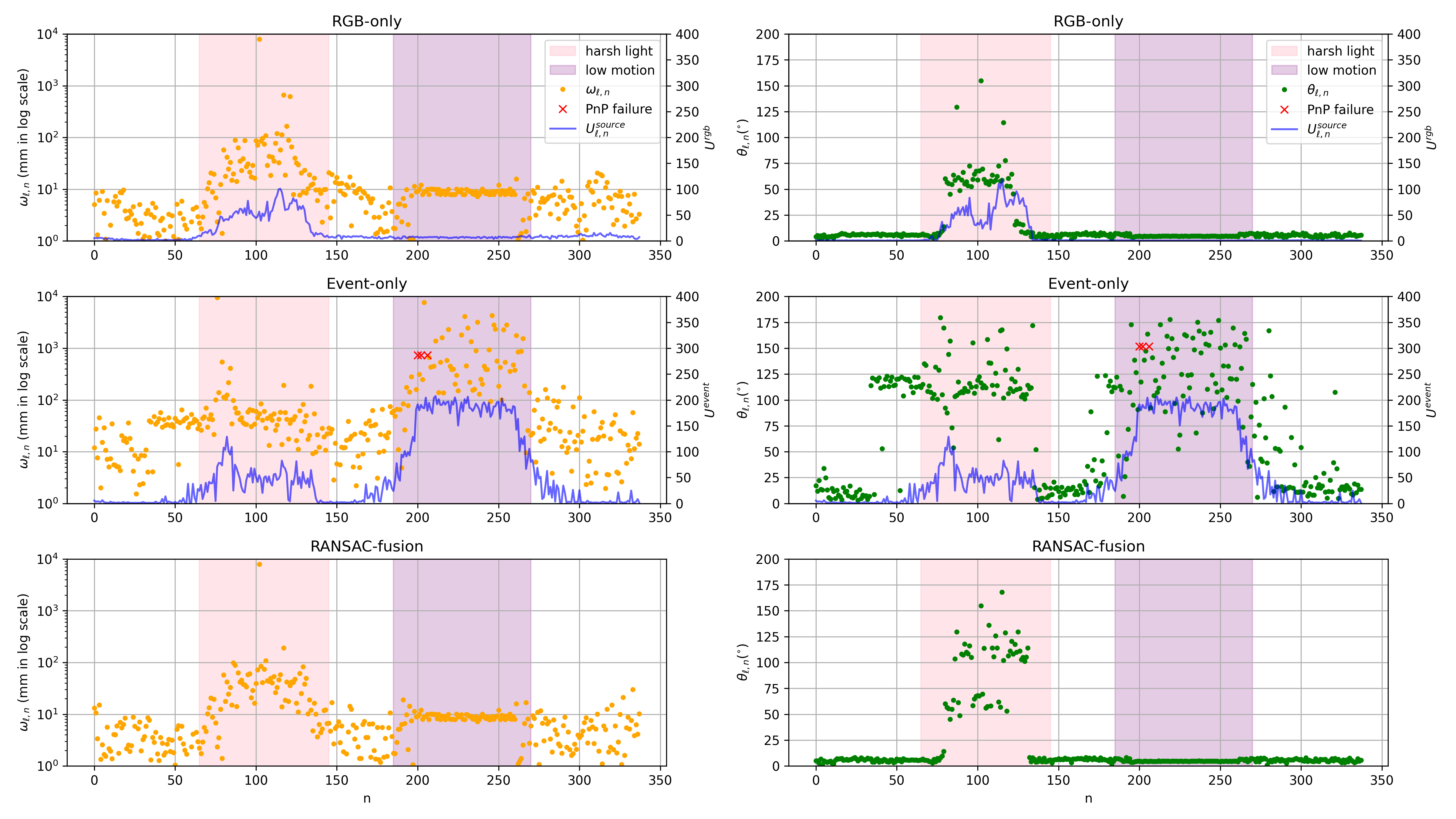}
    \caption{Error and uncertainty plots of the sequence \texttt{satty-3-far}.}
    \label{fig:appendixsatty6}
\end{figure}

\begin{figure}[H]
    \centering
    \includegraphics[width=0.95\linewidth]{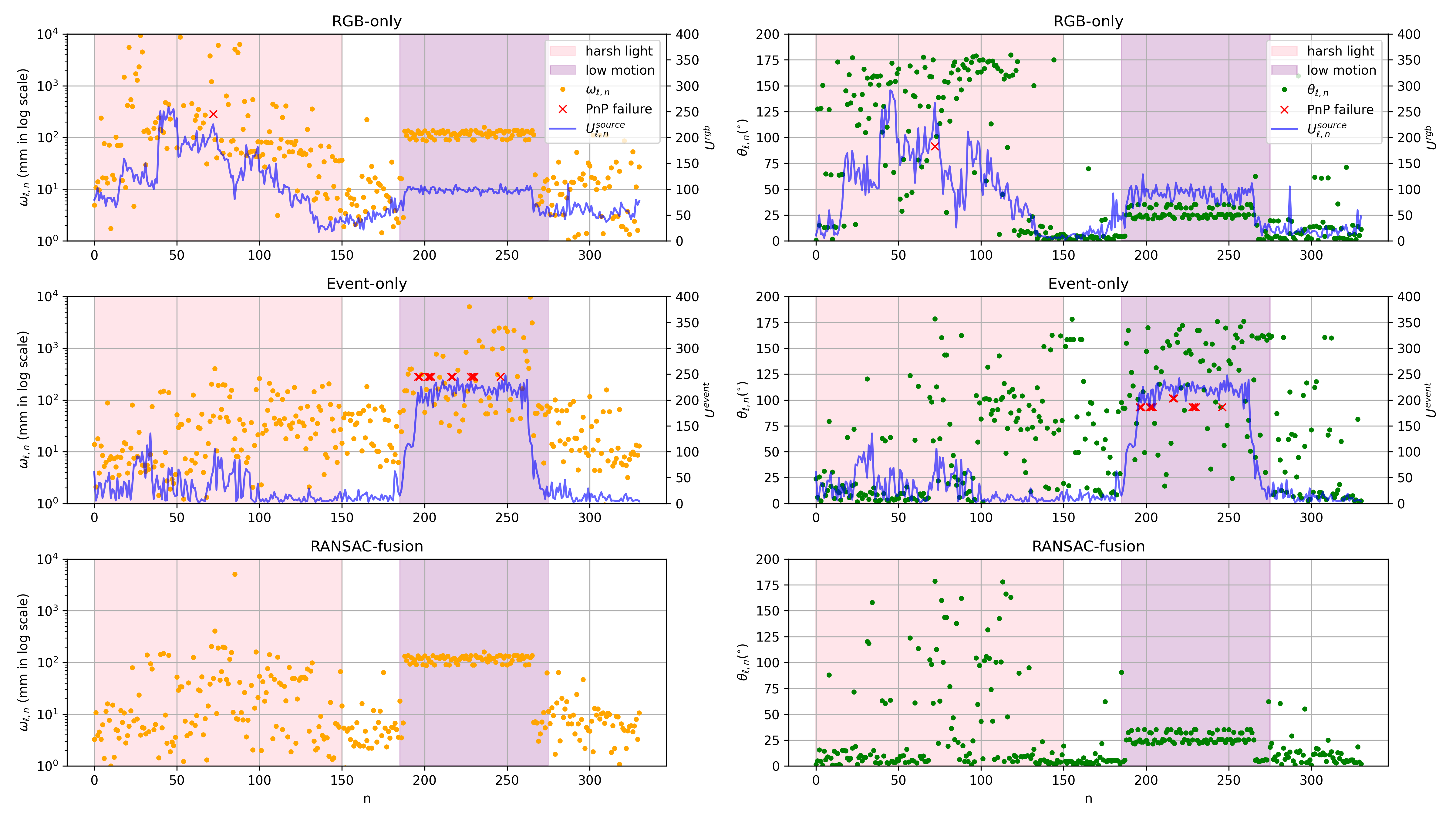}
    \caption{Error and uncertainty plots of the sequence \texttt{satty-4-close}.}
    \label{fig:appendixsatty7}
\end{figure}

\begin{figure}[H]
    \centering
    \includegraphics[width=0.95\linewidth]{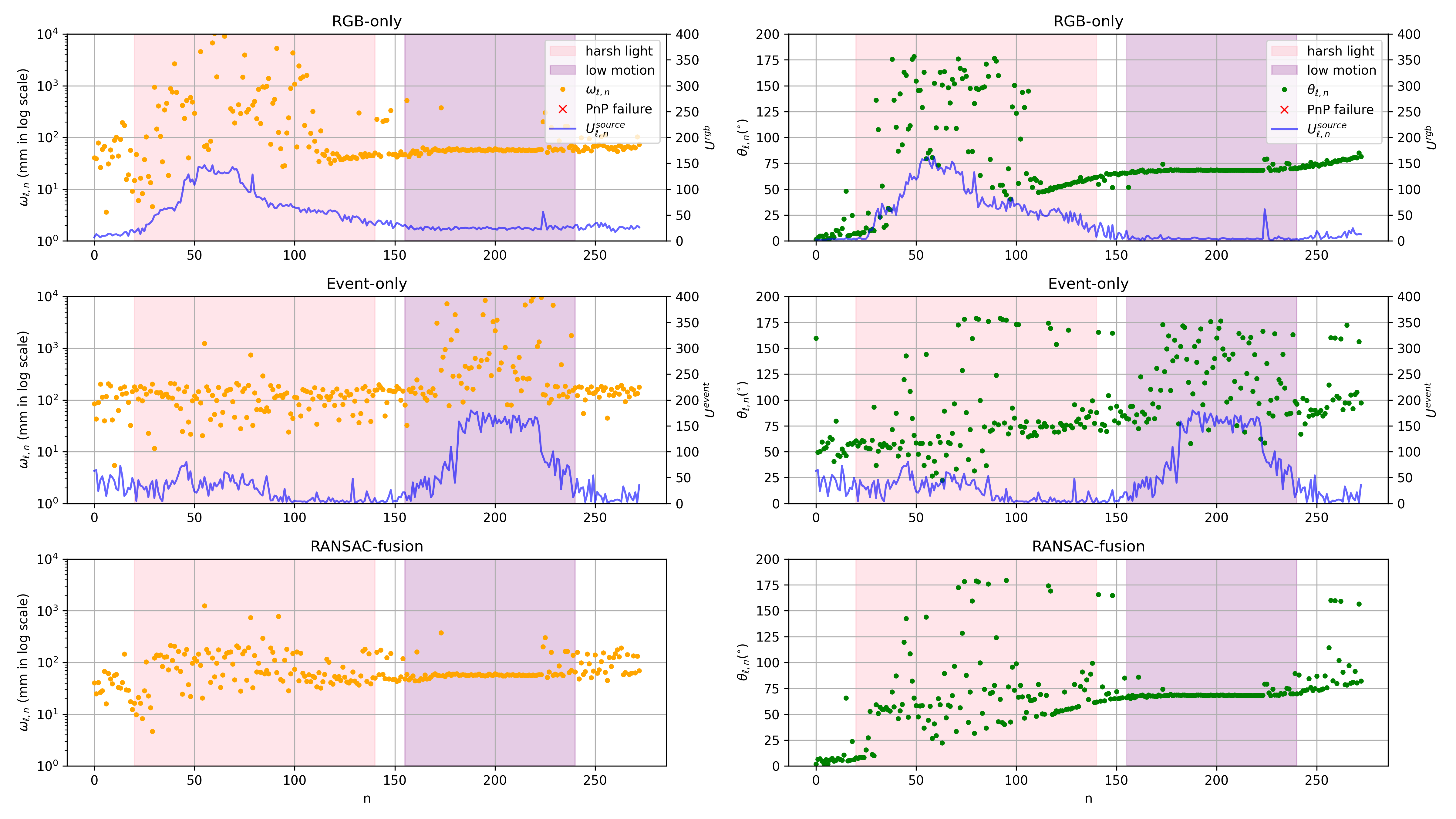}
    \caption{Error and uncertainty plots of the sequence \texttt{satty-4-far}.}
    \label{fig:appendixsatty8}
\end{figure}

\begin{figure}[H]
    \centering
    \includegraphics[width=0.95\linewidth]{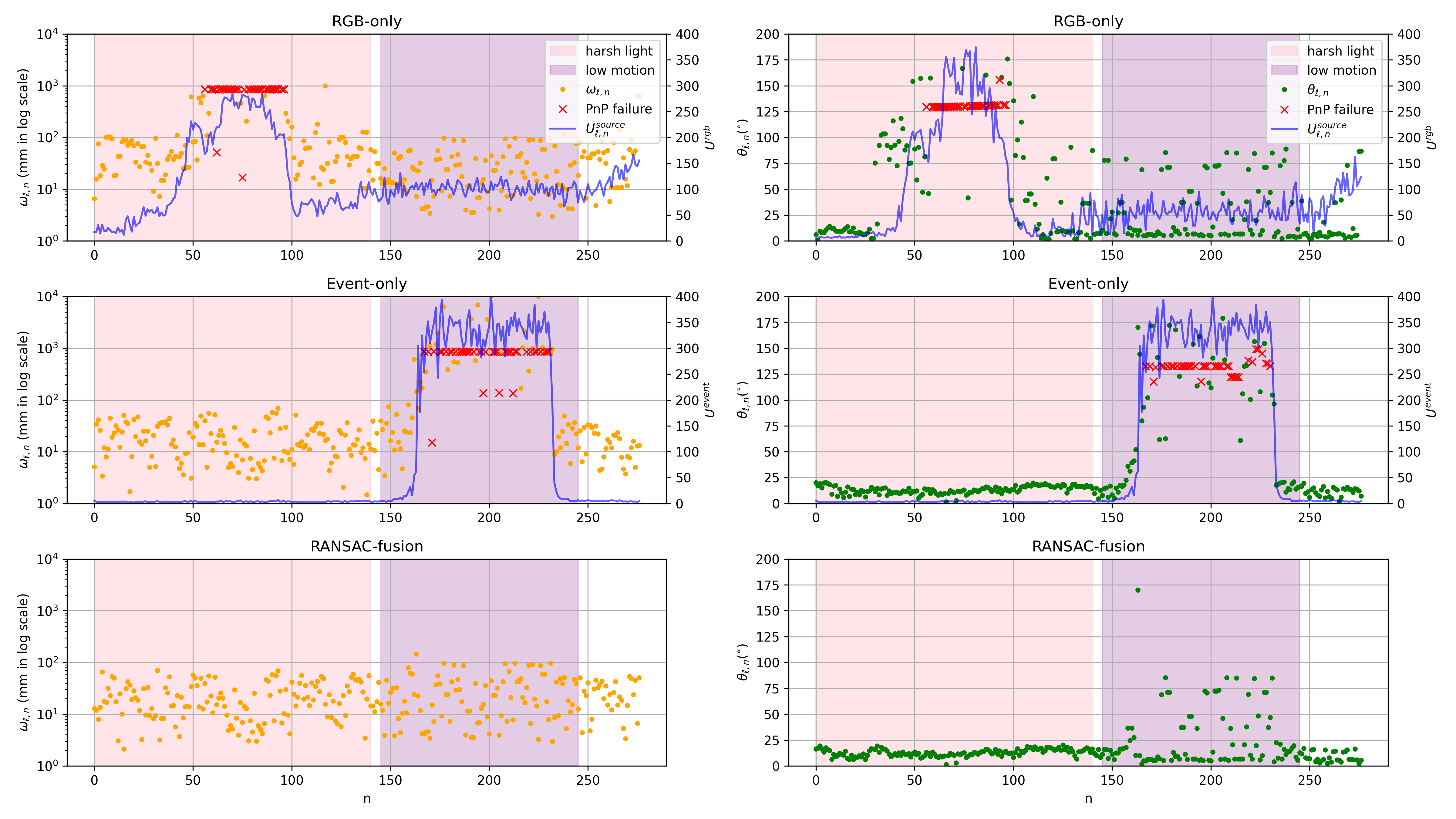}
    \caption{Error and uncertainty plots of the sequence \texttt{cassini-1-close}.}
    \label{fig:appendixcassini1}
\end{figure}

\begin{figure}[H]
    \centering
    \includegraphics[width=0.95\linewidth]{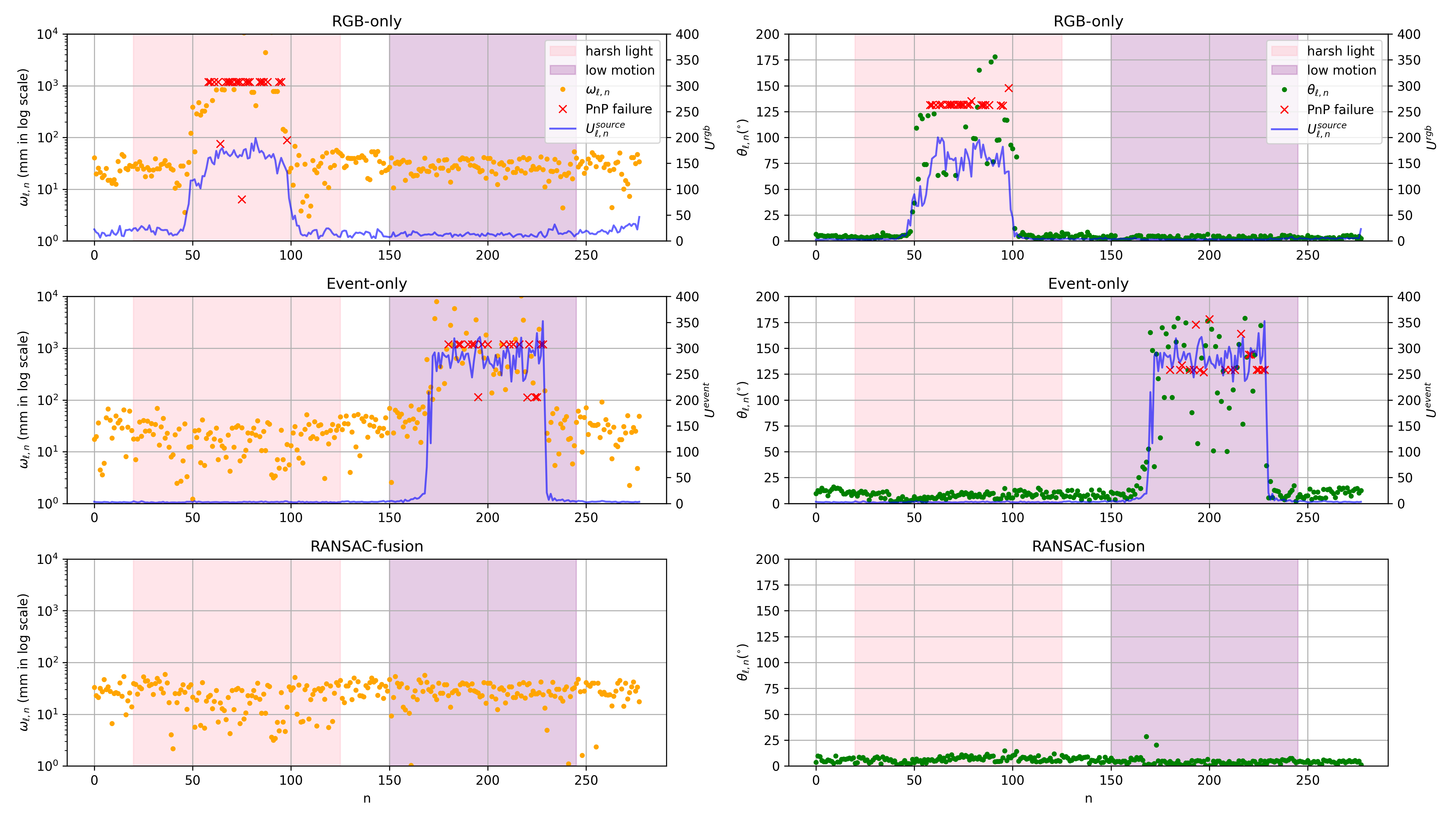}
    \caption{Error and uncertainty plots of the sequence \texttt{cassini-1-far}.}
    \label{fig:appendixcassini2}
\end{figure}

\begin{figure}[H]
    \centering
    \includegraphics[width=0.95\linewidth]{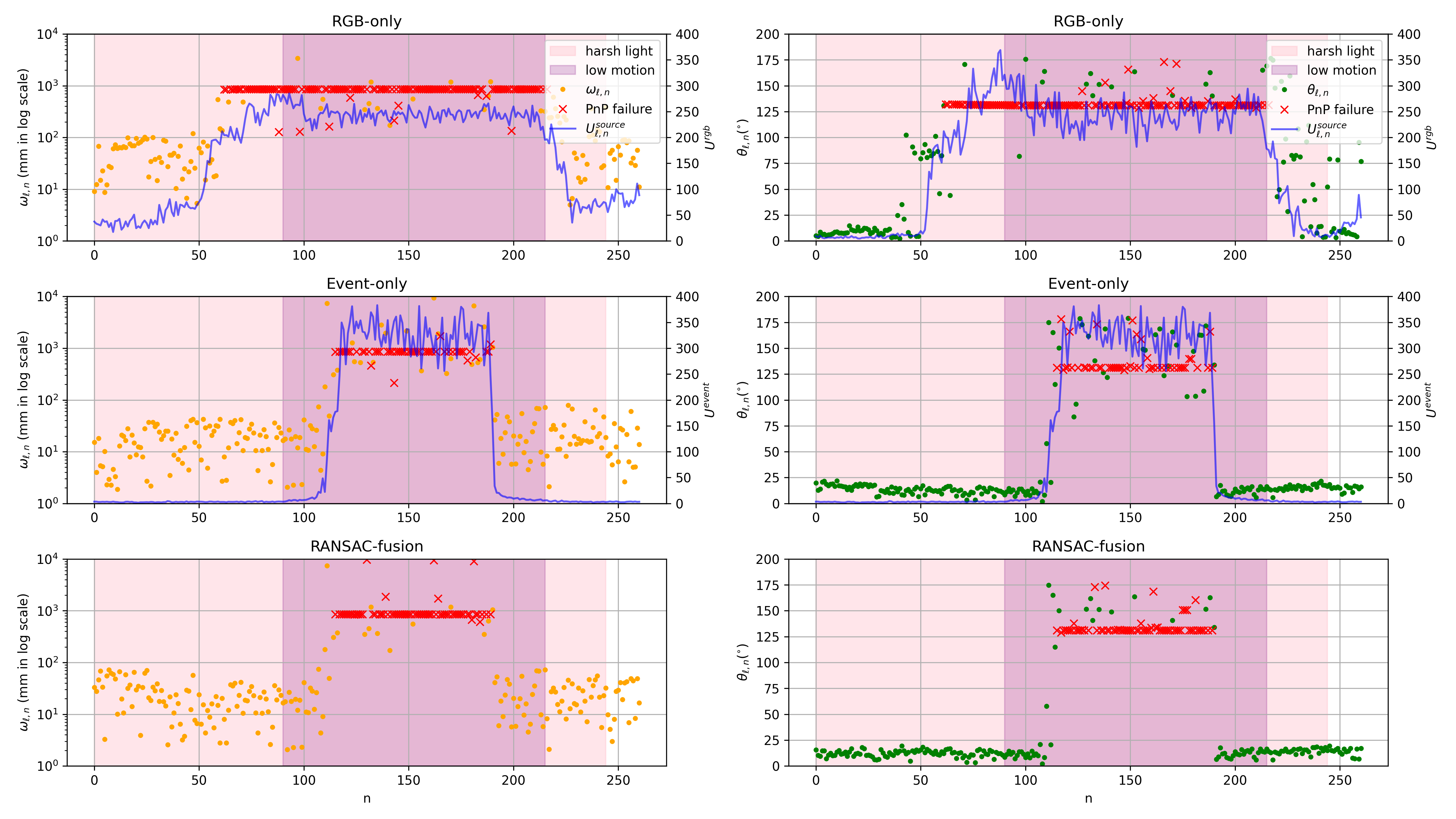}
    \caption{Error and uncertainty plots of the sequence \texttt{cassini-2-close}.}
    \label{fig:appendixcassini3}
\end{figure}

\begin{figure}[H]
    \centering
    \includegraphics[width=0.95\linewidth]{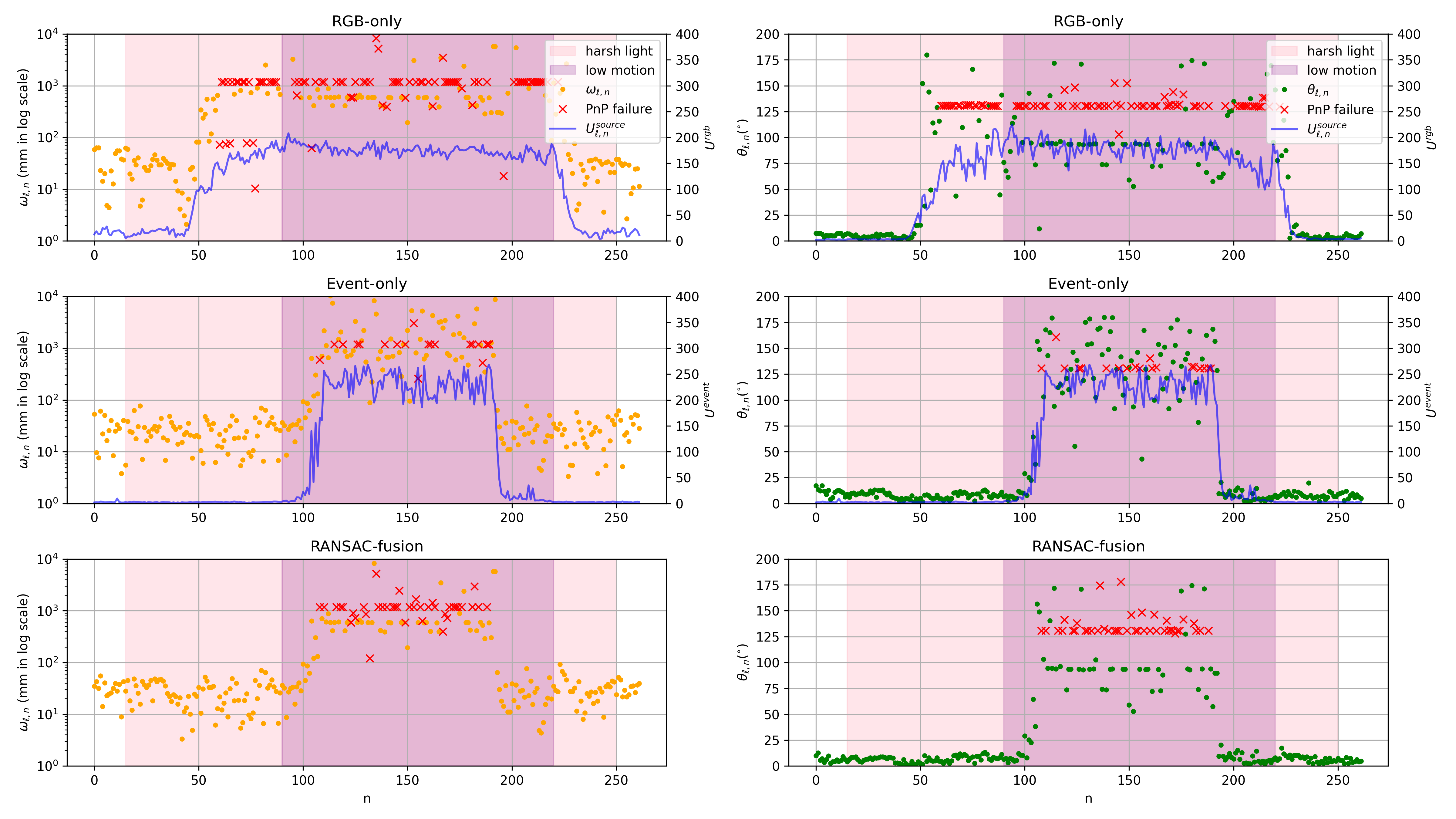}
    \caption{Error and uncertainty plots of the sequence \texttt{cassini-2-far}.}
    \label{fig:appendixcassini4}
\end{figure}

\begin{figure}[H]
    \centering
    \includegraphics[width=0.95\linewidth]{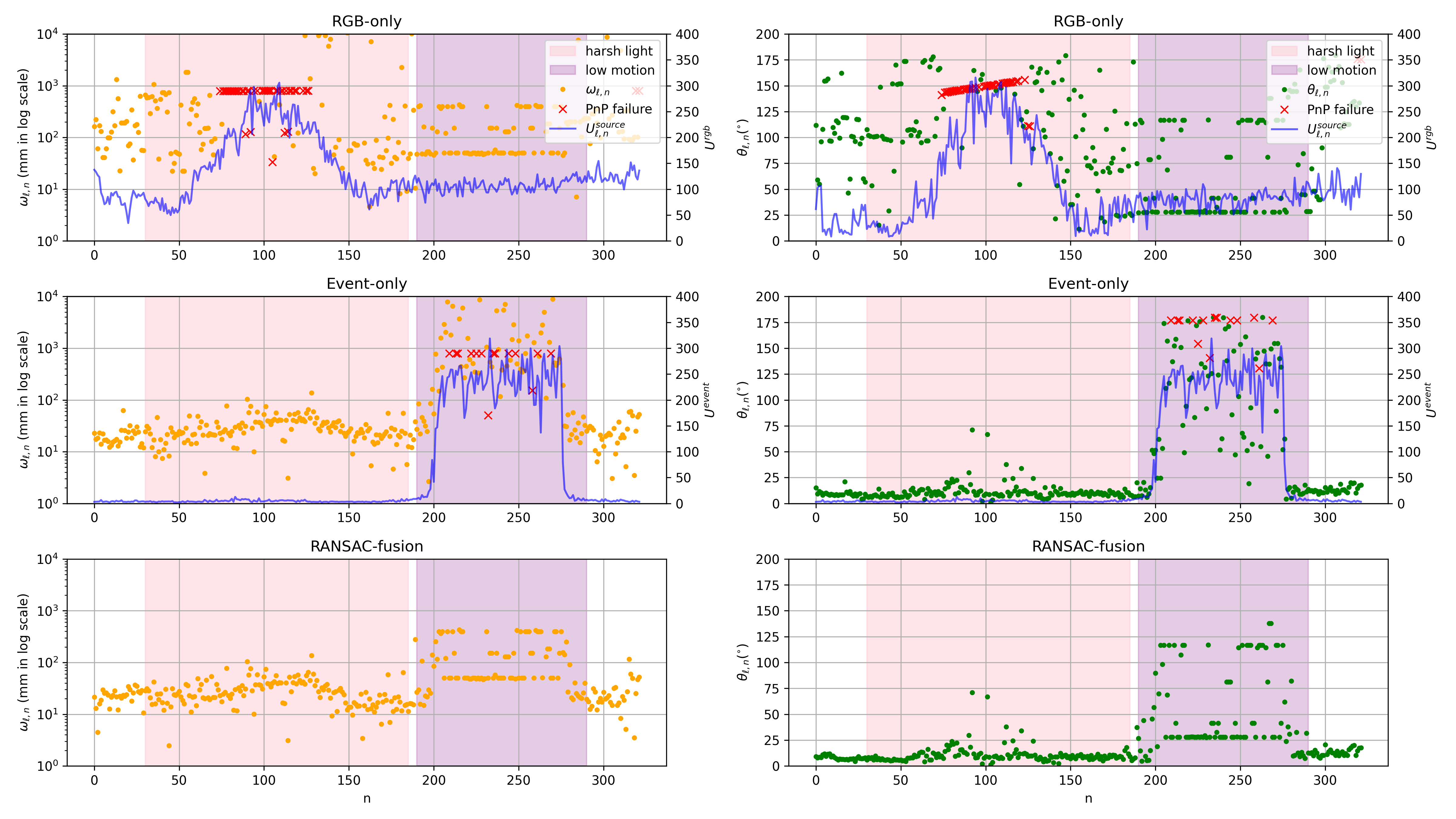}
    \caption{Error and uncertainty plots of the sequence \texttt{cassini-3-close}.}
    \label{fig:appendixcassini5}
\end{figure}

\begin{figure}[H]
    \centering
    \includegraphics[width=0.95\linewidth]{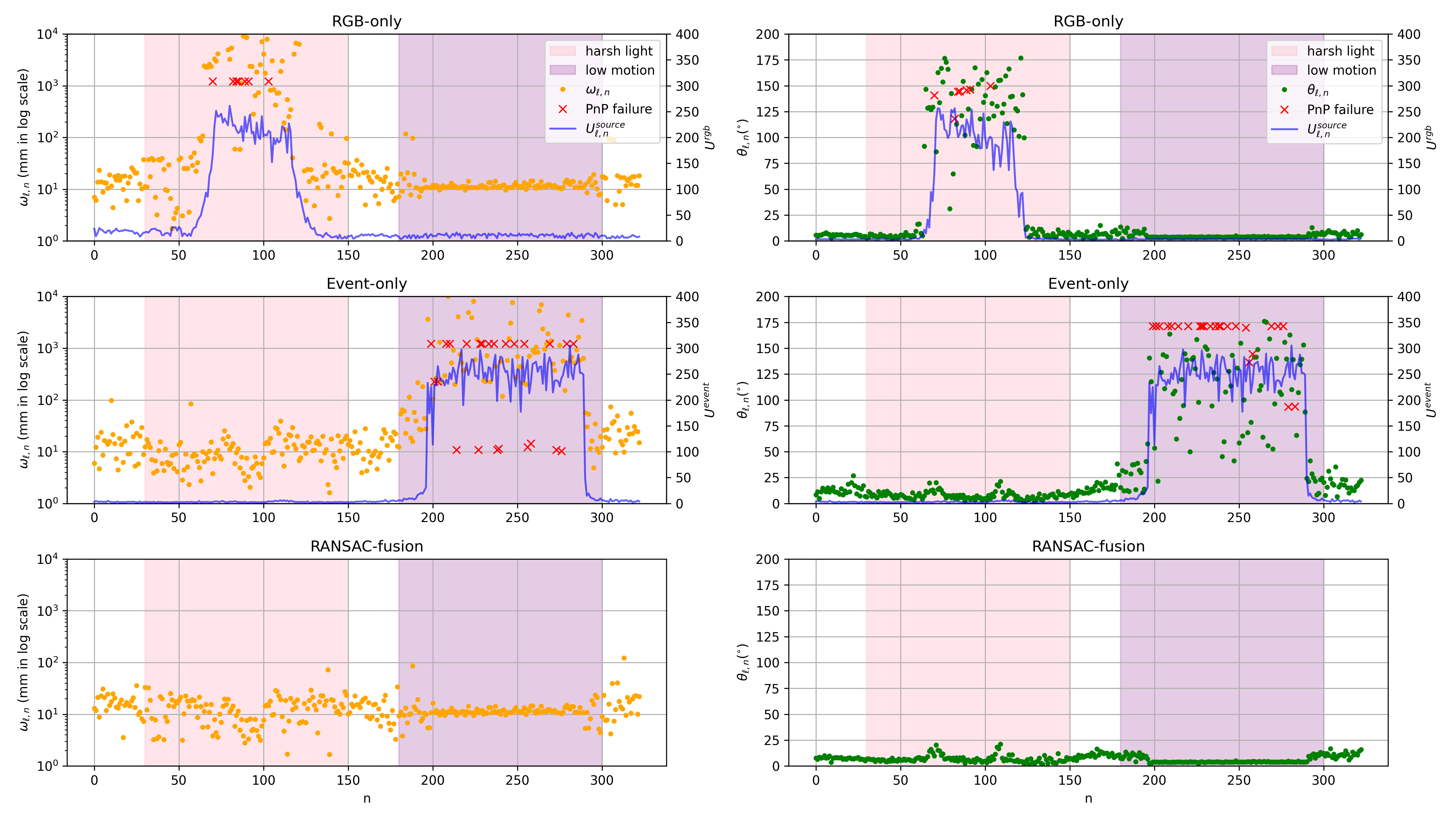}
    \caption{Error and uncertainty plots of the sequence \texttt{cassini-3-far}.}
    \label{fig:appendixcassini6}
\end{figure}

\begin{figure}[H]
    \centering
    \includegraphics[width=0.95\linewidth]{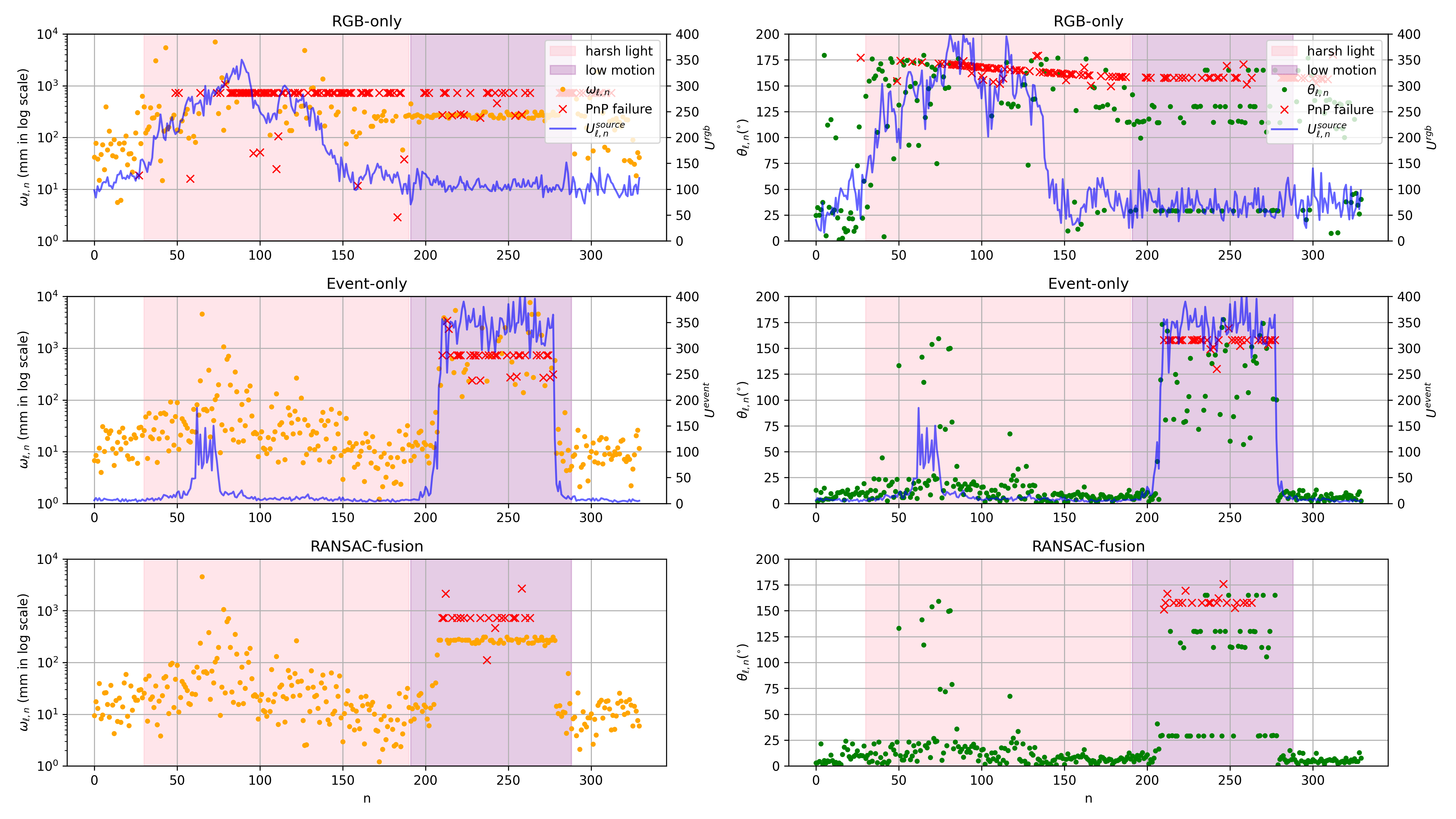}
    \caption{Error and uncertainty plots of the sequence \texttt{cassini-4-close}.}
    \label{fig:appendixcassini7}
\end{figure}

\begin{figure}[H]
    \centering
    \includegraphics[width=0.95\linewidth]{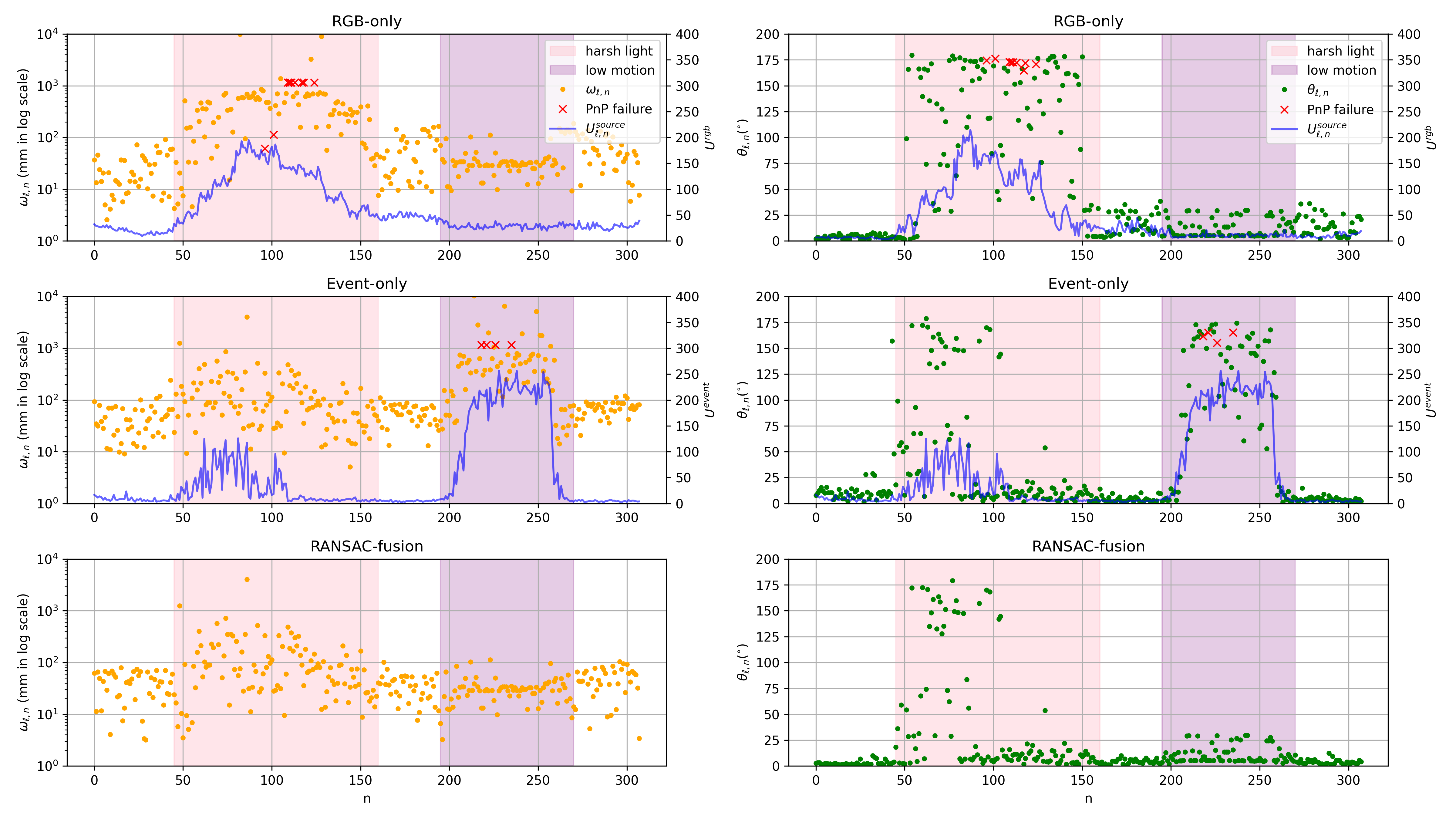}
    \caption{Error and uncertainty plots of the sequence \texttt{cassini-4-far}.}
    \label{fig:appendixcassini8}
\end{figure}

\begin{figure}[H]
    \centering
    \includegraphics[width=0.95\linewidth]{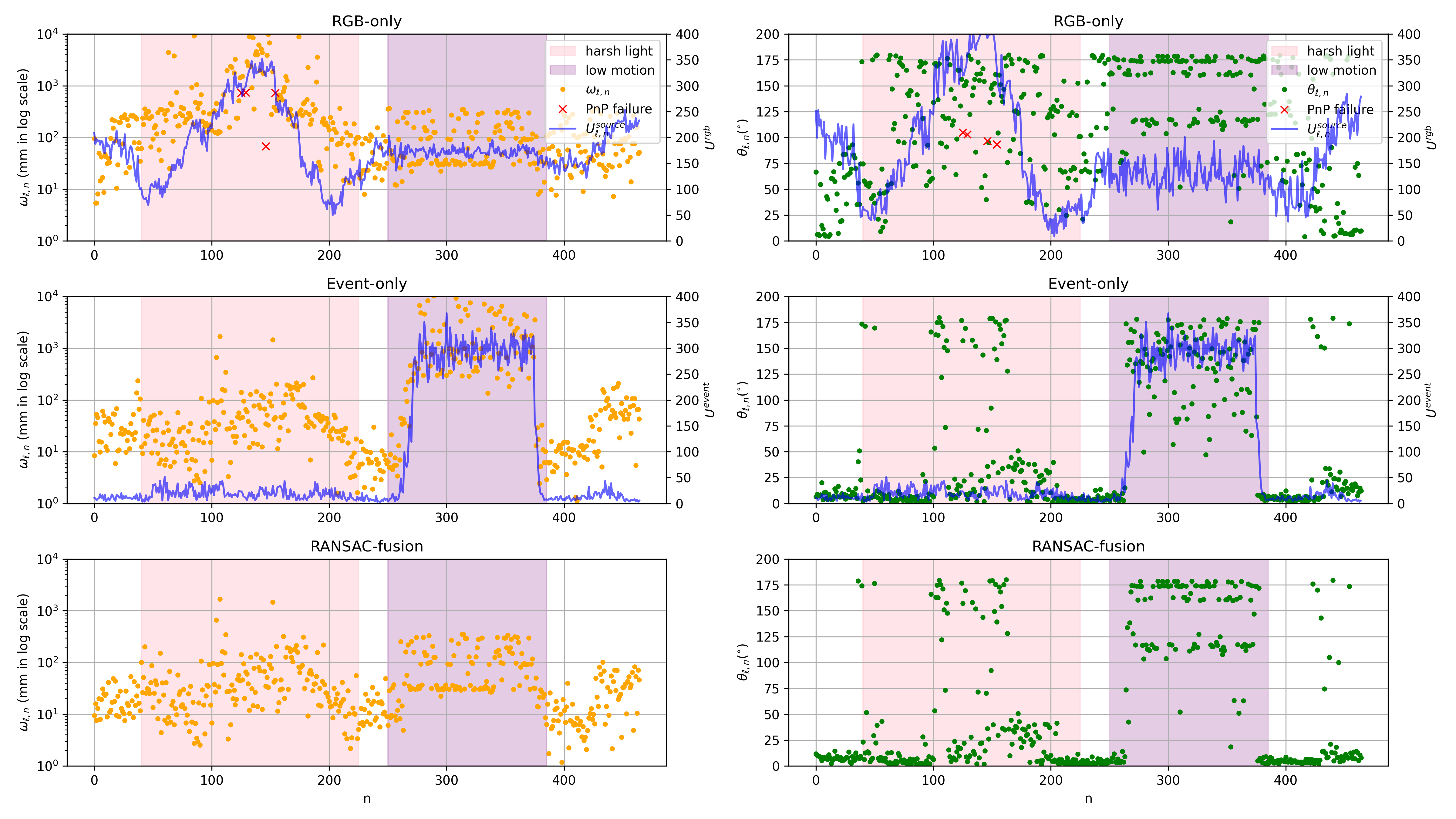}
    \caption{Error and uncertainty plots of the sequence \texttt{soho-1-close}.}
    \label{fig:appendixsoho1}
\end{figure}

We provide the plots of per-frame errors $\omega_{\ell,n}$ and $\theta_{\ell,n}$ for all sequences in our dataset in Figs.~\ref{fig:appendixsatty1}-\ref{fig:appendixsoho1}.

\begin{figure}[H]
    \centering
    \includegraphics[width=0.95\linewidth]{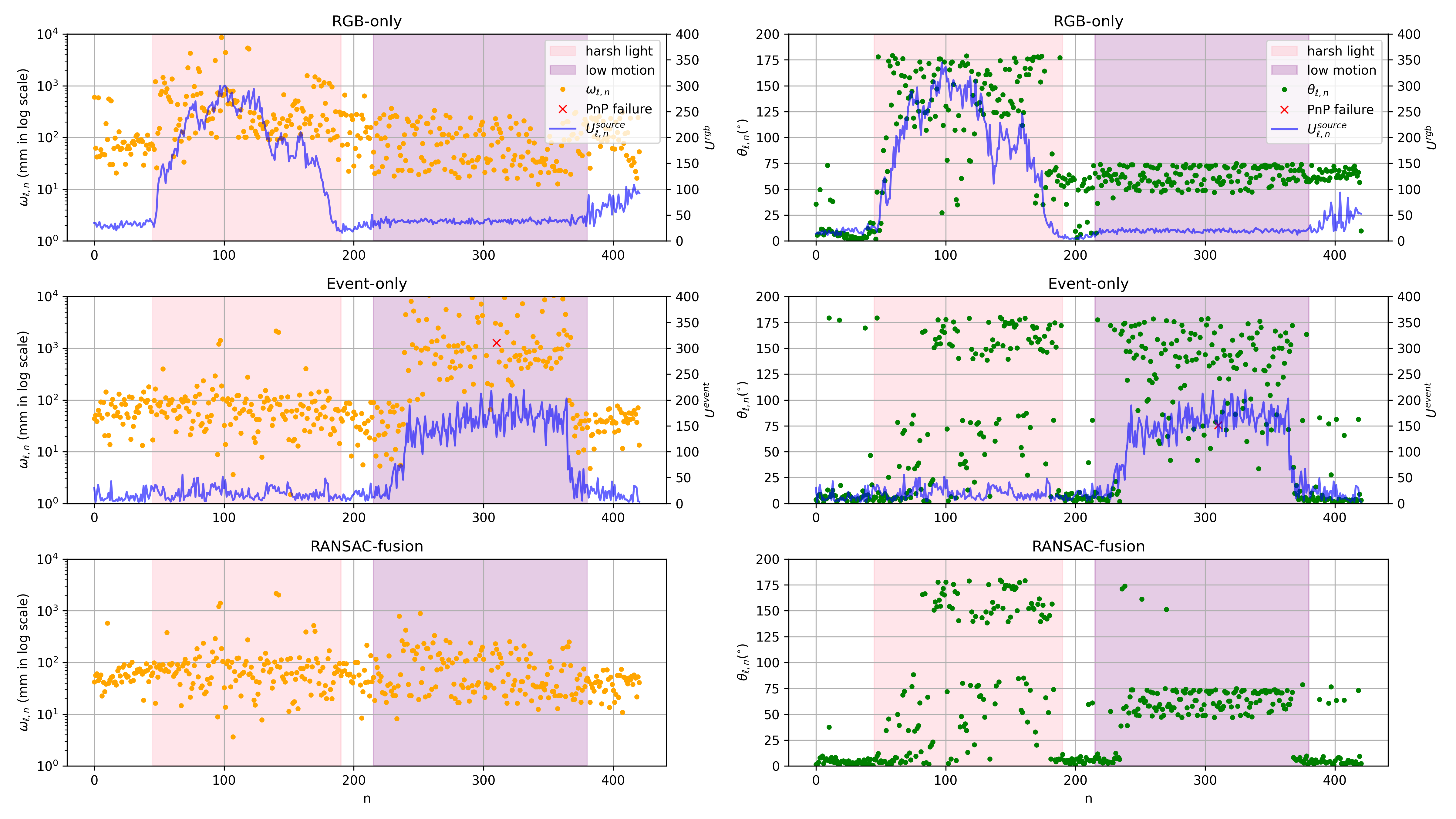}
    \caption{Error and uncertainty plots of the sequence \texttt{soho-1-far}.}
    \label{fig:appendixsoho2}
\end{figure}

\begin{figure}[H]
    \centering
    \includegraphics[width=0.95\linewidth]{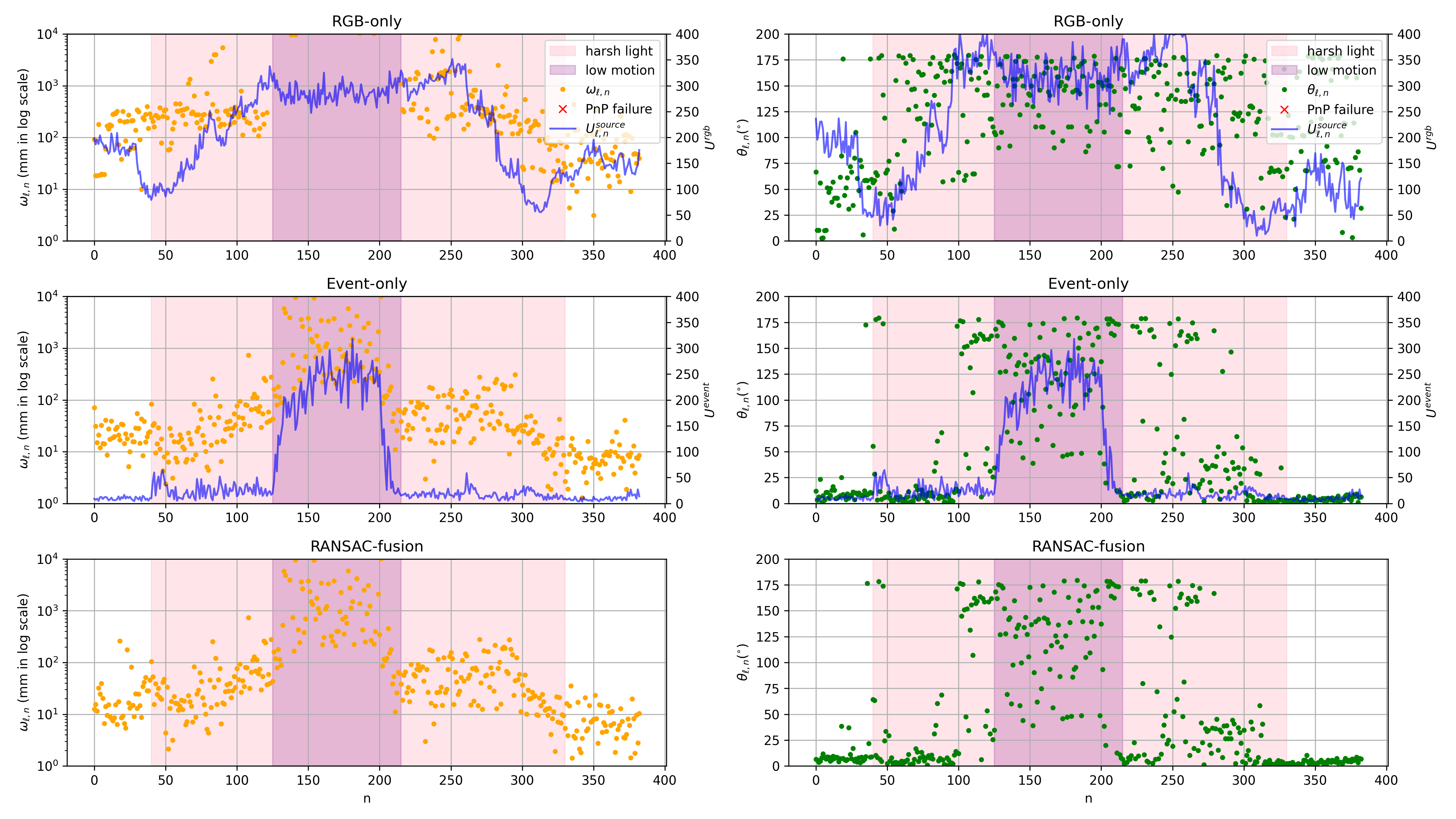}
    \caption{Error and uncertainty plots of the sequence \texttt{soho-2-close}.}
    \label{fig:appendixsoho3}
\end{figure}

\begin{figure}[H]
    \centering
    \includegraphics[width=0.95\linewidth]{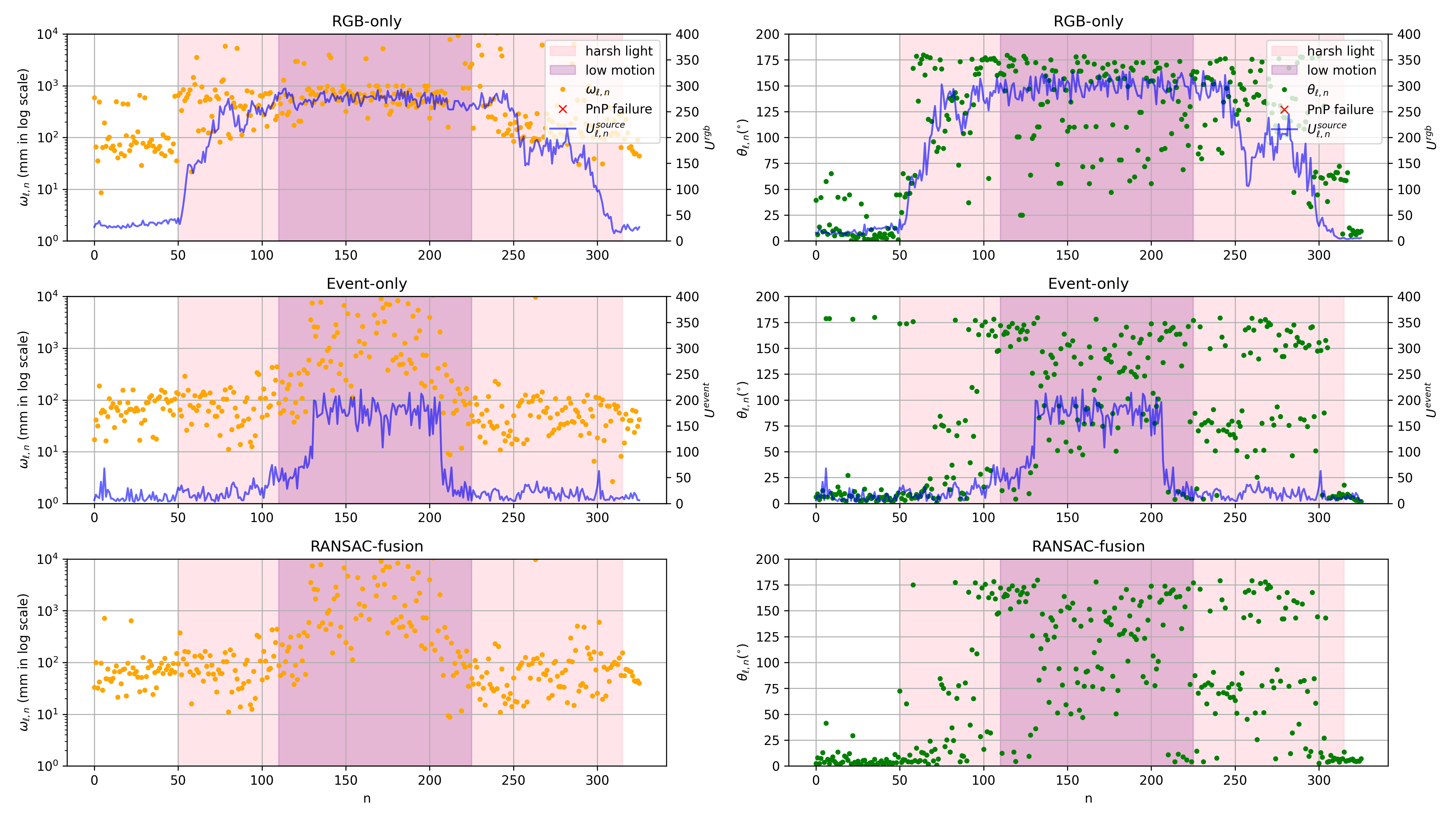}
    \caption{Error and uncertainty plots of the sequence \texttt{soho-2-far}.}
    \label{fig:appendixsoho4}
\end{figure}

\begin{figure}[H]
    \centering
    \includegraphics[width=0.95\linewidth]{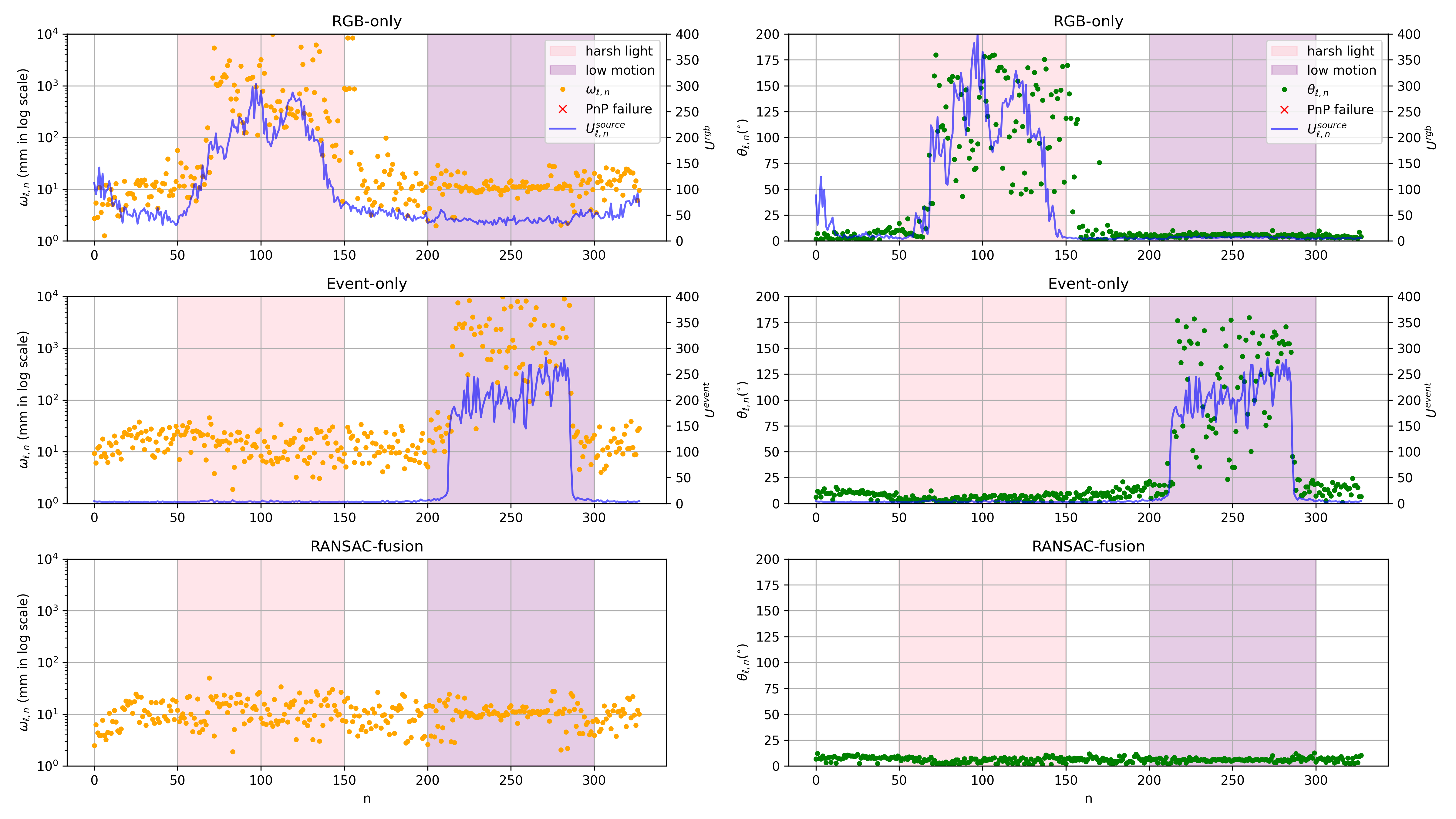}
    \caption{Error and uncertainty plots of the sequence \texttt{soho-3-close}.}
    \label{fig:appendixsoho5}
\end{figure}

\begin{figure}[H]
    \centering
    \includegraphics[width=0.95\linewidth]{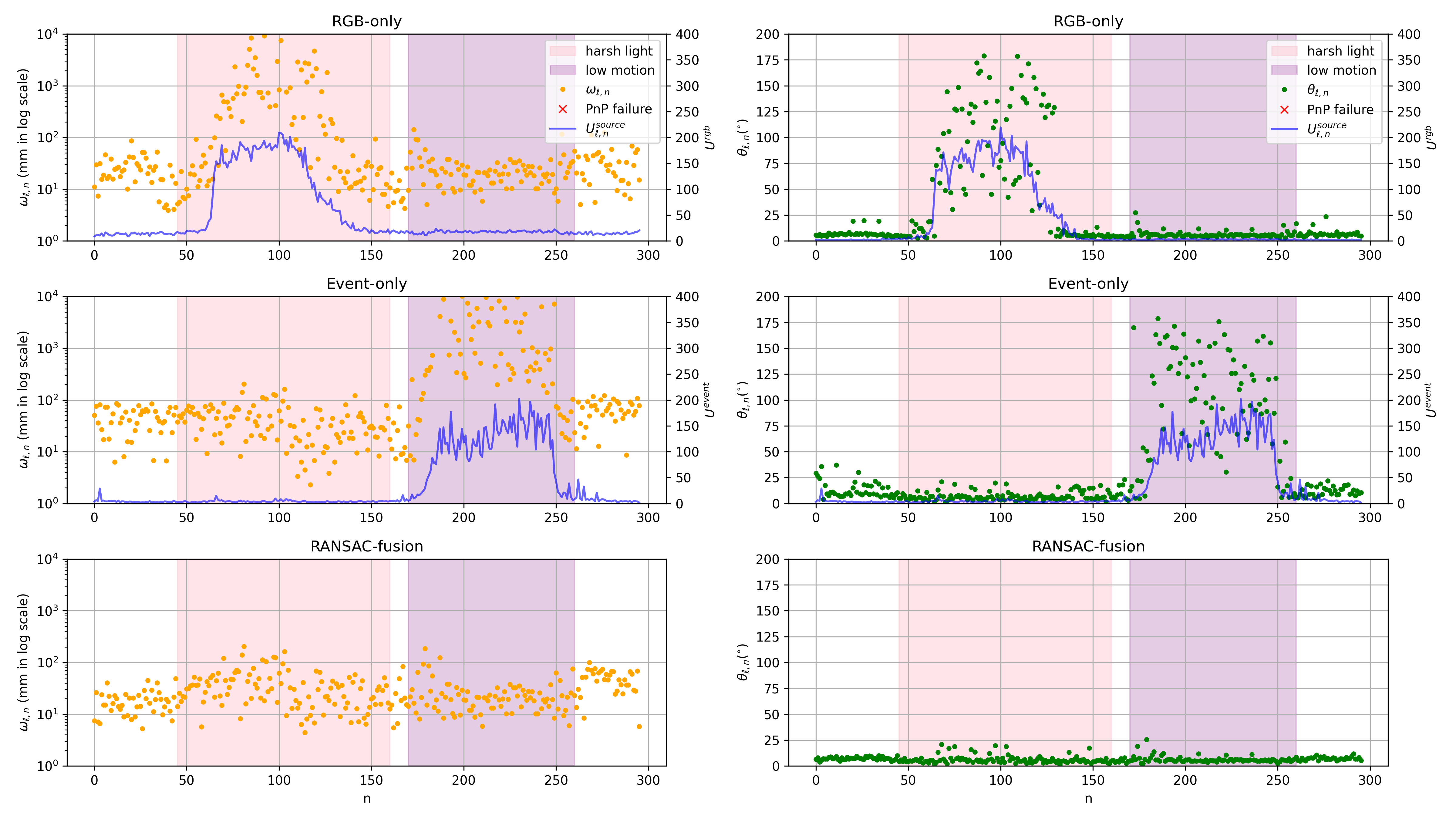}
    \caption{Error and uncertainty plots of the sequence \texttt{soho-3-far}.}
    \label{fig:appendixsoho6}
\end{figure}

\begin{figure}[H]
    \centering
    \includegraphics[width=0.95\linewidth]{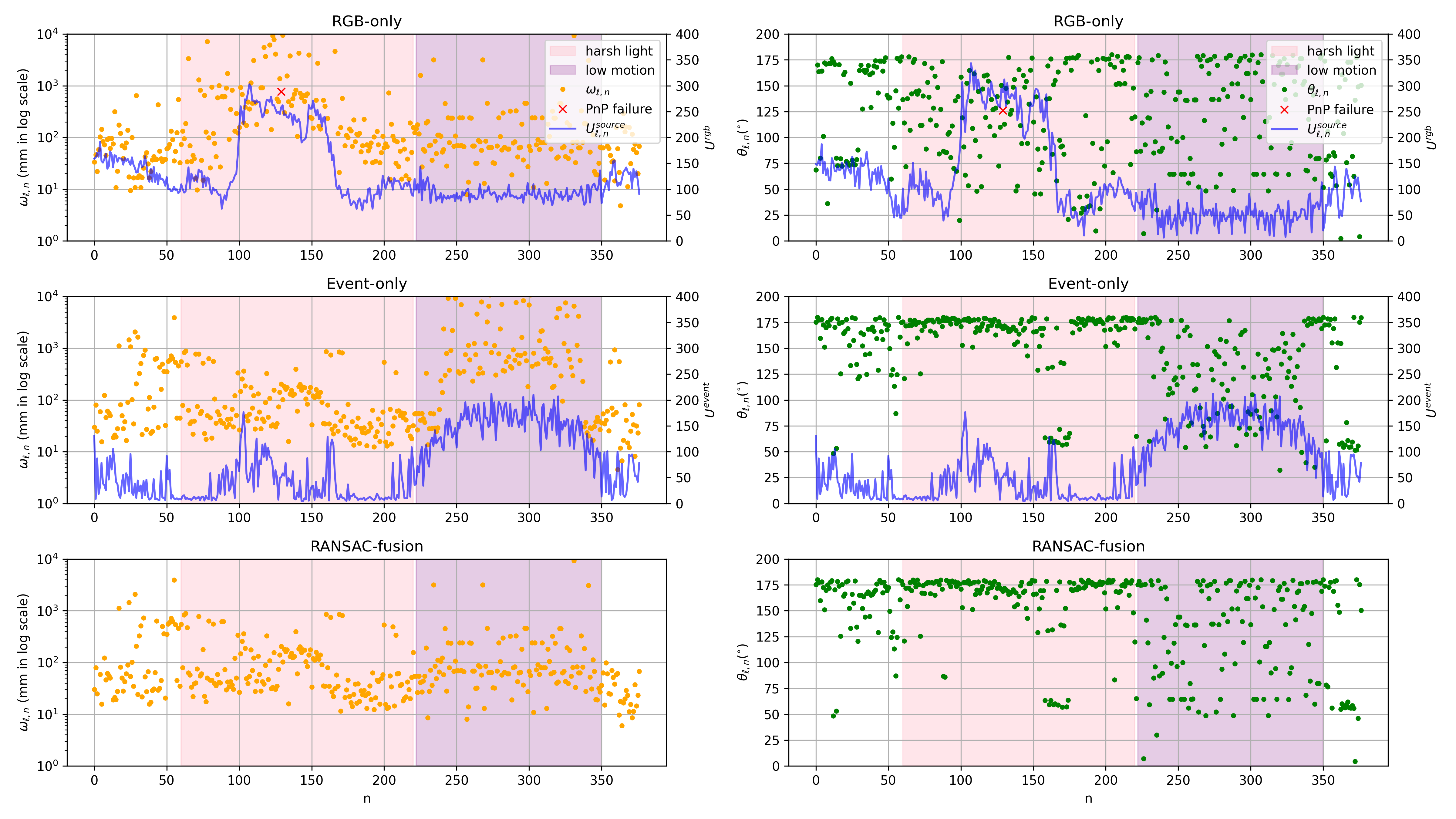}
    \caption{Error and uncertainty plots of the sequence \texttt{soho-4-close}.}
    \label{fig:appendixsoho7}
\end{figure}

\begin{figure}[H]
    \centering
    \includegraphics[width=0.95\linewidth]{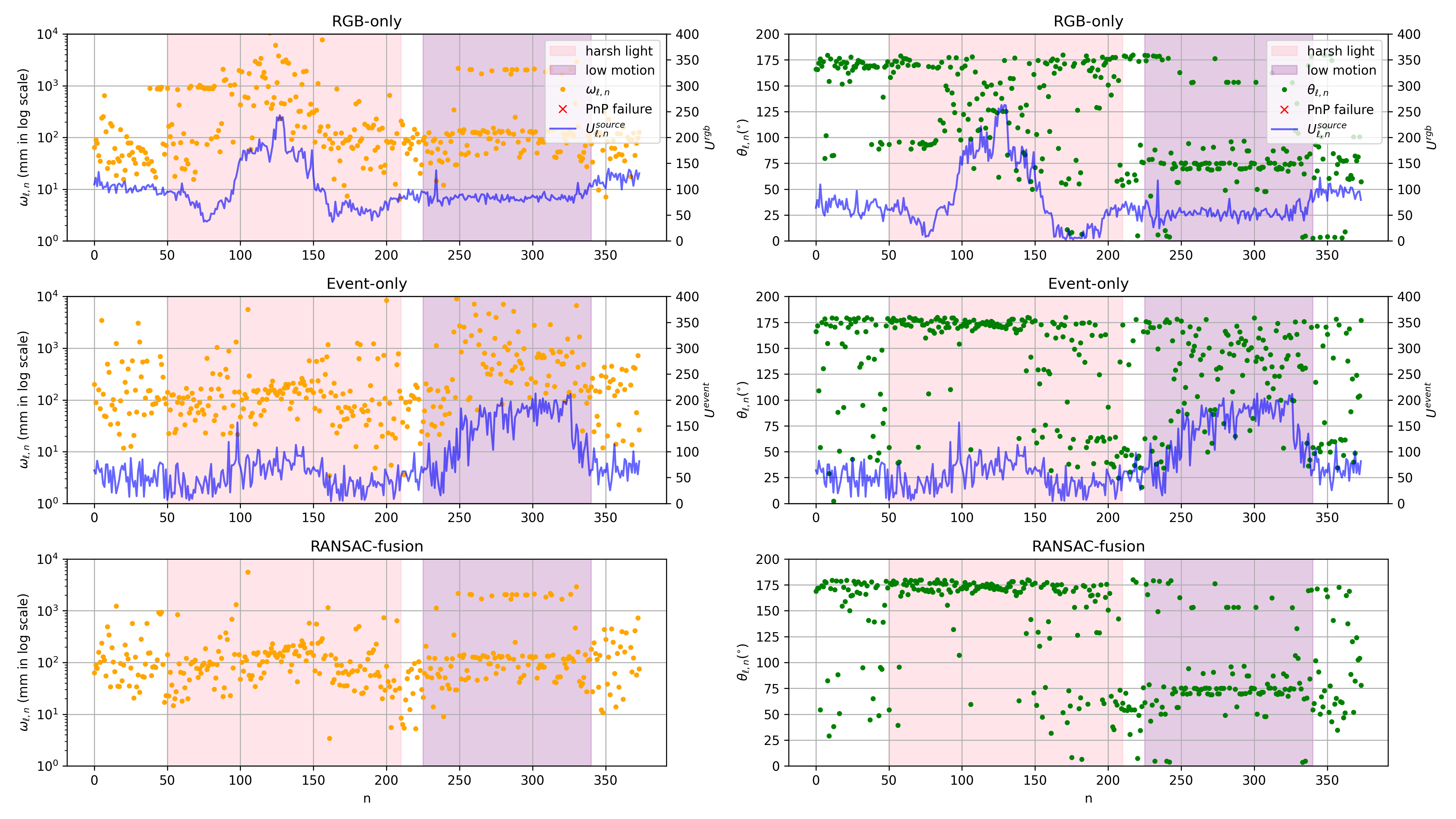}
    \caption{Error and uncertainty plots of the sequence \texttt{soho-4-far}.}
    \label{fig:appendixsoho8}
\end{figure}

\end{appendices}

\end{document}